\pdfoutput=1

\documentclass[11pt]{article}

\usepackage[final]{acl}

\usepackage{times}
\usepackage{latexsym}
\usepackage{enumitem}
\usepackage[T1]{fontenc}

\usepackage[utf8]{inputenc}
\usepackage{CJKutf8}
\usepackage{microtype}

\usepackage{inconsolata}

\usepackage{graphicx}

\usepackage{amsmath,amsfonts,bm}

\def\eqref#1{equation~\ref{#1}}

\def\1{\bm{1}}

\DeclareMathAlphabet{\mathsfit}{\encodingdefault}{\sfdefault}{m}{sl}
\SetMathAlphabet{\mathsfit}{bold}{\encodingdefault}{\sfdefault}{bx}{n}

\definecolor{color1}{HTML}{1f77b4}
\definecolor{color2}{HTML}{ff7f0e}
\definecolor{color3}{HTML}{2ca02c}
\definecolor{color4}{HTML}{d62728}
\definecolor{color5}{HTML}{9467bd}
\definecolor{color6}{HTML}{8c564b}
\definecolor{color7}{HTML}{e377c2}
\definecolor{color8}{HTML}{7f7f7f}
\definecolor{color9}{HTML}{bcbd22}
\definecolor{color10}{HTML}{17becf}
\usepackage{hyperref}
\usepackage{url}
\usepackage{xspace}
\usepackage{listings}
\usepackage{booktabs}
\usepackage{subfigure}
\usepackage{wrapfig} 
\usepackage{multirow}
\usepackage{multicol}
\usepackage{xcolor}
\usepackage{tcolorbox}
\usepackage{arydshln}
\usepackage{color}
\usepackage{comment}
\usepackage{pgfplots}
\pgfplotsset{compat=1.18}
\usepgfplotslibrary{external}

\title{LLMs Can Achieve High-quality Simultaneous Machine Translation as Efficiently as Offline}

\author{Biao Fu$^{1,3,}$\thanks{\,\, Work done during Biao Fu’s internship at Tongyi Lab.}, Minpeng Liao$^{2,}$\thanks{\,\, Corresponding author.}, Kai Fan$^{2,}$\footnotemark[2], Chengxi Li$^{2}$,\\
\textbf{Liang Zhang}$^{1}$,
\textbf{Yidong Chen}$^{1,3}$,
\textbf{Xiaodong Shi}$^{1,3,}$\footnotemark[2] \\
$^{1}$School of Informatics, Xiamen University \ $^{2}$Tongyi Lab \\
$^{3}$Key Laboratory of Digital Protection and Intelligent Processing of Intangible Cultural \\ Heritage of Fujian and Taiwan (Xiamen University), Ministry of Culture and Tourism \\
\texttt{biaofu@stu.xmu.edu.cn,mandel@xmu.edu.cn} \\ 
\texttt{\{minpeng.lmp,xiji.lcx,k.fan\}@alibaba-inc.com}
}

\begin{document}
\maketitle
\begin{abstract}

When the complete source sentence is provided, Large Language Models (LLMs) perform excellently in offline machine translation even with a simple prompt "\textit{Translate the following sentence from [src lang] into [tgt lang]:}". 
However, in many real scenarios, the source tokens arrive in a streaming manner and simultaneous machine translation (SiMT) is required, then the \textbf{efficiency} and \textbf{performance} of decoder-only LLMs are significantly limited by their auto-regressive nature. 
To enable LLMs to achieve high-quality SiMT as efficiently as offline translation, we propose a novel paradigm that includes constructing supervised fine-tuning (SFT) data for SiMT, along with new training and inference strategies. 
To replicate the token input/output stream in SiMT, the source and target tokens are rearranged into an interleaved sequence, separated by special tokens according to varying latency requirements. 
This enables powerful LLMs to learn read and write operations adaptively, based on varying latency prompts, while still maintaining efficient auto-regressive decoding. 
Experimental results show that, even with limited SFT data, our approach achieves state-of-the-art performance across various SiMT benchmarks, and preserves the original abilities of offline translation.
Moreover, our approach generalizes well to document-level SiMT setting without requiring specific fine-tuning, even beyond the offline translation model\footnote{The data and code are available at \url{https://github.com/biaofuxmu/EAST}.}. 
\end{abstract}

\section{Introduction}

\begin{figure}[t]
\centering
\includegraphics[width=\linewidth]{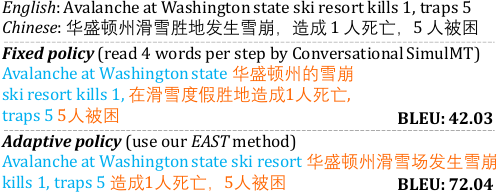}
\caption{Examples of fixed policy and adaptive policy. The fixed policy (Conversational SimulMT) reads a predetermined number of words per step, leading to translation errors due to incomplete semantic understanding (e.g., prematurely translating "Avalanche at Washington state" as "\begin{CJK}{UTF8}{gbsn}\small{华盛顿州的雪崩}\end{CJK}"). In contrast, EAST uses a adaptive policy and recognizes the incompleteness of the semantic context and continues reading more words until "ski resort", resulting in the correct translation.}
\label{fig:policy_case}
\end{figure}

\begin{figure*}[t]
\centering
\subfigure[Previous LLM SiMT with R/W policy, \emph{e.g.}, wait-3: $p(y_{t-2}|\mathbf{x}_{\leq t},\mathbf{y}_{\leq t-3})$, where KV cache needs to re-calculate as $t$ increments.]{
\label{fig:waitk_infer}
\includegraphics[width=\linewidth]{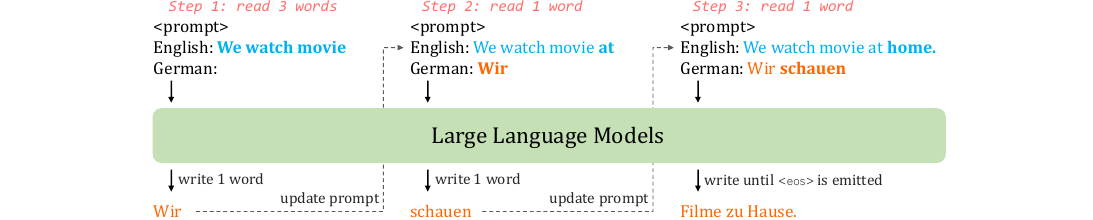}
}
~
\subfigure[The inference process of EAST in auto-regressive manner $p(\mathbf{c}^y_t|\mathbf{c}^x_1, \mathbf{c}^y_1, ..., \mathbf{c}^x_t)$, where $\mathbf{c}^{\cdot}_t$ is the $t$-th chunk of source or target.]{
\label{fig:east_infer}
\includegraphics[width=\linewidth]{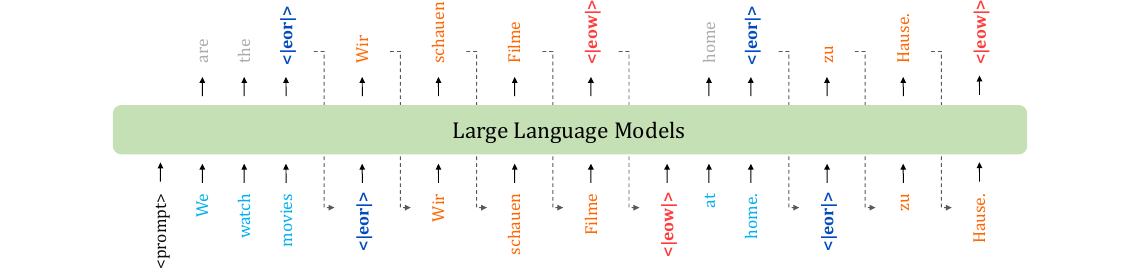}
}
\caption{Comparison between existing LLM-based SiMT methods and our EAST. (a) Existing LLM-based SiMT typically reuses the sequence organization $\mathbf{x}_{\leq t},\mathbf{y}_{\leq t^\prime}$, leading the prompt to generate new target tokens always changing. (b) Our inference follows the nature of auto-regressive decoding without recalculating the KV cache. For simplicity, we use \texttt{<|eor|>} and \texttt{<|eow|>} to represent the special tokens without hurting readability.}
\label{fig:infer_comparison}
\end{figure*}
Simultaneous machine translation (SiMT) \citep{gu-etal-2017-learning} is a critical technique for enabling seamless cross-linguistic communication in real-time scenarios, such as international conferences.
Unlike offline machine translation (OMT), where the entire source sentence is available before translation begins, SiMT systems start translating before receiving the complete input, achieving a balance between translation quality and latency.

Large language models (LLMs) have achieved significant advances in OMT, demonstrating impressive capacity when translating full sentences in offline settings \citep{xu2024a,xu2024contrastive,ye2025how}.
However, their application to SiMT remains underexplored and faces several significant challenges.
First, most existing SiMT models \citep{zhao-etal-2023-adaptive,guo2024agent,raffel2024simultaneous} are typically trained on OMT data due to the scarcity of SiMT-specific datasets. 
This training setup does not align well with the demands of SiMT, which hinders the model's ability to learn how to translate effectively with incomplete input \citep{wang-etal-2023-better,sakai2024simultaneous}.
Second, many SiMT approaches focus on optimizing prompt structures to simulate SiMT for LLMs \citep{wang2023simultaneous,koshkin2024llms,koshkin2024transllama,guo2024agent,agostinelli-etal-2024-simul,cheng2024towards}, as shown in Figure \ref{fig:waitk_infer}, which requires recomputing the key-value (KV) cache as the prompt changes continuously with the update of the source and target. 
This recomputation significantly increases the computational cost and inference latency, limiting the efficiency of SiMT systems \citep{raffel2024simultaneous}. 
Lastly, LLMs-based methods typically employs fixed policies \citep{wang2023simultaneous,agostinelli-etal-2024-simul,sakai2024simultaneous,wang2024conversational,raffel2024simultaneous}, such as the wait-$k$, for their simplicity. 
However, these methods fail to adaptively adjust its read/write actions based on sentence structure and context, leading to suboptimal translation quality, as shown in Figure \ref{fig:policy_case}.

\vspace{-1mm}
In this paper, we introduce \textbf{EAST}, an \textbf{E}fficient and \textbf{A}daptive \textbf{S}imultaneous machine \textbf{T}ranslation method with LLMs, which aims to achieve high-quality SiMT as efficiently as offline translation. 
Specifically, we first leverage the instruction-following capability of LLMs to generate the SiMT data \citep{sakai2024simultaneous} with different latency levels (low, medium, and high). 
Two SiMT datasets are constructed for supervised fine-tuning (SFT), including the German-English dataset \textbf{SiMT-De-En-660K} and the multilingual dataset \textbf{SiMT-Multi-90K}. 
We structure the SFT data by alternating between the source and target segments of the generated SiMT data and introducing two special tokens (\texttt{<|end-of-read|>} and \texttt{<|end-of-write|>}) as explicit read-write signals. 
By performing SFT on this structured data on the LLMs, it can learn to effectively determine when to read more source input and when to generate the translation.
A two-stage fine-tuning process is conducted to enhance its multilingual translation capabilities, with full-weight fine-tuning on the SiMT-De-En-660K dataset, followed by LoRA \citep{hu2022lora} fine-tuning on a combination of the SiMT-Multi-90K and Off-Multi-120K datasets.
During inference, EAST employs an adaptive read-write policy that aligns with its SFT recipe. 
The model predicts token-by-token and then dynamically switches between read and write actions based on whether the predicted token is a read or write signal. 
Due to the auto-regressiveness of our token input/output (I/O) sequence, where new source input and target translations are incrementally appended, EAST can efficiently reuse the KV cache without modifying historical sequence. 
This significantly reduces computational costs and inference latency, improving the overall efficiency of SiMT. 
A comparison to highlight the main difference between the previous LLM SiMT and ours is shown in Figure~\ref{fig:infer_comparison}. 
Experimental results show that EAST achieves high-quality SiMT across eight translation directions and near-offline decoding speeds, without compromising offline translation performance, and generalizes well to document-level SiMT.

Our contributions are summarized as follows:

\begin{itemize} 
    \item We construct two novel latency-aware datasets, including a German-English dataset (SiMT-De-En-660K) and a multilingual dataset (SiMT-Multi-90K), where latency-awareness is often neglected in SiMT studies.
    \item We propose a novel LLM-based adaptive read/write policy which achieves high-quality SiMT as efficiently as offline model. To the best of our knowledge, this is the first efficient and adaptive LLM-based SiMT method.
    \item Experimental results on multilingual and document-level test sets demonstrate the effectiveness of our method, where document-level evaluation is particularly underexplored in prior work.
    \item Our findings reveal that only 10K SiMT examples may be sufficient to achieve commendable translation quality, offering valuable insights for future research. 
\end{itemize}

\section{Related Work}
\textbf{Traditional SiMT} 
SiMT requires starting translation before the full source sentence is available, aiming to balance translation quality and latency. To achieve this, a read-write policy is introduced to determine whether to wait for more input or begin translating.
Traditional SiMT models are often built on encoder-decoder architectures, with fixed or trainable policies. 
The widely studied fixed wait-$k$ policy \citep{ma-etal-2019-stacl,elbayad2020efficient,zhang-feng-2021-universal,zhang-etal-2023-training,fu-etal-2023-adapting} is simple but struggles with complex contexts or non-monotonic language pairs \citep{zhang-etal-2022-learning}.
Adaptive policies, which dynamically determine when to read and write, offer improved translation quality.
To enable models to learn effective read-write decisions, a variety of techniques have been applied, including reinforcement learning \citep{gu-etal-2017-learning,arthur-etal-2021-learning,miao-etal-2021-generative}, dynamic programming \citep{miao-etal-2021-generative,liu-etal-2021-cross,fu-etal-2024-wav2vec}, data augmentation \cite{zhang-etal-2020-learning-adaptive,deng2023imporving}, information transport theory \citep{zhang-feng-2022-information}, Hidden Markov Models \citep{zhang2023hidden}, and decoder-only architecture~\citep{guo-etal-2024-decoder}.

\noindent \textbf{LLM-based SiMT}
Recent studies have explored leveraging LLMs for SiMT, but traditional adaptive policies in encoder-decoder architectures not well-suited for LLMs.
Some approaches \citep{wang2023simultaneous,koshkin2024llms,koshkin2024transllama,agostinelli-etal-2024-simul} optimize prompts with a fixed wait-$k$ policy. Others, like the Agent-SiMT \cite{guo2024agent}, combine traditional adaptive SiMT models to guide read/write decisions.
However, these approaches face issues: updating prompts during inference prevents KV cache reuse, leading to increased recomputation and latency, and Agent-SiMT requires training an additional SiMT model, complicating the process.
Recent efforts \citep{sakai2024simultaneous,cheng2024towards} use LLMs to generate SiMT data for adaptive policy learning but still struggle with inefficiencies in LLM-based SiMT systems.

To improve translation efficiency, \citet{wang2024conversational} introduce the Conversational SimulMT framework, which employs a multi-turn dialogue decoding approach with generating SFT data by segmenting parallel sentences with an alignment tool. 
However, the method employs a fixed policy during inference that reads a fixed number of words at each step, leading to a mismatch with the fine-tuning process. 
The differences between Conversational SimulMT and our method are described in the Appendix \ref{app:key_diff}.
Moreover, SimulMask \cite{raffel2024simultaneous} improves prompt-based efficiency by introducing a policy-specific attention mask during fine-tuning. It mimics inference behavior, limiting target token attention to the relevant source prompt. However, its complex masking requires prior knowledge of policy decisions, making it unsuitable for adaptive policies.

In addition, these SiMT methods primarily focus on leveraging the translation capabilities of LLMs, without exploring the adaptive read-write policies and generalization abilities.
Unlike previous methods, EAST enables the LLMs to learn adaptive read-write policies in various latency requirements, and utilizes an interleaved text structure to significantly improve the inference efficiency of the LLMs while maintaining consistency between fine-tuning and inference.

\section{Methods}
\label{sec:method}

In this paper, we propose EAST, an \underline{E}fficient and \underline{A}daptive \underline{S}imultaneous machine \underline{T}ranslation method with LLMs, which involves three key components: the construction of SiMT data, the training of the LLMs on the SFT data, and the inference with adaptive read-write policy.

\subsection{SiMT Data Curation via Latency-aware Chunk Segmentation}
\label{sec:simt_data}

The availability of SiMT-specific datasets is scarce, and the annotation by professional interpreters is time-consuming and expensive. 
To address this issue, we leverage the powerful instruction-following capability of LLMs after RLHF \citep{ouyang2022training} (\emph{e.g.}, GPT-4 \citep{openai2024gpt4technicalreport}) and design a prompt that instructs LLMs to act as a professional simultaneous interpreter, segmenting sentences into independent semantic chunks and generating corresponding translations for each chunk.

In practice, SiMT must accommodate varying latency requirements depending on different use cases, such as live broadcasts that prioritize low latency and formal conferences that demand high-quality translation with higher latency. 
Importantly, different latencies naturally influence how sentences are segmented, reordered, and translated. 
Therefore, we prompt LLMs to generate SiMT data at three latency levels: "low", "medium", and "high". 
The prompt template is provided in Figure~\ref{fig:gpt_prompt} in Appendix. 
Concretely, given a language pair $\mathbf{x}_{1:T_x}, \mathbf{y}_{1:T_y}$, the LLM output of the proposed low latency prompt can be represented as follows:
\begin{align}
    \mathbf{x}_{1:T_x} = [\mathbf{c}_1^x, \cdots, \mathbf{c}_{T_{low}}^x], \\ \mathbf{y}_{1:T_y} = [\mathbf{c}_1^y, \cdots, \mathbf{c}_{T_{low}}^y], \label{eq:simt_data}
\end{align}
where $\mathbf{c}^{[\cdot]}_t$ is the $t$-th semantic chunk of source or target. 
With simple length filtering, the two chunk sequences should be well aligned with the same length. 
Similarly, we can obtain the medium and high latency output. 
In general, for the same pair, we have $T_{low} \geq T_{medium} \geq T_{high}$.

In this study, we curated a dataset of 660K SiMT samples by extracting language pairs from the WMT15 De-En training data, allocating one-third of the samples to each latency requirement. 
While existing LLM-based SiMT methods typically train separate models for different language pairs \citep{guo2024agent,raffel2024simultaneous}, they often overlook the inherent multilingual capabilities of LLMs. 
In contrast, we constructed a smaller multilingual SiMT dataset of 90K samples encompassing eight translation directions. 

\subsection{Training LLMs with SFT} 
\label{sec:sft}

To tackle the challenge proposed at the beginning of this section, we propose a two-stage SFT training process on the two curated SiMT datasets.

\textbf{Stage I: Activate SiMT of LLMs} The objective of this stage is to teach the LLMs how to perform adaptive simultaneous translation by learning when to read and write in our designed format. 
To enable the model to learn these adaptive behaviors, we reorganized the aligned chunks in the SiMT data by interleaving between source and target chunks and introduce two special tokens (\texttt{<|end-of-read|>} and \texttt{<|end-of-write|>}), \emph{i.e.}, 
\begin{multline}
[\mathbf{c}_1^x, \texttt{<|eor|>}, \mathbf{c}_1^y, \texttt{<|eow|>}, \cdots, \\
\mathbf{c}_T^x, \texttt{<|eor|>}, \mathbf{c}_T^y,\texttt{<|eow|>} ].
\end{multline}
The special tokens act as explicit signals for the model to transition between reading and writing. 
The SiMT annotation process shows that each chunk contains enough semantic meaning for LLMs to carry out translations, ensuring that sequence reorganization does not lead to any loss of information for the model's reading or writing decisions. 
Since the annotation process also encodes the degree of fragmentation into latency indicator tokens—"low", "medium" or "high", the SFT can effectively guide the model to adapt to varying latency requirements. 
Figure~\ref{fig:data_example} provides a comprehensive example of the SFT data.

We train for one epoch during this stage on the larger SiMT-De-En-660K, employing full parameter tuning. 
As SiMT defined in our proposed format is generally a novel task for LLMs, full parameter tuning ensures that the LLM can effectively and successfully learn the auto-regressive SiMT. 
Training for just one epoch helps mitigate the risk of overfitting during the full parameter tuning.

\textbf{Stage II: Generalize to Multilingual SiMT} As the LLM acquires its auto-regressive SiMT capability in Stage I, its inherent multilingual proficiency enables it to generalize to multilingual SiMT, even with limited SFT data. Consequently, we apply LoRA \citep{hu2022lora} fine-tuning to a smaller multilingual dataset of 90K instances including eight language directions. 
Additionally, during this stage, we incorporate an OMT task to bolster the model's ability to translate full sentences and enhance overall translation performance. 
In fact, we can view offline translation as a specific instance of SiMT by treating the entire sentence as a complete semantic chunk, \emph{e.g.}, $\mathbf{x}_{1:T} = \mathbf{c}_1^x$.

\noindent \textbf{Loss} In previous LLM-based SiMT methods \cite{wang2024conversational}, loss calculation \emph{w.r.t.} source text is typically masked out, as it does not contribute to the training. 
However, in our case, the cross-entropy loss is calculated on both the target text and the source text, as well as on special tokens. 
The primary goal is to align with the auto-regressive design of interleaved sequences and establish the appropriate reading and writing timing.

\subsection{Efficient Inference for Adaptive SiMT}
Unlike Conversational SimulMT \cite{wang2024conversational}, which adopts a fixed policy  (e.g., reading a fixed number of tokens at each step) during inference and suffers from a mismatch between training and inference phases, EAST performs auto-regressive token-by-token prediction aligned with the training process as shown in Figure \ref{fig:east_infer}. 
The process unfolds in two main phases:

\textbf{Read-Predict-Discard} During the read phase, the model sequentially receives source tokens and predicts the next token. 
If the predicted token is not \texttt{<|end-of-read|>}, it is discarded, and the next source token is appended to the current source chunk. 
Once \texttt{<|end-of-read|>} is predicted, the model transitions to the translating phase. 
Note that the discarding operation in the read phase does not violate the incremental appending of contexts, enabling EAST to efficiently utilize KV-cache for faster generation. 

\textbf{Predict-Append} Once the model enters the translation phase, it directly begins predicting the next token. 
If the predicted token is not \texttt{<|end-of-write|>}, it is appended to the current target chunk. 
When \texttt{<|end-of-write|>} is predicted, the model completes the current translation and returns to the reading phase. 

Similar to the training phase, inference controls latency through indicator tokens—"low", "medium", or "high". 
Interestingly, an \textbf{Interpolation Effect} is observed, allowing for generalization to other latency levels using indicator tokens such as "low-medium" or "medium-high". 
Consequently, we can gather 3 to 5 observations to draw the BLEU-AL curve.

\section{Experiments}
\label{sec:experiments}

\subsection{Experimental Settings}

\textbf{Datasets} 
Following ALMA \cite{xu2024a}, we collect test data from WMT17 to WMT21, covering the 8 language directions: De$\leftrightarrow$En, Zh$\leftrightarrow$En, Ru$\leftrightarrow$En, and Cs$\leftrightarrow$En, and refer to this collection as \textbf{Off-Multi-120K} for OMT training. 
As introduced in the previous section, the primary SiMT SFT dataset to initiate novel task learning is \textbf{SiMT-De-En-660K}, derived from the WMT15 De$\rightarrow$En training dataset. 
In addition, we construct a smaller multilingual SiMT SFT dataset, \textbf{SiMT-Multi-90K}, derived from Off-Multi-120K dataset\footnote{See Appendix \ref{apd:data_filter} for details of data processing.}. 
As shown in Table \ref{tab:data_statistics} of Appendix, the \textbf{sentence-level test data} is extracted from WMT22 across the same 8 translation directions as the OMT data. 
The majority of existing research primarily focuses on sentence-level evaluation. 
However, in many real-world applications, such as speech delivery, the input for SiMT often comes at the document level rather than isolated sentences. 
Moreover, LLMs have demonstrated impressive capabilities in long-form generation.
Thus, we directly evaluate EAST on WMT22 \textbf{document-level test data} without additional fine-tuning.

\noindent \textbf{Metrics} 
For quality evaluation, we use automatic metric--SacreBLEU\footnote{\url{https://github.com/mjpost/sacrebleu}} to compute the corpus-level BLEU, along with neural evaluation metrics BLEURT\footnote{\texttt{BLEURT-20}} \citep{sellam-etal-2020-bleurt,pu-etal-2021-learning} and COMET\footnote{\texttt{wmt22-comet-da}} \citep{rei-etal-2020-comet,rei-etal-2022-cometkiwi}. 
For latency evaluation, we adopt Average Latency (AL) \citep{ma-etal-2019-stacl}, 
computation-aware AL (AL-CA)\footnote{The AL-CA metric is calculated by adding the machine processing time to the policy delay .}, 
and Length-Adaptive AL (LAAL) \citep{papi-etal-2022-generation}.
In addition, we use Word Wall Time (WWT) \cite{wang2024conversational} to evaluate the model's decoding speed by calculating the actual inference time per word.
For the implementation details of our model, please refer to Appendix \ref{apd:train_detail}.

\begin{figure}[t]
\pgfplotsset{
    every axis y label/.append style={at={(-0.1,0.5)}},
    every axis/.append style={line width=0.8pt},
}
\centering
\resizebox{0.45\textwidth}{!}{
\begin{tikzpicture}[baseline]
\begin{axis}[
    ylabel=SacreBLEU,
    xlabel=AL,
    enlargelimits=0.03,
    font=\small,
    width=10cm,height=8cm,
    legend cell align=left,
    legend style={font=\tiny,
    at={(0.69,0.01)},
    anchor=south,
    legend columns=1},
    xmajorgrids=true,
    ymajorgrids=true,
    grid style=dashed,
    xmin=2, 
    xmax=10,
    ymin=24,
    ymax=37,
    ytick={20,24,28,32,36},
]

\addplot[color=color4, dashed, mark=triangle*, mark size=1.8pt,line width=0.8pt] coordinates {(2.68,32.46)(3.06,33.12)(4.77,35.21)(5.33,35.55)(7.78,36.62)};

\addplot[color=color4, mark=triangle*, mark size=1.8pt,line width=0.8pt] coordinates {(3.18,32.55)(3.46,32.8)(5.84,34.62)(7.05,34.93)(9.73,35.49)}; 

\addplot[color=color1,mark=triangle*, mark size=1.8pt,line width=0.8pt] coordinates {(2.3,30.03)(3.48,31.99)(4.71,33.06)(5.88,33.69)(7.01,34.15)(9.29,35.12)};  

\addplot[color=color2,mark=triangle*, mark size=1.8pt,line width=0.8pt] coordinates {(2.67,27.93)(4.09,30.12)(5.21,31.3)(6.28,32.38)(7.36,32.60)(8.38,32.96)}; 

\addplot[color=color3,mark=triangle*, mark size=1.8pt,line width=0.8pt] coordinates {(2.94,28.30)(4.20,30.60)(5.26,31.80)(7.01,32.90)(9.87,33.40)};

\addplot[color=color5,mark=*, mark size=1.4pt,line width=0.8pt] coordinates {(1.52,24.74)(2.73,28.48)(3.73,30.17)(5.49,31.11)(7.03,31.42)(9.22,31.65)(12.33,31.92)};

\addplot[color=color7,mark=*, mark size=1.4pt,line width=0.8pt] coordinates {(2.15,24.91)(2.45,27.50)(3.16,29.13)(4.34,30.01)(6.17,30.98)(8.59,31.41)(12.09,31.58)}; 

\addplot[color=color10,mark=*, mark size=1.4pt,line width=0.8pt] coordinates {(2.15,24.88)(2.69,28.25)(3.74,29.50)(5.28,30.54)(7.21,31.00)(9.50,31.22)(12.39,31.21)};

\legend{EAST-Stage-I w/ Llama3, EAST-Stage-I w/ Llama2, Conversational SimulMT w/ SiMT-De-En-660K, Conversational SimulMT, Agent-SiMT, SM$^2$-Bi, Mono-KD, ITST}

\end{axis}
\end{tikzpicture}}
\caption{BLEU vs. AL on WMT15 De$\rightarrow$En test set. }
\label{fig:result_wmt15_bleu}
\end{figure}
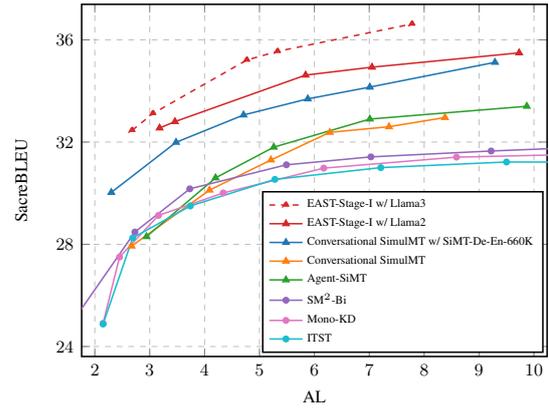
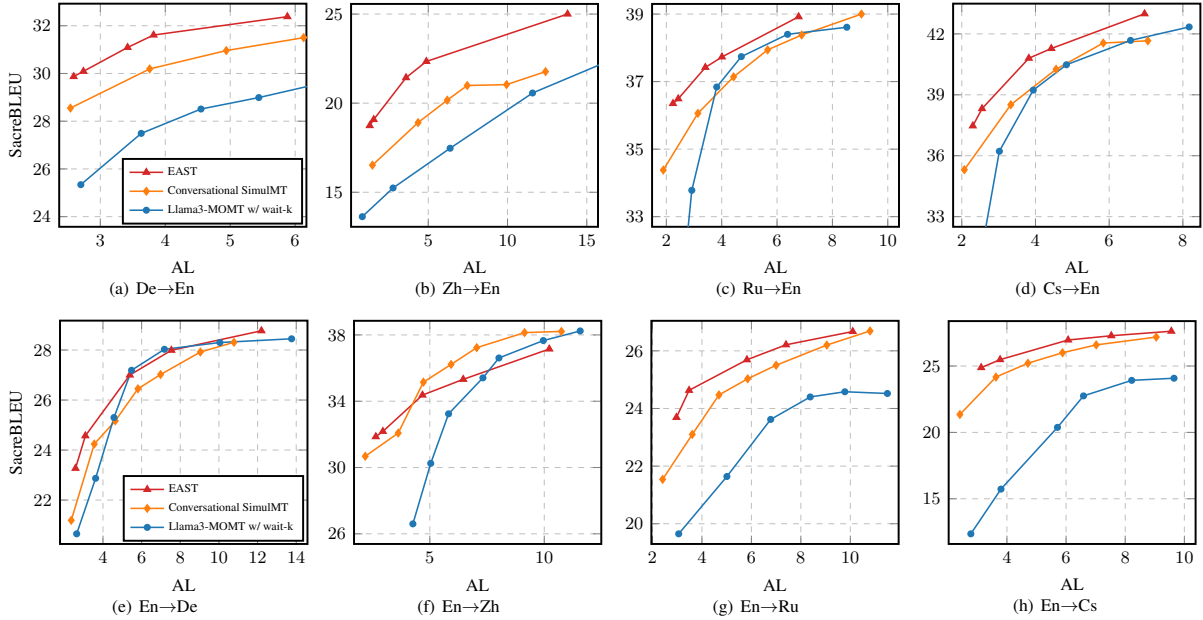
\begin{figure*}[t]
\pgfplotsset{
    every axis y label/.append style={at={(-0.12,0.5)}},
    every axis/.append style={line width=1.0pt},
}
\centering

\resizebox{0.26\textwidth}{!}{
\subfigure[De$\rightarrow$En]{
\begin{tikzpicture}[baseline]
\begin{axis}[
    ylabel=SacreBLEU,
    xlabel=AL,
    enlargelimits=0.05,
    font=\small,
    width=6cm,height=5.6cm,
    legend cell align=left,
    legend style={font=\tiny,
    at={(0.62,0.01)},
    anchor=south,
    legend columns=1},
    xmajorgrids=true,
    ymajorgrids=true,
    grid style=dashed,
    ymin=24,ymax=32.5,
    xmax=6,
]
\addplot[color=color4,mark=triangle*, mark size=1.8pt,line width=0.8pt] coordinates {(2.59,29.87)(2.74,30.09)(3.42,31.09)(3.82,31.61)(5.88,32.38)}; 

\addplot[color=color2,mark=diamond*, mark size=1.8pt,line width=0.8pt] coordinates {(2.54,28.55)(3.76,30.19)(4.94,30.96)(6.13,31.5)(7.27,32.09)(9.29,32.38)}; 

\addplot[color=color1,mark=*, mark size=1.4pt,line width=0.8pt] coordinates {(2.7,25.34)(3.63,27.49)(4.55,28.51)(5.44,28.99)(6.29,29.52)(7.83,29.66)(8.52,29.68)}; 

\legend{EAST, Conversational SimulMT, Llama3-MOMT w/ wait-k}

\end{axis}
\end{tikzpicture}}}
\hspace{-2mm}
\resizebox{0.248\textwidth}{!}{
\subfigure[Zh$\rightarrow$En]{
\begin{tikzpicture}[baseline]
\begin{axis}[
    xlabel=AL,
    enlargelimits=0.05,
    width=6cm,height=5.6cm,
    font=\small,
    legend style={font=\tiny,
    at={(0.5,0.35)},
    anchor=south,
    legend columns=1},
    xmajorgrids=true,
    ymajorgrids=true,
    grid style=dashed,
    xmax=15,
]

\addplot[color=color4,mark=triangle*, mark size=1.8pt,line width=0.8pt] coordinates {(1.31,18.74)(1.56,19.08)(3.6,21.43)(4.88,22.34)(13.79,25.0)}; 

\addplot[color=color2,mark=diamond*, mark size=1.8pt,line width=0.8pt] coordinates {(1.48,16.52)(4.36,18.91)(6.2,20.17)(7.46,20.99)(9.94,21.04)(12.4,21.77)}; 

\addplot[color=color1,mark=*, mark size=1.4pt,line width=0.8pt] coordinates {(0.84,13.63)(2.78,15.24)(6.38,17.47)(11.58,20.57)(15.91,22.19)(19.22,22.74)}; 

\end{axis}
\end{tikzpicture}}
}
\hspace{-2mm}
\resizebox{0.248\textwidth}{!}{
\subfigure[Ru$\rightarrow$En]{
\begin{tikzpicture}[baseline]
\begin{axis}[
    xlabel=AL,
    enlargelimits=0.05,
    width=6cm,height=5.6cm,
    font=\small,
    legend style={font=\tiny,
    at={(0.64,0.45)},
    anchor=south,
    legend columns=1},
    xmajorgrids=true,
    ymajorgrids=true,
    grid style=dashed,
    xmax=10,
    ymin=33,
    ytick={33,35,37,39},
]
\addplot[color=color4,mark=triangle*, mark size=1.8pt,line width=0.8pt] coordinates {(2.24,36.35)(2.43,36.49)(3.41,37.42)(4.01,37.73)(6.78,38.92)}; 

\addplot[color=color2,mark=diamond*, mark size=1.8pt,line width=0.8pt] coordinates {(1.89,34.38)(3.14,36.06)(4.43,37.14)(5.66,37.94)(6.89,38.38)(9.05,39.0)}; 

\addplot[color=color1,mark=*, mark size=1.4pt,line width=0.8pt] coordinates {(2.01,26.25)(2.92,33.78)(3.82,36.84)(4.71,37.74)(6.38,38.4)(8.52,38.61)};

\end{axis}
\end{tikzpicture}}
}
\hspace{-2mm}
\resizebox{0.245\textwidth}{!}{
\subfigure[Cs$\rightarrow$En]{
\begin{tikzpicture}[baseline]
\begin{axis}[
    xlabel=AL,
    enlargelimits=0.05,
    width=6cm,height=5.6cm,
    font=\small,
    legend style={font=\tiny,
    at={(0.5,0.35)},
    anchor=south,
    legend columns=1},
    xmajorgrids=true,
    ymajorgrids=true,
    grid style=dashed,
    ymin=33,
    ytick={33,36,39,42},
]
\addplot[color=color4,mark=triangle*, mark size=1.8pt,line width=0.8pt] coordinates {(2.3,37.47)(2.55,38.32)(3.82,40.8)(4.43,41.28)(6.96,42.99)}; 

\addplot[color=color2,mark=diamond*, mark size=1.8pt,line width=0.8pt] coordinates {(2.07,35.31)(3.33,38.51)(4.56,40.26)(5.84,41.55)(7.05,41.66)}; 

\addplot[color=color1,mark=*, mark size=1.4pt,line width=0.8pt] coordinates {(2.1,26.86)(3.02,36.22)(3.94,39.23)(4.84,40.48)(6.58,41.68)(8.17,42.34)}; 

\end{axis}
\end{tikzpicture}}
}
~
\resizebox{0.268\textwidth}{!}{
\subfigure[En$\rightarrow$De]{
\begin{tikzpicture}[baseline]
\begin{axis}[
    ylabel=SacreBLEU,
    xlabel=AL,
    enlargelimits=0.05,
    font=\small,
    width=6cm,height=5.6cm,
    legend style={font=\tiny,
    at={(0.62,0.01)},
    anchor=south,
    legend columns=1},
    legend cell align=left,
    xmajorgrids=true,
    ymajorgrids=true,
    grid style=dashed,
    xmax = 14.0,
    xtick={4,6,8,10,12,14},
]
\addplot[color=color4,mark=triangle*, mark size=1.8pt,line width=0.8pt] coordinates {(2.58,23.27)(3.09,24.57)(5.39,27.0)(7.54,27.99)(12.2,28.77)}; 

\addplot[color=color2,mark=diamond*, mark size=1.8pt,line width=0.8pt] coordinates {(2.36,21.19)(3.55,24.24)(4.64,25.17)(5.81,26.45)(6.98,27.02)(9.04,27.92)(10.78,28.31)}; 

\addplot[color=color1,mark=*, mark size=1.4pt,line width=0.8pt] coordinates {(2.64,20.65)(3.62,22.87)(4.57,25.3)(5.48,27.19)(7.18,28.03)(10.05,28.3)(13.76,28.45)}; 3 4 6 8 15 20

\legend{EAST, Conversational SimulMT, Llama3-MOMT w/ wait-k}

\end{axis}
\end{tikzpicture}}
}
\hspace{-3mm}
\resizebox{0.245\textwidth}{!}{
\subfigure[En$\rightarrow$Zh]{
\begin{tikzpicture}[baseline]
\begin{axis}[
    xlabel=AL,
    enlargelimits=0.05,
    font=\small,
    width=6cm,height=5.6cm,
    legend style={font=\tiny,
    at={(0.5,0.35)},
    anchor=south,
    legend columns=1},
    xmajorgrids=true,
    ymajorgrids=true,
    grid style=dashed,
    xmax=12,
    ymin=26,
    ytick={26,30,34,38},
]
\addplot[color=color4,mark=triangle*, mark size=1.8pt,line width=0.8pt] coordinates {(2.64,31.86)(2.95,32.18)(4.68,34.37)(6.46,35.32)(10.22,37.15)}; 

\addplot[color=color2,mark=diamond*, mark size=1.8pt,line width=0.8pt] coordinates {(2.18,30.68)(3.62,32.08)(4.72,35.15)(5.93,36.22)(7.05,37.23)(9.14,38.14)(10.75,38.21)}; 

\addplot[color=color1,mark=*, mark size=1.4pt,line width=0.8pt] coordinates {(4.26,26.6)(5.04,30.25)(5.82,33.24)(7.32,35.41)(8.02,36.61)(9.96,37.66)(11.57,38.24)}; 

\end{axis}
\end{tikzpicture}}
}
\hspace{-2mm}
\resizebox{0.245\textwidth}{!}{
\subfigure[En$\rightarrow$Ru]{
\begin{tikzpicture}[baseline]
\begin{axis}[
    xlabel=AL,
    enlargelimits=0.05,
    font=\small,
    width=6cm,height=5.6cm,
    legend style={font=\tiny,
    at={(0.64,0.45)},
    anchor=south,
    legend columns=1},
    xmajorgrids=true,
    ymajorgrids=true,
    grid style=dashed,
]
\addplot[color=color4,mark=triangle*, mark size=1.8pt,line width=0.8pt] coordinates {(2.97,23.69)(3.48,24.63)(5.82,25.69)(7.4,26.21)(10.1,26.67)}; 

\addplot[color=color2,mark=diamond*, mark size=1.8pt,line width=0.8pt] coordinates {(2.41,21.54)(3.61,23.1)(4.69,24.47)(5.85,25.03)(7.0,25.5)(9.05,26.2)(10.8,26.69)}; 

\addplot[color=color1,mark=*, mark size=1.4pt,line width=0.8pt] coordinates {(3.06,19.65)(5.01,21.64)(6.78,23.62)(8.38,24.4)(9.78,24.58)(11.5,24.52)};

\end{axis}
\end{tikzpicture}}
}
\hspace{-2mm}
\resizebox{0.248\textwidth}{!}{
\subfigure[En$\rightarrow$Cs]{
\begin{tikzpicture}[baseline]
\begin{axis}[
    xlabel=AL,
    enlargelimits=0.05,
    font=\small,
    width=6cm,height=5.6cm,
    legend style={font=\tiny,
    at={(0.5,0.35)},
    anchor=south,
    legend columns=1},
    xmajorgrids=true,
    ymajorgrids=true,
    grid style=dashed,
    xmax=10.0,
]
\addplot[color=color4,mark=triangle*, mark size=1.8pt,line width=0.8pt] coordinates {(3.13,24.88)(3.77,25.48)(6.07,26.96)(7.53,27.28)(9.56,27.62)}; 

\addplot[color=color2,mark=diamond*, mark size=1.8pt,line width=0.8pt] coordinates {(2.41,21.35)(3.63,24.17)(4.71,25.21)(5.88,26.0)(7.02,26.59)(9.05,27.18)}; 

\addplot[color=color1,mark=*, mark size=1.4pt,line width=0.8pt] coordinates {(2.78,12.36)(3.8,15.73)(5.71,20.38)(6.59,22.75)(8.22,23.92)(9.65,24.08)}; 

\end{axis}
\end{tikzpicture}}
}

\caption{SacreBLEU against AL on the WMT22 X$\rightarrow$En and En$\rightarrow$X test sets. }
\label{fig:result_wmt22_bleu}
\end{figure*}

\noindent \textbf{System Settings}
In this paper, we conduct comparative experiments between EAST and the following baselines. We employ the same training setup as EAST in these baselines.

\begin{itemize}

\item \textbf{EAST}: The proposed pipeline includes two-stage training, \emph{i.e.}, full-weight fine-tuning on SiMT-De-En-660K followed by LoRA fine-tuning on SiMT-Multi-90K and Off-Multi-120K datasets.

\item \textbf{EAST-Stage-I}: Full-weight fine-tuning on the SiMT-De-En-660K dataset. 

\item \textbf{Conversational SiMT}~\cite{wang2024conversational}: It first generates SiMT data by segmenting parallel sentences using an alignment tool, and then formats it into a multi-round dialogue prompts for SFT. During inference, it reads $k$ tokens each step and then incrementally decodes them.

\item \textbf{Llama3-MOMT}: LoRA fine-tuning on the Off-Multi-120K dataset for OMT using Llama-3-8B-Instruct as the base model.

\item \textbf{Llama3-MOMT w/ wait-$k$}: Applying the wait-$k$ policy on the trained \textbf{Llama3-MOMT} model for streaming inference.

\end{itemize}

\subsection{Main Results}
\label{sec:main_results}

\noindent \textbf{SiMT WMT15 De$\rightarrow$En}
To compare with the previous SOTA models, We first evaluate our method on the WMT15  De$\rightarrow$En test set.
As Figure~\ref{fig:result_wmt15_bleu} shows, 
we compare our method with two categories of baselines: (1) Traditional SiMT methods, including ITST \citep{zhang-feng-2022-information}, Mono-KD \citep{wang-etal-2023-better}, and SM$^2$-Bi \citep{yu-etal-2024-self}; (2) LLM-based SiMT methods, including Agent-SiMT \citep{guo2024agent} and Conversational SimulMT \cite{wang2024conversational}.
Our EAST-Stage-I achieves superior BLEU-AL curves, outperforming these traditional approaches with a large margin. 
We admit that the LLMs are typically pre-trained on extensive multilingual corpora, giving them an inherent advantage over smaller SiMT models. 
However, it is important to recognize that our definition of auto-regressive SiMT with an adaptive policy represents a completely novel challenge for LLMs, and the size of our SFT dataset considerably smaller than that of these methods. 
In addition, even the recent LLM-based SiMT method Conversational SiMT and Agent-SiMT can only achieve on-par performance with traditional SiMT methods and does not show significant advantages.

When Conversational SiMT is trained on our SiMT-De-En-660K dataset, it achieves more than a 2 BLEU improvement across all latency settings compared to its original counterpart with a larger dataset (4M examples). This demonstrates the effectiveness of our dataset. When incorporating our adaptive policy (EAST-Stage-I), we observe an additional 0.9 BLEU improvement, showing the effectiveness of our policy.  
Upgrading the backbone model from Llama2 to Llama3 results in +0.5 BLEU in low-latency regions and +1 BLEU in high-latency regions.

\begin{table*}[t]
\centering
\small
\begin{tabular}{lccccc}
\toprule
Method & Low & Low-Medium & Medium & Medium-High & High \\
\midrule
Conversational-SiMT & 47.85 / 2.18 & 47.29 / 3.62 & 45.68 / 4.72 & 43.04 / 5.93 & 32.70 / 10.75 \\
EAST & 38.98 / 2.64 & 38.60 / 2.95 & 35.25 / 4.68 & 33.88 / 6.46 & 32.12 / 10.22 \\
\bottomrule
\end{tabular}
\caption{PPL and AL results on the WMT22 En$\rightarrow$Zh test set at different latency settings.}
\label{tab:ppl_al}
\end{table*}

\begin{table*}[t]
\centering
\resizebox{0.95\textwidth}{!}{
\begin{tabular}{lcccccccccc}
\toprule
\multirow{2}{*}{Models} & \multicolumn{2}{c}{De$\rightarrow$En} & \multicolumn{2}{c}{Zh$\rightarrow$En} & \multicolumn{2}{c}{Ru$\rightarrow$En} & \multicolumn{2}{c}{Cs$\rightarrow$En} & \multicolumn{2}{c}{Average} \\
\cmidrule(lr){2-3} \cmidrule(lr){4-5} \cmidrule(lr){6-7} \cmidrule(lr){8-9} \cmidrule(lr){10-11}
 & BLEU & COMET & BLEU & COMET & BLEU & COMET & BLEU & COMET & BLEU & COMET \\
\midrule 
GPT-4 & 33.87 & 85.62 & 27.20 & 82.79 & 43.51 & 86.18 & 48.67 & 87.43 & 38.31 & 85.51 \\ 
\hdashline
Bayling-7B & 28.16 & 83.19 & 20.31 & 77.48 & 34.74 & 82.48 & 35.98 & 82.03 & 29.80 & 81.30 \\
ALMA-7B-LoRA & 29.56 & 83.95 & 23.64 & 79.78 & 39.21 & 84.84 & 43.49 & 85.93 & 33.98 & 83.63 \\
\midrule
Llama3-MOMT & 31.98 & \textbf{84.89} & \textbf{25.48} & \textbf{81.26} & \textbf{39.83} & \textbf{85.19} & 44.92 & \textbf{86.23} & \textbf{35.55} & \textbf{84.39} \\
EAST & \textbf{32.55} & 84.77 & 23.80 & 80.86 & \textbf{39.83} & 85.04 & \textbf{45.61} & 86.20 & 35.45 & 84.22 \\
\bottomrule
\toprule
\multirow{2}{*}{Models} & \multicolumn{2}{c}{En$\rightarrow$De} & \multicolumn{2}{c}{En$\rightarrow$Zh} & \multicolumn{2}{c}{En$\rightarrow$Ru} & \multicolumn{2}{c}{En$\rightarrow$Cs} & \multicolumn{2}{c}{Average} \\ \cmidrule(lr){2-3} \cmidrule(lr){4-5} \cmidrule(lr){6-7} \cmidrule(lr){8-9} \cmidrule(lr){10-11}
 & BLEU & COMET & BLEU & COMET & BLEU & COMET & BLEU & COMET & BLEU & COMET \\
\midrule
GPT-4 & 35.38 & 87.44 & 43.98 & 87.49 & 30.45 & 88.87 & 34.53 & 90.77 & 36.09 & 88.64 \\
\hdashline
Bayling-7B & 25.66 & 82.18 & 38.19 & 84.43 & 14.85 & 74.72 & 15.64 & 76.85 & 23.59 & 79.55 \\
ALMA-7B-LoRA & 30.16 & 85.45 & 36.47 & 84.87 & \textbf{26.93} & 87.05 & \textbf{30.17} & \textbf{89.05} & 30.93 & 86.61 \\
\midrule 
Llama3-MOMT & 30.45 & \textbf{85.63} & \textbf{40.68} & \textbf{86.53} & 24.83 & \textbf{87.27} & 27.92 & 88.36 & 30.97 & \textbf{86.95} \\
EAST & \textbf{30.84} & 85.49 & 40.17 & 86.31 & 26.79 & 87.13 & 26.63 & 88.17 & \textbf{31.11} & 86.78 \\
\bottomrule
\end{tabular}
}
\caption{Offline results on the WMT22 X$\rightarrow$En and En$\rightarrow$X test sets.}
\label{tab:offline_mt}
\end{table*}

\noindent \textbf{SiMT WMT22 X$\leftrightarrow$En} 
We further evaluate EAST on the WMT22 X$\leftrightarrow$En test sets, as shown in Figure \ref{fig:result_wmt22_bleu}.
We compare EAST with two LLM-based baselines: Llama-MOMT w/ wait-$k$ and Conversation SimulMT, where Conversation SimulMT is reproduced based on the Llama3 backbone model with EAST's SFT data and the two-stage training method for fair comparisons.
Llama-MOMT w/ wait-$k$ shows inferior translation quality in all latency areas due to the limitations of the fixed policy.
Compared to Conversation SimulMT, EAST achieves higher BLEU in all latency regions across eight directions, with an average improvement of 1.5 BLEU in low latency regions.
In the En$\rightarrow$Zh direction, while Conversational SimulMT achieves a higher BLEU at high latency area, it is lower than EAST in COMET and BLEURT metrics, as shown in Figures \ref{fig:result_wmt22_comet} and \ref{fig:result_wmt22_brt}.
These metrics better reflect semantic quality that better correlate with human judgments, whereas BLEU primarily measures surface-level n-gram overlap with the reference.
This result can be attributed to the fixed policy used in Conversational SimulMT. It often generates translations before the complete semantic context is available, which can lead to surface-level matches with the reference (thus increasing BLEU), but at the cost of semantic completeness or fluency—resulting in lower COMET and BLEURT scores.
To further verify this, we use \texttt{Llama-3-8B} to compute the PPL of translations.
As shown in Table \ref{tab:ppl_al}, EAST achieves lower PPL scores across all latency levels, demonstrating that its translations are more fluent and semantically coherent.
A detailed comparison of the COMET and BLEURT metrics is provided in the Appendix \ref{apd:comet_bleurt}.

\noindent \textbf{Offline Performance} 
We also evaluate the performance of offline translation on the WMT22 test set, as presented in Table \ref{tab:offline_mt}. 
Our results are superior to previous studies, Bayling \citep{zhang2023bayling} and ALMA \citep{xu2024a}, except for a slight lag in En$\rightarrow$Cs.
Compared to OMT model Llama3-MOMT and other variants, EAST maintains comparable or superior offline translation performance across the eight language directions, indicating that our two-stage SFT process effectively maintains translation quality for OMT. 

In summary, these results highlight that EAST not only excels in high-quality simultaneous translation but also ensures that the offline translation capabilities are not compromised.

\begin{figure}[t]
\pgfplotsset{
  width=7.5cm,height=6.8cm,
    every axis y label/.append style={at={(-0.09,0.5)}},
    every axis/.append style={line width=1.0pt},
}
\centering
\resizebox{0.248\textwidth}{!}{
\subfigure[De$\rightarrow$En]{
\begin{tikzpicture}[baseline]
\begin{axis}[
    ylabel=SacreBLEU,
    xlabel=LAAL,
    enlargelimits=0.05,
    legend cell align=left,
    legend style={font=\tiny,
    at={(0.72,0.01)},
    anchor=south,
    legend columns=1},
    xmajorgrids=true,
    ymajorgrids=true,
    grid style=dashed,
    ymin=24,ymax=35,
    xmin=2.0, 
    xmax=17,
]
\addplot[color=color4, dashed, mark=none, mark size=1.8pt,line width=0.8pt] coordinates {(0, 33.37)(18, 33.37)};

\addplot[color=color2, dashed, mark=none, mark size=1.8pt,line width=0.8pt] coordinates {(0, 32.55)(18, 32.55)}; 

\addplot[color=color4,mark=triangle*, mark size=1.8pt,line width=0.8pt] coordinates {(4.24,32.76)(4.39,33.09)(5.35,34.26)(5.73,34.76)(8.29,35.21)}; 

\addplot[color=color2,mark=triangle*, mark size=1.8pt,line width=0.8pt] coordinates {(2.89,29.87)(3.02,30.09)(3.64,31.09)(4.02,31.61)(6.01,32.38)}; 

\addplot[color=color1, dashed, mark=none, mark size=1.4pt,line width=0.8pt] coordinates {(0, 33.14)(18, 33.14)};

\addplot[color=color5, dashed, mark=none, mark size=1.4pt,line width=0.8pt] coordinates {(0, 31.98)(18, 31.98)}; 

\addplot[color=color1, mark=*, mark size=1.4pt,line width=0.8pt] coordinates {(9.21,31.18)(10.26,31.59)(11.02,31.93)(11.96,32.24)(12.96,32.51)(14.8,32.61)(16.45,33.02)};

\addplot[color=color5, mark=*, mark size=1.4pt,line width=0.8pt] coordinates {(2.06,24.55)(2.94,26.5)(3.82,27.6)(4.71,28.48)(6.4,29.48)(8.59,29.71)(11.65,30.36)(13.08,30.35)}; 

\end{axis}
\end{tikzpicture}}
}
\hspace{-3mm}
\resizebox{0.234\textwidth}{!}{
\subfigure[Ru$\rightarrow$En]{
\begin{tikzpicture}[baseline]
\begin{axis}[
    xlabel=LAAL,
    enlargelimits=0.05,
    legend style={font=\tiny,
    at={(0.72,0.01)},
    anchor=south,
    legend columns=1},
    xmajorgrids=true,
    ymajorgrids=true,
    grid style=dashed,
    xmin=2.5,
    xmax = 31.0,
]

\addplot[color=color4, dashed, mark=none, mark size=1.8pt,line width=0.8pt] coordinates {(0, 40.04)(35, 40.04)}; 

\addplot[color=color2, dashed, mark=none, mark size=1.8pt,line width=0.8pt] coordinates {(0, 39.80)(35, 39.80)}; 

\addplot[color=color4,mark=triangle*, mark size=1.8pt,line width=0.8pt] coordinates {(4.17,40.49)(4.49,41.37)(6.15,41.9)(7.15,42.28)(9.54,42.72)}; 

\addplot[color=color2,mark=triangle*, mark size=1.8pt,line width=0.8pt] coordinates {(2.77,36.35)(2.95,36.49)(3.85,37.42)(4.41,37.73)(7.05,38.92)}; 

\addplot[color=color1, dashed, mark=none, mark size=1.8pt,line width=0.8pt] coordinates {(0,42.28)(35,42.28)}; 

\addplot[color=color5, dashed, mark=none, mark size=1.8pt,line width=0.8pt] coordinates {(0,39.84)(35,39.84)}; 

\addplot[color=color1,mark=*, mark size=1.4pt,line width=0.8pt] coordinates {(19.91,40.14)(21.17,41.49)(22.92,41.66)(25.26,41.95)(27.96,42.26)}; 

\addplot[color=color5,mark=*, mark size=1.4pt,line width=0.8pt] coordinates {(4.12,37.05)(5.78,37.26)(6.58,38.48)(8.64,38.61)(10.25,38.71)}; 

\legend{EAST Doc-Off, EAST Sent-Off, EAST Doc-Si, EAST Sent-Si, Llama3-MOMT Doc-Off, Llama3-MOMT Sent-Off, Llama3-MOMT Doc-Si, Llama3-MOMT Sent-Si}

\end{axis}
\end{tikzpicture}}
}

\caption{SacreBLEU-LAAL curves on the WMT22 document-level De/Ru$\rightarrow$En test set. Methods labeled with ``-Off" refer to offline translation, \emph{i.e.}, including the entire document in the prompt. Methods marked with ``-Si" denote simultaneous translation, involving the streaming input.}
\label{fig:result_wmt22_x2en_doc}

\end{figure}
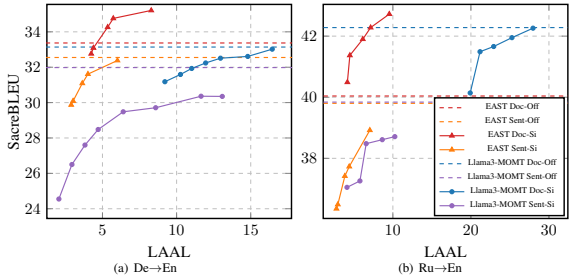

\begin{table*}[t]
\centering
\small
\begin{tabular}{lcccc}
\toprule
Method & BLEU ($\uparrow$) & AL ($\downarrow$) & AL-CA ($\downarrow$) &  WWT (ms) ($\downarrow$) \\
\cmidrule(lr){1-5}
EAST-offline & 32.55 & 14.62 & 15.22 & 38.96 \\ 
EAST & 29.87/31.08/32.38 & 2.59/3.42/5.87 & 3.26/4.29/6.55 & 49.87 ($\pm$1,21) \\
Llama3-MOMT w/ wait-k & 26.50/27.60/28.95 & 2.70/3.63/5.44 & 3.69/4.69/6.43 & 977.2 ($\pm$4.49) \\ 
\bottomrule
\end{tabular}
\caption{Comparison of inference latency and speed on WMT22 De$\rightarrow$En test set.  The BLEU, AL, and AL-CA scores are given for low, medium, and high latency settings respectively. WWT refers to the actual inference time per word and is reported as mean and standard deviations (in parentheses) over the three latency.}
\label{tab:infer_latency}
\end{table*}

\subsection{Zero-Shot Generalization to Document-level SiMT}
Since real-world applications often involve streaming inputs that are typically long and unsegmented, we further evaluate the EAST directly on the document-level test set from WMT22 De/Ru$\rightarrow$En without fine-tuning on document-level data. 
In our experiments, the document-level test set is derived from the same data as the sentence-level set but without sentence segmentation.
The results \emph{w.r.t.} the corpus BLEU are depicted in Figure~\ref{fig:result_wmt22_x2en_doc}. 
EAST shows superior performance in document-level settings, as this enhancement is due to the model's improved ability to leverage historical context, thus enhancing translation accuracy and coherence.

Originally, document-level offline translation was expected to be one of the strongest capabilities of LLMs. 
Surprisingly, our proposed EAST model significantly outperforms both EAST and Llama3-MOMT in offline performance within a document-level context. 
This discrepancy in previous works may arise from models being trained exclusively on sentence-level data, which can lead to a training-inference mismatch during offline translation. 
Additionally, the long context of the source document may contribute to forgetting issues during the generation of the target document. 
However, our training approach, which alternates between source and target texts, effectively minimizes these mismatches. 
These results indicate that EAST is better suited for longer text sequences, making it particularly suitable for streaming scenarios.

Moreover, it can be observed that there is a significant rightward shift on the document-level BLEU-LAAL curve of the wait-$k$ method (Llama3-MOMT Doc-Si) compared to its sentence-level counterpart (Llama3-MOMT Sent-Si).
Our statistical data indicates that English texts are substantially longer than their German and Russian counterparts in document-level test set—averaging 16.1 words longer than German and 35.8 words longer than Russian. 
This discrepancy is much greater than in the sentence-level test set, where English texts are only 2.2 and 3.1 words longer, respectively.
Instead, EAST uses an adaptive read/write policy that effectively mitigates the above problems.

\subsection{Inference as Efficient as Offline}

In this section, we measure the overall efficiency of the EAST model on an NVIDIA A100 through 
computation-aware latency (AL-CA) and 
decoding speed (WWT), as shown in Table \ref{tab:infer_latency}. 
EAST achieves comparable translation performance to its offline counterpart EAST-offline with lower latency, and significantly outperforms Llama3-MOMT w/ wait-$k$ method under similar latency conditions.
For decoding speed, Llama3-MOMT w/ wait-$k$ shows the slowest inference speeds, taking up to 977.2ms to generate a single word. 
This inefficiency is attributed to the inability of this method to efficiently utilize the KV cache, necessitating the re-encoding of historical content at each decoding step, which limits its practical use in real-time scenarios.
Conversely, EAST efficiently leverages KV-cache during inference, taking only 49ms to decode a word, achieving comparable decoding speeds to its offline counterpart (38.96ms per word).
This small difference of about 10ms shows that EAST maintains near-offline efficiency, even under the streaming input conditions of SiMT.

\begin{figure*}[t]
\pgfplotsset{width=4.7cm,height=4.5cm,
    every axis/.append style={line width=0.6pt},
}
\centering
\subfigure[Low Latency]{
\begin{tikzpicture}[baseline]
\begin{axis}[
    xlabel={Data Size (k)},
    ylabel={BLEU},
    axis y line*=left,
    enlargelimits=0.03,
    xmode=log,
    font=\scriptsize,
    ymax=32,
    ymin=9,
    xmin=1, xmax=660,
    legend style={font=\tiny, at={(0.98,0.65)}},
    grid style=dashed,
    xmajorgrids=true,
    ymajorgrids=true,
    ylabel style={yshift=-4pt},
]

\addplot[
    color=color1,
    mark=none,
    ]
    coordinates {
    (1.28,9.08)(2.56,9.35)(5.12,13.98)(6.4,12.02)(7.68,14.41)(8.96,26.37)(10.24,28.95)(12.8,27.28)(15.36,28.57)(20.48,27.58)(25.6,28.56)(30.72,28.14)(40.96,28.1)(51.2,28.84)(61.44,29.38)(71.68,29.26)(81.92,28.03)(92.16,28.27)(102.4,27.75)(122.88,27.74)(153.6,27.51)(163.84,27.62)(184.32,27.73)(204.8,27.72)(225.28,28.15)(256.0,28.3)(276.48,28.8)(307.2,28.78)(327.68,28.39)(358.4,28.3)(389.12,28.41)(409.6,28.61)(440.32,28.96)(460.8,28.48)(491.52,28.58)(512.0,28.68)(552.96,28.62)(563.2,28.72)(614.4,28.7)(659.968,28.73)
    };

\addlegendentry{BLEU}
\end{axis}

\begin{axis}[
    axis y line*=right,
    axis x line=none,
    font=\scriptsize,
    legend style={font=\tiny, at={(0.895, 0.45)}},
    ylabel={AL},
    xmin=1, xmax=660,
    enlargelimits=0.03,
    ylabel near ticks,
    yticklabel pos=right,
    xmode=log,
    ylabel style={yshift=4pt},
]

\addplot[
    color=color2,
    mark=none,
    ]
    coordinates {
    (1.28,14.62)(2.56,14.62)(5.12,14.62)(6.4,12.94)(8.96,5.78)(10.24,5.26)(12.8,3.16)(15.36,3.71)(20.48,2.99)(25.6,3.52)(30.72,3.0)(40.96,3.35)(51.2,3.45)(61.44,3.97)(71.68,4.48)(81.92,3.43)(92.16,3.07)(102.4,2.85)(122.88,2.81)(153.6,2.71)(163.84,2.89)(184.32,2.88)(204.8,2.8)(225.28,3.06)(256.0,2.97)(276.48,3.1)(307.2,3.01)(327.68,2.95)(358.4,2.94)(389.12,2.93)(409.6,3.05)(440.32,3.23)(460.8,3.1)(491.52,2.93)(512.0,3.11)(552.96,2.98)(563.2,2.99)(614.4,3.01)(659.968,3.03)
    }; 

\addlegendentry{AL}
\end{axis}
\end{tikzpicture}}
~
\subfigure[Medium Latency]{
\begin{tikzpicture}[baseline]
\begin{axis}[
    xlabel={Data Size (k)},
    ylabel={BLEU},
    axis y line*=left,
    enlargelimits=0.03,
    font=\scriptsize,
    xmode=log,
    xmin=1, xmax=660,
    legend style={font=\tiny, at={(0.58,0.98)}},
    grid style=dashed,
    xmajorgrids=true,
    ymajorgrids=true,
    ylabel style={yshift=-4pt},
]

\addplot[
    color=color1,
    mark=none,
    ]
    coordinates {
    (1.28,8.96)(2.56,9.11)(5.12,12.19)(6.4,12.47)(7.68,14.22)(8.96,26.04)(10.24,29.14)(12.8,28.23)(15.36,30.36)(20.48,29.44)(25.6,30.2)(30.72,29.71)(40.96,29.83)(51.2,30.04)(71.68,30.08)(92.16,29.46)(102.4,29.41)(153.6,28.96)(184.32,29.54)(204.8,29.75)(225.28,29.84)(256.0,29.93)(276.48,30.2)(307.2,29.94)(327.68,29.74)(389.12,29.66)(409.6,29.87)(440.32,30.07)(491.52,29.94)(512.0,29.68)(552.96,29.86)(614.4,29.89)(659.968,29.89)
    };

\end{axis}

\begin{axis}[
    axis y line*=right,
    axis x line=none,
    font=\scriptsize,
    legend style={font=\tiny, at={(1.0, 0.98)}},
    ylabel={AL},
    xmode=log,
    xmin=1, xmax=660,
    enlargelimits=0.03,
    ylabel near ticks,
    yticklabel pos=right,
    ylabel style={yshift=4pt},
]

\addplot[
    color=color2,
    mark=none,
    ]
    coordinates {
    (1.28,14.62)(2.56,14.62)(5.12,14.62)(6.4,12.75)(7.68,-0.42)(8.96,5.71)(10.24,5.38)(12.8,3.98)(15.36,6.14)(20.48,4.54)(25.6,6.1)(30.72,4.77)(40.96,5.4)(51.2,5.78)(71.68,7.2)(92.16,5.38)(102.4,5.09)(153.6,4.24)(184.32,4.81)(204.8,4.7)(225.28,5.69)(256.0,5.16)(276.48,5.54)(307.2,5.32)(327.68,5.39)(389.12,5.2)(409.6,5.14)(440.32,5.71)(491.52,5.16)(512.0,5.18)(552.96,5.15)(614.4,5.22)(659.968,5.28)
    }; 

\end{axis}
\end{tikzpicture}}
~
\subfigure[High Latency]{
\begin{tikzpicture}[baseline]
\begin{axis}[
    xlabel={Data Size (k)},
    ylabel={BLEU},
    axis y line*=left,
    enlargelimits=0.03,
    xmode=log,
    xmin=1, xmax=660,
    font=\scriptsize,
    legend style={font=\tiny, at={(0.57,0.98)}},
    grid style=dashed,
    xmajorgrids=true,
    ymajorgrids=true,
    ylabel style={yshift=-4pt},
]

\addplot[
    color=color1,
    mark=none,
    ]
    coordinates {
    (1.28,6.94)(2.56,6.74)(5.12,7.31)(6.4,10.72)(7.68,15.43)(8.96,26.03)(10.24,29.19)(12.8,29.04)(15.36,30.51)(20.48,30.44)(25.6,30.51)(30.72,30.49)(40.96,30.6)(51.2,30.32)(71.68,30.04)(92.16,29.84)(102.4,30.06)(153.6,30.0)(184.32,30.29)(204.8,30.29)(225.28,30.6)(256.0,30.3)(276.48,30.56)(307.2,30.46)(327.68,30.47)(389.12,30.31)(409.6,30.03)(440.32,30.42)(491.52,30.36)(512.0,30.14)(552.96,30.22)(614.4,30.21)(659.968,30.19)
    };

\end{axis}

\begin{axis}[
    scaled ticks=false,
    axis y line*=right,
    axis x line=none,
    font=\scriptsize,
    legend style={font=\tiny, at={(0.9, 0.5)}},
    ylabel={AL},
    xmode=log,
    xmin=1, xmax=660,
    enlargelimits=0.03,
    ylabel near ticks,
    yticklabel pos=right,
    xmin=1.0, xmax=660,
    ylabel style={yshift=4pt},
]

\addplot[
    color=color2,
    mark=none,
    ]
    coordinates {
    (1.28,14.61)(2.56,14.62)(5.12,14.62)(6.4,13.38)(7.68,0.15)(8.96,5.92)(10.24,5.53)(12.8,4.91)(15.36,7.72)(20.48,7.16)(25.6,8.62)(30.72,8.05)(40.96,8.48)(51.2,8.39)(71.68,9.7)(92.16,8.18)(102.4,8.05)(153.6,7.85)(184.32,8.01)(204.8,7.62)(225.28,8.51)(256.0,7.59)(276.48,8.21)(307.2,8.27)(327.68,8.28)(389.12,8.21)(409.6,7.84)(440.32,8.46)(491.52,7.91)(512.0,8.1)(552.96,8.22)(614.4,8.17)(659.968,8.2)
    }; 

\end{axis}
\end{tikzpicture}}

\caption{BLEU scores (left $y$-axis ) and AL values (right $y$-axis ) over data size. We use log scale for scale the x-axis to more clearly observe the effect of data size. The original plots are also provided in Figure \ref{fig:data_scale_ori}.}
\label{fig:data_scale}
\end{figure*}
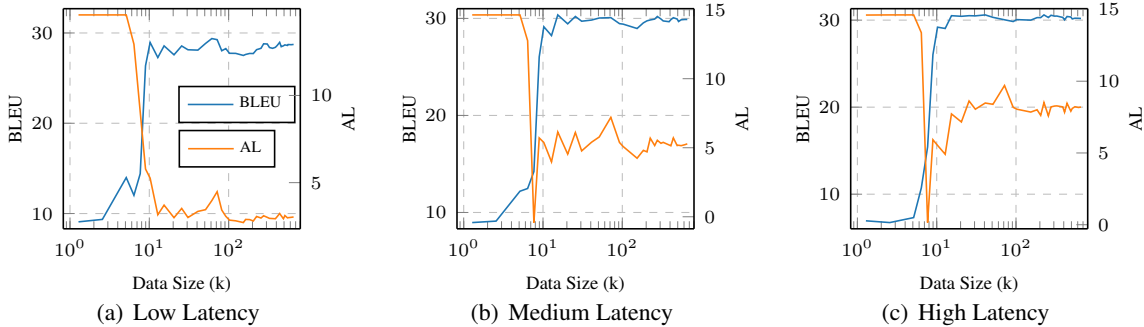

\subsection{How Many Examples Are Needed to Teach LLMs the Novel SiMT Task?} 
In this section, we investigate the data size required for efficiently training LLMs on the novel SiMT task based on the SiMT-De-En-660K dataset.
Figure \ref{fig:data_scale} illustrates the changes in BLEU and AL scores with data sizes for different latency settings--``low", ``medium", and ``high". 
First, there is a significant improvement in BLEU score as the data size increases to about 10K. 
Beyond this point, the rate of increase in BLEU score diminishes. 
Similarly, AL metrics also decrease notably as the data size reaches around 10k, before stabilizing or showing minor fluctuations. 
This pattern is consistent across all latency settings, indicating a general learning behavior of the model: rapid enhancements in translation quality are achieved with the first 10K examples, followed by a phase where the model  focuses on refining its read/write policy across different latency levels, up to 100k examples.

The results suggest that a relatively small dataset of just 10K SiMT examples may be sufficient to achieve commendable translation quality. 
This finding aligns well with our multilingual dataset, which contains approximately 10K examples per language, facilitating good performance across different languages. 
Moreover, expanding the data size (up to 100K examples) can further optimize the model's read/write policy.

\textbf{More analysis are included in the Appendix.}
Appendix \ref{apd:diff_train} analyzes the performance of EAST under various training strategies.
Appendix \ref{apd:hallu_rate} compares the hallucination rate of different methods.
Appendix \ref{apd:qua_policy} evaluates the quality of different translation policies.
Appendix \ref{apd:diff_backbone} explores the impact of different backbones on translation performance.
Appendix \ref{apd:gpt4} compares the SiMT performance of EAST with GPT-4.
Appendix \ref{apd:fluency} evaluates the fluency of EAST translations.

\section{Conclusion}
\label{sec:conclusion}
In this paper, we introduce an \textbf{E}fficient and \textbf{A}daptive \textbf{S}imultaneous \textbf{T}ranslation method using LLMs, EAST, designed to achieve high-quality SiMT with the efficiency of offline systems. 
By constructing SFT data, leveraging an interleaved token structure with explicit read-write signals and incorporating latency-aware prompts, EAST enables LLMs to perform adaptive reading and translation based on varying latency requirements.
Our experimental results demonstrate that EAST not only achieves state-of-the-art performance on SiMT benchmarks but also maintains high-quality translations in offline settings.
Additionally, EAST shows excellent generalization to document-level SiMT, highlighting its suitability for streaming translation in real-world scenarios.

\section*{Limitations}
The proposed method assumes an idealized setting where the input is clean and fluent. In real-world applications, however, simultaneous translation often involves noisy or disfluent input. The model's performance under such conditions has not been evaluated.
Additionally, our approach is designed and evaluated for simultaneous text-to-text translation tasks, leaving the domain of simultaneous speech-to-text translation unexplored.

\section*{Acknowledgements}
We would like to thank all the anonymous reviewers for the insightful and helpful comments.
This work is supported by the National Science and Technology Major Project (Grant No. 2022ZD0116101),
the Major Scientific Research Project of the State Language Commission in the 13th Five-Year Plan (Grant No. WT135-38), the public technology service platform project of Xiamen City (No. 3502Z20231043), and Alibaba Research Intern Program.

\bibliography{anthology,custom}

\appendix
\appendix

\begin{figure*}[t]
\pgfplotsset{width=5.2cm,height=4.5cm,
    every axis/.append style={line width=0.6pt},
}
\centering
\hspace{-3mm}
\subfigure[Low Latency]{
\begin{tikzpicture}[baseline]
\begin{axis}[
    xlabel={Data Size (k)},
    ylabel={BLEU},
    axis y line*=left,
    enlargelimits=0.03,
    font=\scriptsize,
    ymax=32,
    ymin=9,
    xmin=0, xmax=660,
    xtick={0,100,200,400,600},
    legend style={font=\tiny, at={(0.98,0.65)}},
    grid style=dashed,
    xmajorgrids=true,
    ymajorgrids=true,
    ylabel style={yshift=-4pt},
]

\addplot[
    color=color1,
    mark=none,
    ]
    coordinates {
    (1.28,9.08)(2.56,9.35)(5.12,13.98)(6.4,12.02)(7.68,14.41)(8.96,26.37)(10.24,28.95)(12.8,27.28)(15.36,28.57)(20.48,27.58)(25.6,28.56)(30.72,28.14)(40.96,28.1)(51.2,28.84)(61.44,29.38)(71.68,29.26)(81.92,28.03)(92.16,28.27)(102.4,27.75)(122.88,27.74)(153.6,27.51)(163.84,27.62)(184.32,27.73)(204.8,27.72)(225.28,28.15)(256.0,28.3)(276.48,28.8)(307.2,28.78)(327.68,28.39)(358.4,28.3)(389.12,28.41)(409.6,28.61)(440.32,28.96)(460.8,28.48)(491.52,28.58)(512.0,28.68)(552.96,28.62)(563.2,28.72)(614.4,28.7)(659.968,28.73)
    };

\addlegendentry{BLEU}
\end{axis}

\begin{axis}[
    axis y line*=right,
    axis x line=none,
    font=\scriptsize,
    legend style={font=\tiny, at={(0.895, 0.45)}},
    ylabel={AL},
    xmin=0, xmax=660,
    enlargelimits=0.03,
    ylabel near ticks,
    yticklabel pos=right,
    ylabel style={yshift=4pt},
]

\addplot[
    color=color2,
    mark=none,
    ]
    coordinates {
    (1.28,14.62)(2.56,14.62)(5.12,14.62)(6.4,12.94)(8.96,5.78)(10.24,5.26)(12.8,3.16)(15.36,3.71)(20.48,2.99)(25.6,3.52)(30.72,3.0)(40.96,3.35)(51.2,3.45)(61.44,3.97)(71.68,4.48)(81.92,3.43)(92.16,3.07)(102.4,2.85)(122.88,2.81)(153.6,2.71)(163.84,2.89)(184.32,2.88)(204.8,2.8)(225.28,3.06)(256.0,2.97)(276.48,3.1)(307.2,3.01)(327.68,2.95)(358.4,2.94)(389.12,2.93)(409.6,3.05)(440.32,3.23)(460.8,3.1)(491.52,2.93)(512.0,3.11)(552.96,2.98)(563.2,2.99)(614.4,3.01)(659.968,3.03)
    }; 

\addlegendentry{AL}
\end{axis}
\end{tikzpicture}}
\hspace{-3mm}
\subfigure[Medium Latency]{
\begin{tikzpicture}[baseline]
\begin{axis}[
    xlabel={Data Size (k)},
    ylabel={BLEU},
    axis y line*=left,
    enlargelimits=0.03,
    font=\scriptsize,
    xmin=0, xmax=660,
    xtick={0,100,200,400,600},
    legend style={font=\tiny, at={(0.58,0.98)}},
    grid style=dashed,
    xmajorgrids=true,
    ymajorgrids=true,
    ylabel style={yshift=-4pt},
]

\addplot[
    color=color1,
    mark=none,
    ]
    coordinates {
    (1.28,8.96)(2.56,9.11)(5.12,12.19)(6.4,12.47)(7.68,14.22)(8.96,26.04)(10.24,29.14)(12.8,28.23)(15.36,30.36)(20.48,29.44)(25.6,30.2)(30.72,29.71)(40.96,29.83)(51.2,30.04)(71.68,30.08)(92.16,29.46)(102.4,29.41)(153.6,28.96)(184.32,29.54)(204.8,29.75)(225.28,29.84)(256.0,29.93)(276.48,30.2)(307.2,29.94)(327.68,29.74)(389.12,29.66)(409.6,29.87)(440.32,30.07)(491.52,29.94)(512.0,29.68)(552.96,29.86)(614.4,29.89)(659.968,29.89)
    };

\end{axis}

\begin{axis}[
    axis y line*=right,
    axis x line=none,
    font=\scriptsize,
    legend style={font=\tiny, at={(1.0, 0.98)}},
    ylabel={AL},
    xmin=0, xmax=660,
    enlargelimits=0.03,
    ylabel near ticks,
    yticklabel pos=right,
    ylabel style={yshift=4pt},
]

\addplot[
    color=color2,
    mark=none,
    ]
    coordinates {
    (1.28,14.62)(2.56,14.62)(5.12,14.62)(6.4,12.75)(7.68,-0.42)(8.96,5.71)(10.24,5.38)(12.8,3.98)(15.36,6.14)(20.48,4.54)(25.6,6.1)(30.72,4.77)(40.96,5.4)(51.2,5.78)(71.68,7.2)(92.16,5.38)(102.4,5.09)(153.6,4.24)(184.32,4.81)(204.8,4.7)(225.28,5.69)(256.0,5.16)(276.48,5.54)(307.2,5.32)(327.68,5.39)(389.12,5.2)(409.6,5.14)(440.32,5.71)(491.52,5.16)(512.0,5.18)(552.96,5.15)(614.4,5.22)(659.968,5.28)
    }; %

\end{axis}
\end{tikzpicture}}
\hspace{-3mm}
\subfigure[High Latency]{
\begin{tikzpicture}[baseline]
\begin{axis}[
    xlabel={Data Size (k)},
    ylabel={BLEU},
    axis y line*=left,
    enlargelimits=0.03,
    xmin=0, xmax=660,
    font=\scriptsize,
    xtick={0,100,200,400,600},
    legend style={font=\tiny, at={(0.57,0.98)}},
    grid style=dashed,
    xmajorgrids=true,
    ymajorgrids=true,
    ylabel style={yshift=-4pt},
]

\addplot[
    color=color1,
    mark=none,
    ]
    coordinates {
    (1.28,6.94)(2.56,6.74)(5.12,7.31)(6.4,10.72)(7.68,15.43)(8.96,26.03)(10.24,29.19)(12.8,29.04)(15.36,30.51)(20.48,30.44)(25.6,30.51)(30.72,30.49)(40.96,30.6)(51.2,30.32)(71.68,30.04)(92.16,29.84)(102.4,30.06)(153.6,30.0)(184.32,30.29)(204.8,30.29)(225.28,30.6)(256.0,30.3)(276.48,30.56)(307.2,30.46)(327.68,30.47)(389.12,30.31)(409.6,30.03)(440.32,30.42)(491.52,30.36)(512.0,30.14)(552.96,30.22)(614.4,30.21)(659.968,30.19)
    };

\end{axis}

\begin{axis}[
    scaled ticks=false,
    axis y line*=right,
    axis x line=none,
    font=\scriptsize,
    legend style={font=\tiny, at={(0.9, 0.5)}},
    ylabel={AL},
    xmin=0, xmax=660,
    enlargelimits=0.03,
    ylabel near ticks,
    yticklabel pos=right,
    xmin=1.0, xmax=660,
    ylabel style={yshift=4pt},
]

\addplot[
    color=color2,
    mark=none,
    ]
    coordinates {
    (1.28,14.61)(2.56,14.62)(5.12,14.62)(6.4,13.38)(7.68,0.15)(8.96,5.92)(10.24,5.53)(12.8,4.91)(15.36,7.72)(20.48,7.16)(25.6,8.62)(30.72,8.05)(40.96,8.48)(51.2,8.39)(71.68,9.7)(92.16,8.18)(102.4,8.05)(153.6,7.85)(184.32,8.01)(204.8,7.62)(225.28,8.51)(256.0,7.59)(276.48,8.21)(307.2,8.27)(327.68,8.28)(389.12,8.21)(409.6,7.84)(440.32,8.46)(491.52,7.91)(512.0,8.1)(552.96,8.22)(614.4,8.17)(659.968,8.2)
    }; %

\end{axis}
\end{tikzpicture}}

\caption{BLEU scores (left $y$-axis ) and AL values (right $y$-axis ) over data size using normal scale.}
\label{fig:data_scale_ori}
\end{figure*}
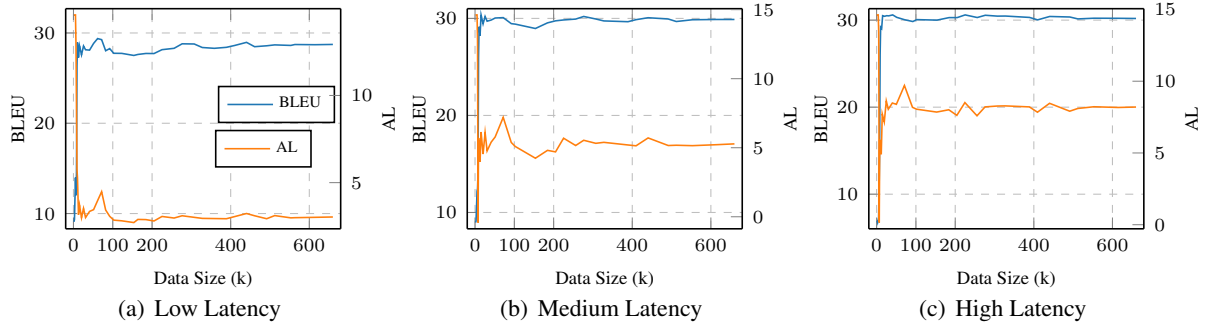

\begin{table*}[ht]
\centering
\resizebox{\textwidth}{!}{
\begin{tabular}{lcccccccccc}
\toprule
\multirow{2}{*}{Latency} & \multicolumn{9}{c}{\textbf{SiMT-Multi-90K}} & \multicolumn{1}{l}{\textbf{SiMT-De-En-660K}} \\
\cmidrule(lr){2-10} \cmidrule(lr){11-11}
 & De$\rightarrow$En & Zh$\rightarrow$En & Ru$\rightarrow$En & Cs$\rightarrow$En & En$\rightarrow$De & EN$\rightarrow$Zh & En$\rightarrow$Ru & En$\rightarrow$Cs & Total & De$\rightarrow$En \\ \cmidrule(lr){1-11}
Low & 3,325 & 1,635 & 3,642 & 2,507 & 3,267 & 2,423 & 4,945 & 4,226 & 25,970 & 230,902 \\
Medium & 3,631 & 2,763 & 3,719 & 2,472 & 3,997 & 2,433 & 5,830 & 5,035 & 29,880 & 227,131 \\
High & 4,102 & 4,166 & 4,254 & 2,746 & 4,921 & 2,920 & 6,322 & 5,433 & 34,864 & 202,843 \\ \cmidrule(lr){1-11}
Total & 11,058 & 8,564 & 11,615 & 7,725 & 12,185 & 7,776 & 17,097 & 14,694 & 90,714 & 660,876 \\
\bottomrule
\end{tabular}
}
\caption{The statistics for the two SiMT datasets we constructed.}
\label{tab:simt_data}
\end{table*}

\begin{table*}[ht]
\centering
\resizebox{0.9\textwidth}{!}{
\begin{tabular}{lcccccc}
\toprule
\multirow{2}{*}{Language} & \multicolumn{3}{c}{Sentence-level Parallel Data} & \multicolumn{3}{c}{Document-level   Parallel Data} \\
\cmidrule(lr){2-4} \cmidrule(lr){5-7}
 & Train & Test (from En) & Test (to En) & Test (to English) & Avg. Words & Max. Words \\
\cmidrule(lr){1-7}
German (De) & 14211 & 2037 & 1984 & 217 & 107/123 & 839/1003 \\
Chinese (Zh) & 15406 & 2037 & 1875 & - & - & - \\ 
Russia (Ru) & 15000 & 2037 & 2016 & 128 & 175/211 & 699/843 \\
Czech (Cs) & 12076 & 2037 & 1448 & - & - & - \\
\bottomrule
\end{tabular}
}
\caption{The statistics for the parallel data from the WMT. "Avg. Words" indicates the average number of words per document in the source/target language. "Max. Words" represents the maximum number of words per document in the source/target language.}
\label{tab:data_statistics}
\end{table*}

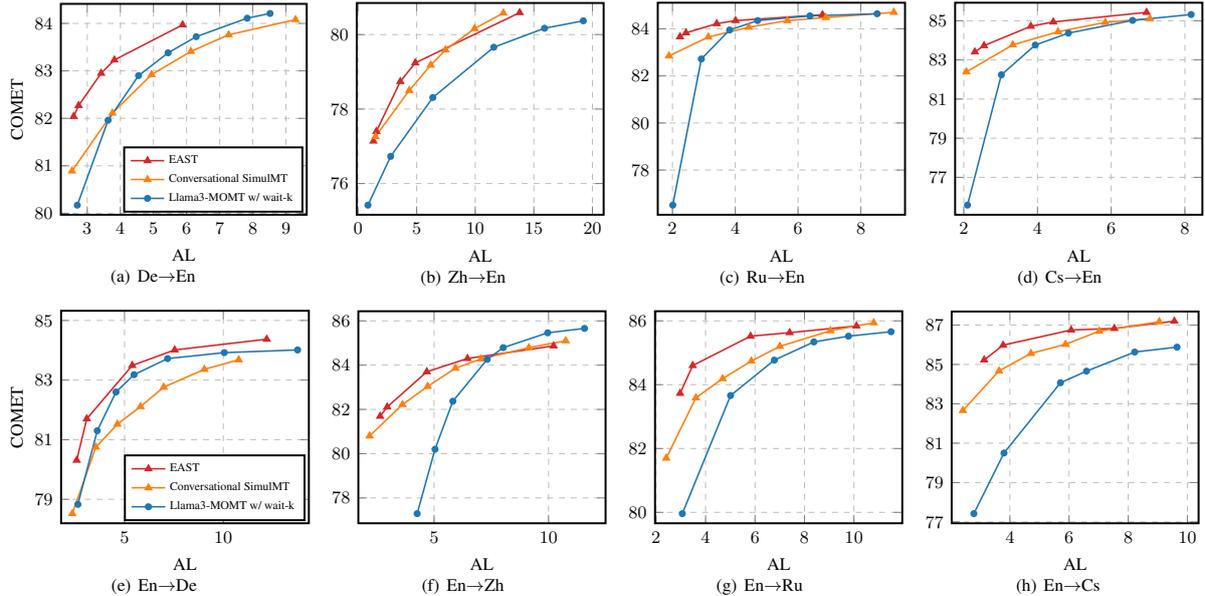
\begin{figure*}[t]
\pgfplotsset{
    every axis y label/.append style={at={(-0.12,0.5)}},
    every axis/.append style={line width=1.0pt},
}
\centering
\resizebox{0.263\textwidth}{!}{
\subfigure[De$\rightarrow$En]{
\begin{tikzpicture}[baseline]
\begin{axis}[
    ylabel=COMET,
    xlabel=AL,
    enlargelimits=0.05,
    font=\small,
    width=6cm,height=5.4cm,
    legend cell align=left,
    legend style={font=\tiny,
    at={(0.62,0.01)},
    anchor=south,
    legend columns=1},
    xmajorgrids=true,
    ymajorgrids=true,
    grid style=dashed,
    xtick={3,4,5,6,7,8,9},
]
\addplot[color=color4,mark=triangle*, mark size=1.8pt,line width=0.8pt] coordinates {(2.59,82.04)(2.74,82.27)(3.42,82.95)(3.82,83.23)(5.88,83.97)}; 

\addplot[color=color2,mark=triangle*, mark size=1.8pt,line width=0.8pt] coordinates {(2.54,80.89)(3.76,82.11)(4.94,82.92)(6.13,83.41)(7.27,83.76)(9.29,84.08)};

\addplot[color=color1,mark=*, mark size=1.4pt,line width=0.8pt] coordinates {(2.7,80.17)(3.63,81.96)(4.55,82.9)(5.44,83.38)(6.29,83.72)(7.83,84.11)(8.52,84.21)}; 

\legend{EAST, Conversational SimulMT, Llama3-MOMT w/ wait-k}

\end{axis}
\end{tikzpicture}}
}
\hspace{-2mm}
\resizebox{0.248\textwidth}{!}{
\subfigure[Zh$\rightarrow$En]{
\begin{tikzpicture}[baseline]
\begin{axis}[
    xlabel=AL,
    enlargelimits=0.05,
    font=\small,
    width=6cm,height=5.4cm,
    legend style={font=\tiny,
    at={(0.5,0.35)},
    anchor=south,
    legend columns=1},
    xmajorgrids=true,
    ymajorgrids=true,
    grid style=dashed,
    xmax=20,
]
\addplot[color=color4,mark=triangle*, mark size=1.8pt,line width=0.8pt] coordinates {(1.31,77.14)(1.56,77.4)(3.6,78.74)(4.88,79.24)(13.79,80.59)}; 

\addplot[color=color2,mark=triangle*, mark size=1.8pt,line width=0.8pt] coordinates {(1.48,77.26)(4.36,78.5)(6.2,79.18)(7.46,79.59)(9.94,80.16)(12.4,80.58)}; 

\addplot[color=color1,mark=*, mark size=1.4pt,line width=0.8pt] coordinates {(0.84,75.42)(2.78,76.73)(6.38,78.31)(11.58,79.66)(15.91,80.17)(19.22,80.37)}; 

\end{axis}
\end{tikzpicture}}
}
\hspace{-2mm}
\resizebox{0.245\textwidth}{!}{
\subfigure[Ru$\rightarrow$En]{
\begin{tikzpicture}[baseline]
\begin{axis}[
    xlabel=AL,
    enlargelimits=0.05,
    font=\small,
    width=6cm,height=5.4cm,
    legend style={font=\tiny,
    at={(0.64,0.45)},
    anchor=south,
    legend columns=1},
    xmajorgrids=true,
    ymajorgrids=true,
    grid style=dashed,
]
\addplot[color=color4,mark=triangle*, mark size=1.8pt,line width=0.8pt] coordinates {(2.24,83.66)(2.43,83.83)(3.41,84.22)(4.01,84.35)(6.78,84.6)}; 

\addplot[color=color2,mark=triangle*, mark size=1.8pt,line width=0.8pt] coordinates {(1.89,82.85)(3.14,83.65)(4.43,84.07)(5.66,84.35)(6.89,84.48)(9.05,84.7)}; 

\addplot[color=color1,mark=*, mark size=1.4pt,line width=0.8pt] coordinates {(2.01,76.51)(2.92,82.72)(3.82,83.95)(4.71,84.35)(6.38,84.55)(8.52,84.64)}; 

\end{axis}
\end{tikzpicture}}
}
\hspace{-2mm}
\resizebox{0.245\textwidth}{!}{
\subfigure[Cs$\rightarrow$En]{
\begin{tikzpicture}[baseline]
\begin{axis}[
    xlabel=AL,
    enlargelimits=0.05,
    font=\small,
    width=6cm,height=5.4cm,
    legend style={font=\tiny,
    at={(0.5,0.35)},
    anchor=south,
    legend columns=1},
    xmajorgrids=true,
    ymajorgrids=true,
    grid style=dashed,
    ytick={75,77,79,81,83,85},
]
\addplot[color=color4,mark=triangle*, mark size=1.8pt,line width=0.8pt] coordinates {(2.3,83.41)(2.55,83.73)(3.82,84.72)(4.43,84.94)(6.96,85.43)}; 

\addplot[color=color2,mark=triangle*, mark size=1.8pt,line width=0.8pt] coordinates {(2.07,82.39)(3.33,83.77)(4.56,84.44)(5.84,84.92)(7.05,85.11)}; 

\addplot[color=color1,mark=*, mark size=1.4pt,line width=0.8pt] coordinates {(2.1,75.58)(3.02,82.24)(3.94,83.76)(4.84,84.37)(6.58,85.02)(8.17,85.32)}; 

\end{axis}
\end{tikzpicture}}
}
~
\resizebox{0.265\textwidth}{!}{
\subfigure[En$\rightarrow$De]{
\begin{tikzpicture}[baseline]
\begin{axis}[
    ylabel=COMET,
    xlabel=AL,
    enlargelimits=0.05,
    font=\small,
    width=6cm,height=5.4cm,
    legend style={font=\tiny,
    at={(0.62,0.01)},
    anchor=south,
    legend columns=1},
    legend cell align=left,
    xmajorgrids=true,
    ymajorgrids=true,
    grid style=dashed,
    ymax=85,
    ytick={75,77,79,81,83,85},
]
\addplot[color=color4,mark=triangle*, mark size=1.8pt,line width=0.8pt] coordinates {(2.58,80.31)(3.09,81.7)(5.39,83.49)(7.54,84.01)(12.2,84.37)}; 

\addplot[color=color2,mark=triangle*, mark size=1.8pt,line width=0.8pt] coordinates {(2.36,78.52)(3.55,80.75)(4.64,81.52)(5.81,82.11)(6.98,82.76)(9.04,83.36)(10.78,83.68)}; 

\addplot[color=color1,mark=*, mark size=1.4pt,line width=0.8pt] coordinates {(2.64,78.83)(3.62,81.3)(4.57,82.6)(5.48,83.18)(7.18,83.72)(10.05,83.92)(13.76,84.01)}; 

\legend{EAST, Conversational SimulMT, Llama3-MOMT w/ wait-k}

\end{axis}
\end{tikzpicture}}
}
\hspace{-2mm}
\resizebox{0.245\textwidth}{!}{
\subfigure[En$\rightarrow$Zh]{
\begin{tikzpicture}[baseline]
\begin{axis}[
    xlabel=AL,
    enlargelimits=0.05,
    font=\small,
    width=6cm,height=5.4cm,
    legend style={font=\tiny,
    at={(0.62,0.01)},
    anchor=south,
    legend columns=1},
    xmajorgrids=true,
    ymajorgrids=true,
    grid style=dashed,
    xmax=12,
    ymax=86,
]
\addplot[color=color4,mark=triangle*, mark size=1.8pt,line width=0.8pt] coordinates {(2.64,81.69)(2.95,82.11)(4.68,83.7)(6.46,84.3)(10.22,84.87)}; 

\addplot[color=color2,mark=triangle*, mark size=1.8pt,line width=0.8pt] coordinates {(2.18,80.8)(3.62,82.22)(4.72,83.04)(5.93,83.86)(7.05,84.28)(9.14,84.79)(10.75,85.1)}; 

\addplot[color=color1,mark=*, mark size=1.4pt,line width=0.8pt] coordinates {(4.26,77.29)(5.04,80.2)(5.82,82.37)(7.32,84.26)(8.02,84.79)(9.96,85.46)(11.57,85.66)}; 

\end{axis}
\end{tikzpicture}}
}
\hspace{-2mm}
\resizebox{0.245\textwidth}{!}{
\subfigure[En$\rightarrow$Ru]{
\begin{tikzpicture}[baseline]
\begin{axis}[
    xlabel=AL,
    enlargelimits=0.05,
    font=\small,
    width=6cm,height=5.4cm,
    legend style={font=\tiny,
    at={(0.64,0.45)},
    anchor=south,
    legend columns=1},
    xmajorgrids=true,
    ymajorgrids=true,
    grid style=dashed,
    ymax=86,
]
\addplot[color=color4,mark=triangle*, mark size=1.8pt,line width=0.8pt] coordinates {(2.97,83.73)(3.48,84.6)(5.82,85.52)(7.4,85.63)(10.1,85.84)}; 

\addplot[color=color2,mark=triangle*, mark size=1.8pt,line width=0.8pt] coordinates {(2.41,81.7)(3.61,83.59)(4.69,84.19)(5.85,84.74)(7.0,85.2)(9.05,85.69)(10.8,85.94)}; 

\addplot[color=color1,mark=*, mark size=1.4pt,line width=0.8pt] coordinates {(3.06,79.96)(5.01,83.66)(6.78,84.77)(8.38,85.34)(9.78,85.52)(11.5,85.66)}; 

\end{axis}
\end{tikzpicture}}
}
\hspace{-2mm}
\resizebox{0.248\textwidth}{!}{
\subfigure[En$\rightarrow$Cs]{
\begin{tikzpicture}[baseline]
\begin{axis}[
    xlabel=AL,
    enlargelimits=0.05,
    font=\small,
    width=6cm,height=5.4cm,
    legend style={font=\tiny,
    at={(0.5,0.35)},
    anchor=south,
    legend columns=1},
    xmajorgrids=true,
    ymajorgrids=true,
    grid style=dashed,
    xmax=10.0,
    ytick={77,79,81,83,85,87},
]
\addplot[color=color4,mark=triangle*, mark size=1.8pt,line width=0.8pt] coordinates {(3.13,85.23)(3.77,85.98)(6.07,86.75)(7.53,86.83)(9.56,87.2)}; 

\addplot[color=color2,mark=triangle*, mark size=1.8pt,line width=0.8pt] coordinates {(2.41,82.66)(3.63,84.67)(4.71,85.56)(5.88,86.02)(7.02,86.69)(9.05,87.17)}; 

\addplot[color=color1,mark=*, mark size=1.4pt,line width=0.8pt] coordinates {(2.78,77.42)(3.8,80.5)(5.71,84.07)(6.59,84.66)(8.22,85.63)(9.65,85.88)};

\end{axis}
\end{tikzpicture}}
}

\caption{COMET-AL curves for main results on the WMT22 X$\rightarrow$En and En$\rightarrow$X test sets.}
\label{fig:result_wmt22_comet}
\end{figure*}

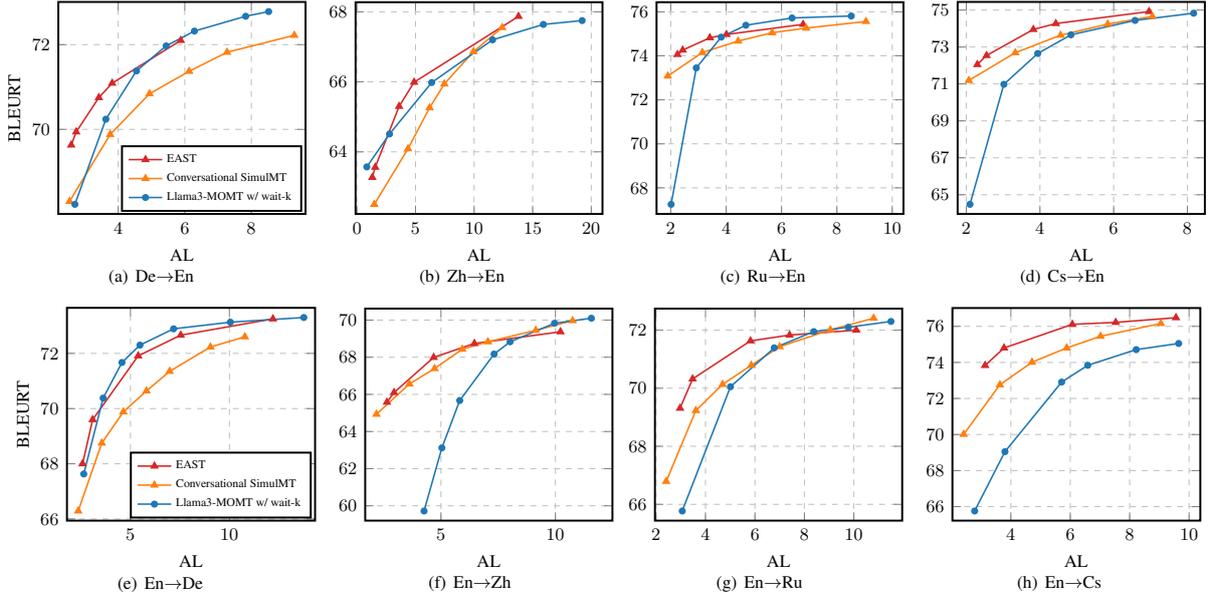
\begin{figure*}[t]
\pgfplotsset{
    every axis y label/.append style={at={(-0.12,0.5)}},
    every axis/.append style={line width=1.0pt},
}
\centering
\hspace{-2mm}
\resizebox{0.264\textwidth}{!}{
\subfigure[De$\rightarrow$En]{
\begin{tikzpicture}[baseline]
\begin{axis}[
    ylabel=BLEURT,
    xlabel=AL,
    enlargelimits=0.05,
    font=\small,
    width=6cm,height=5.4cm,
    legend cell align=left,
    legend style={font=\tiny,
    at={(0.62,0.01)},
    anchor=south,
    legend columns=1},
    xmajorgrids=true,
    ymajorgrids=true,
    grid style=dashed,
]
\addplot[color=color4,mark=triangle*, mark size=1.8pt,line width=0.8pt] coordinates {(2.59,69.63)(2.74,69.94)(3.42,70.75)(3.82,71.09)(5.88,72.1)}; 

\addplot[color=color2,mark=triangle*, mark size=1.8pt,line width=0.8pt] coordinates {(2.54,68.3)(3.76,69.88)(4.94,70.84)(6.13,71.37)(7.27,71.82)(9.29,72.22)}; 

\addplot[color=color1,mark=*, mark size=1.4pt,line width=0.8pt] coordinates {(2.7,68.23)(3.63,70.24)(4.55,71.38)(5.44,71.97)(6.29,72.32)(7.83,72.67)(8.52,72.78)}; 

\legend{EAST, Conversational SimulMT, Llama3-MOMT w/ wait-k}

\end{axis}
\end{tikzpicture}}
}
\hspace{-2mm}
\resizebox{0.248\textwidth}{!}{
\subfigure[Zh$\rightarrow$En]{
\begin{tikzpicture}[baseline]
\begin{axis}[
    xlabel=AL,
    enlargelimits=0.05,
    font=\small,
    width=6cm,height=5.4cm,
    legend style={font=\tiny,
    at={(0.5,0.35)},
    anchor=south,
    legend columns=1},
    xmajorgrids=true,
    ymajorgrids=true,
    grid style=dashed,
    ymax=68,
    xmax=20.0,
]
\addplot[color=color4,mark=triangle*, mark size=1.8pt,line width=0.8pt] coordinates {(1.31,63.27)(1.56,63.56)(3.6,65.3)(4.88,65.99)(13.79,67.87)}; 

\addplot[color=color2,mark=triangle*, mark size=1.8pt,line width=0.8pt] coordinates {(1.48,62.5)(4.36,64.09)(6.2,65.26)(7.46,65.94)(9.94,66.86)(12.4,67.55)}; 

\addplot[color=color1,mark=*, mark size=1.4pt,line width=0.8pt] coordinates {(0.84,63.57)(2.78,64.51)(6.38,65.98)(11.58,67.2)(15.91,67.64)(19.22,67.75)}; 

\end{axis}
\end{tikzpicture}}
}
\hspace{-2mm}
\resizebox{0.248\textwidth}{!}{
\subfigure[Ru$\rightarrow$En]{
\begin{tikzpicture}[baseline]
\begin{axis}[
    xlabel=AL,
    enlargelimits=0.05,
    font=\small,
    width=6cm,height=5.4cm,
    legend style={font=\tiny,
    at={(0.64,0.45)},
    anchor=south,
    legend columns=1},
    xmajorgrids=true,
    ymajorgrids=true,
    grid style=dashed,
    ymax=76,
    xmax=10,
]
\addplot[color=color4,mark=triangle*, mark size=1.8pt,line width=0.8pt] coordinates {(2.24,74.06)(2.43,74.26)(3.41,74.82)(4.01,74.97)(6.78,75.43)}; 

\addplot[color=color2,mark=triangle*, mark size=1.8pt,line width=0.8pt] coordinates {(1.89,73.08)(3.14,74.15)(4.43,74.67)(5.66,75.05)(6.89,75.26)(9.05,75.56)}; 

\addplot[color=color1,mark=*, mark size=1.4pt,line width=0.8pt] coordinates {(2.01,67.25)(2.92,73.45)(3.82,74.85)(4.71,75.39)(6.38,75.72)(8.52,75.81)}; 

\end{axis}
\end{tikzpicture}}
}
\hspace{-2mm}
\resizebox{0.245\textwidth}{!}{
\subfigure[Cs$\rightarrow$En]{
\begin{tikzpicture}[baseline]
\begin{axis}[
    xlabel=AL,
    enlargelimits=0.05,
    font=\small,
    width=6cm,height=5.4cm,
    legend style={font=\tiny,
    at={(0.5,0.35)},
    anchor=south,
    legend columns=1},
    xmajorgrids=true,
    ymajorgrids=true,
    grid style=dashed,
    ytick={65,67,69,71,73,75},
]
\addplot[color=color4,mark=triangle*, mark size=1.8pt,line width=0.8pt] coordinates {(2.3,72.05)(2.55,72.54)(3.82,73.95)(4.43,74.27)(6.96,74.92)}; 

\addplot[color=color2,mark=triangle*, mark size=1.8pt,line width=0.8pt] coordinates {(2.07,71.19)(3.33,72.69)(4.56,73.64)(5.84,74.24)(7.05,74.66)}; 

\addplot[color=color1,mark=*, mark size=1.4pt,line width=0.8pt] coordinates {(2.1,64.48)(3.02,70.98)(3.94,72.65)(4.84,73.66)(6.58,74.44)(8.17,74.83)}; 

\end{axis}
\end{tikzpicture}}
}
\hspace{-2mm}
\resizebox{0.265\textwidth}{!}{
\subfigure[En$\rightarrow$De]{
\begin{tikzpicture}[baseline]
\begin{axis}[
    ylabel=BLEURT,
    xlabel=AL,
    enlargelimits=0.05,
    font=\small,
    width=6cm,height=5.4cm,
    legend style={font=\tiny,
    at={(0.62,0.01)},
    anchor=south,
    legend columns=1},
    legend cell align=left,
    xmajorgrids=true,
    ymajorgrids=true,
    grid style=dashed,
]
\addplot[color=color4,mark=triangle*, mark size=1.8pt,line width=0.8pt] coordinates {(2.58,68.0)(3.09,69.6)(5.39,71.91)(7.54,72.66)(12.2,73.25)}; 

\addplot[color=color2,mark=triangle*, mark size=1.8pt,line width=0.8pt] coordinates {(2.36,66.29)(3.55,68.75)(4.64,69.88)(5.81,70.64)(6.98,71.35)(9.04,72.23)(10.78,72.6)}; 

\addplot[color=color1,mark=*, mark size=1.4pt,line width=0.8pt] coordinates {(2.64,67.63)(3.62,70.38)(4.57,71.67)(5.48,72.3)(7.18,72.89)(10.05,73.13)(13.76,73.3)};

\legend{EAST, Conversational SimulMT, Llama3-MOMT w/ wait-k}

\end{axis}
\end{tikzpicture}}
}
\hspace{-2mm}
\resizebox{0.245\textwidth}{!}{
\subfigure[En$\rightarrow$Zh]{
\begin{tikzpicture}[baseline]
\begin{axis}[
    xlabel=AL,
    enlargelimits=0.05,
    font=\small,
    width=6cm,height=5.4cm,
    legend style={font=\tiny,
    at={(0.5,0.35)},
    anchor=south,
    legend columns=1},
    xmajorgrids=true,
    ymajorgrids=true,
    grid style=dashed,
    xmax=12,
    ytick={60,62,64,66,68,70},
]
\addplot[color=color4,mark=triangle*, mark size=1.8pt,line width=0.8pt] coordinates {(2.64,65.58)(2.95,66.1)(4.68,67.99)(6.46,68.74)(10.22,69.37)}; 

\addplot[color=color2,mark=triangle*, mark size=1.8pt,line width=0.8pt] coordinates {(2.18,64.93)(3.62,66.56)(4.72,67.39)(5.93,68.43)(7.05,68.83)(9.14,69.45)(10.75,69.97)}; 

\addplot[color=color1,mark=*, mark size=1.4pt,line width=0.8pt] coordinates {(4.26,59.72)(5.04,63.12)(5.82,65.67)(7.32,68.17)(8.02,68.83)(9.96,69.83)(11.57,70.1)}; 

\end{axis}
\end{tikzpicture}}
}
\hspace{-3mm}
\resizebox{0.245\textwidth}{!}{
\subfigure[En$\rightarrow$Ru]{
\begin{tikzpicture}[baseline]
\begin{axis}[
    xlabel=AL,
    enlargelimits=0.05,
    font=\small,
    width=6cm,height=5.4cm,
    legend style={font=\tiny,
    at={(0.64,0.45)},
    anchor=south,
    legend columns=1},
    xmajorgrids=true,
    ymajorgrids=true,
    grid style=dashed,
]
\addplot[color=color4,mark=triangle*, mark size=1.8pt,line width=0.8pt] coordinates {(2.97,69.31)(3.48,70.32)(5.82,71.63)(7.4,71.83)(10.1,72.0)}; 

\addplot[color=color2,mark=triangle*, mark size=1.8pt,line width=0.8pt] coordinates {(2.41,66.79)(3.61,69.23)(4.69,70.13)(5.85,70.79)(7.0,71.43)(9.05,72.01)(10.8,72.41)}; 

\addplot[color=color1,mark=*, mark size=1.4pt,line width=0.8pt] coordinates {(3.06,65.77)(5.01,70.05)(6.78,71.39)(8.38,71.95)(9.78,72.1)(11.5,72.3)}; 

\end{axis}
\end{tikzpicture}}
}
\hspace{-2mm}
\resizebox{0.249\textwidth}{!}{
\subfigure[En$\rightarrow$Cs]{
\begin{tikzpicture}[baseline]
\begin{axis}[
    xlabel=AL,
    enlargelimits=0.05,
    font=\small,
    width=6cm,height=5.4cm,
    legend style={font=\tiny,
    at={(0.5,0.35)},
    anchor=south,
    legend columns=1},
    xmajorgrids=true,
    ymajorgrids=true,
    grid style=dashed,
    xmax=10.0,
    ytick={66,68,70,72,74,76},
]
\addplot[color=color4,mark=triangle*, mark size=1.8pt,line width=0.8pt] coordinates {(3.13,73.83)(3.77,74.8)(6.07,76.11)(7.53,76.22)(9.56,76.48)}; 

\addplot[color=color2,mark=triangle*, mark size=1.8pt,line width=0.8pt] coordinates {(2.41,70.01)(3.63,72.76)(4.71,74.01)(5.88,74.8)(7.02,75.45)(9.05,76.16)}; 

\addplot[color=color1,mark=*, mark size=1.4pt,line width=0.8pt] coordinates {(2.78,65.76)(3.8,69.05)(5.71,72.91)(6.59,73.84)(8.22,74.71)(9.65,75.05)}; 

\end{axis}
\end{tikzpicture}}
}

\caption{BLEURT-AL curves for main results on the WMT22 X$\rightarrow$En and En$\rightarrow$X test sets.}
\label{fig:result_wmt22_brt}
\end{figure*}

\section{Key Differences from Conversational SimulMT}
\label{app:key_diff}

\subsection{Data Construction}
Conversational SimulMT relies on an alignment tool and multi-step data augmentation to create SiMT data. However, this approach has significant limitations:
\textbf{(i)} The resulting subsequences may not represent meaningful semantic units.
\textbf{(ii)} The data construction method is based on offline parallel data, which introduces a domain mismatch with the SiMT paradigm.
\textbf{(iii)} The word alignments generated by the alignment tool fast\_align used in Conversational SimulMT have an error rate of approximately 30\%, potentially degrading the quality of synthetic data and leading to suboptimal performance.

In contrast, EAST leverages GPT-4 to segment source text into independent semantic units and generate corresponding simultaneous translations, effectively mitigating these issues. Moreover, our data construction considers varying latency levels, as the models should generate different translations for different latency requirements, an important consideration often overlooked in prior works. As shown in Figure \ref{fig:result_wmt15_bleu}, when training our SiMT-De-En-660K dataset in Conversational SimulMT format, we observe a consistent performance improvement of 2 BLEU across all latency settings. Notably, our dataset (660K samples) is an order of magnitude smaller than that of Conversational SimulMT (4M samples). Moreover, the results in Figure \ref{fig:data_scale} show that only 10K SiMT examples may be sufficient to achieve commendable translation quality.

This concludes that a limited amount of high-quality, latency-aware aligned data is more effective than large-scale, tool-aligned data.

\subsection{Training for Adaptive Read/Write Policy}
In Conversational SimulMT, the SFT data is structured into a chat (or message API) format during training, and only the loss on target tokens is computed to optimize the LLM's translation ability for incomplete text. However, during inference, it adopts a fixed policy (e.g., reading a fixed number of tokens at each step), leading to a mismatch between training and inference phases. In contrast, EAST introduces the training method the same as the pretraining, i.e., or next-token prediction (or completion API) format, to learn an adaptive read/write policy. Particularly, the loss is computed across source, target, and read/write tokens. This not only optimizes the translation performance but also enables LLMs to model adaptive read/write behaviors based on context, ensuring consistency between training and inference. As a result, EAST can effectively segment and translate source text into appropriate semantic units based on latency requirements specified by the instructions, achieving a latency-specific adaptive read/write policy.

\section{Data Statistics}
The data statistics for our SiMT and OMT datasets are illustrated in Table \ref{tab:simt_data} and \ref{tab:data_statistics}, respectively.

\begin{figure*}[t]
\pgfplotsset{
    every axis y label/.append style={at={(-0.12,0.5)}},
    every axis/.append style={line width=1.0pt},
}
\centering
\resizebox{0.262\textwidth}{!}{
\subfigure[De$\rightarrow$En]{
\begin{tikzpicture}[baseline]
\begin{axis}[
    ylabel=SacreBLEU,
    xlabel=AL,
    enlargelimits=0.05,
    font=\small,
    width=6.4cm,height=5.6cm,
    legend cell align=left,
    legend style={font=\tiny,
    at={(0.7,0.01)},
    anchor=south,
    legend columns=1},
    xmajorgrids=true,
    ymajorgrids=true,
    grid style=dashed,
    ymin=27.5,
]
\addplot[color=color4,mark=triangle*, mark size=1.8pt,line width=0.8pt] coordinates {(2.59,29.87)(2.74,30.09)(3.42,31.09)(3.82,31.61)(5.88,32.38)}; 

\addplot[color=color2,mark=triangle*, mark size=1.8pt,line width=0.8pt] coordinates {(2.57,29.79)(2.73,30.05)(3.43,30.76)(3.83,31.19)(6.16,32.1)}; 

\addplot[color=color7,mark=triangle*, mark size=1.8pt,line width=0.8pt] coordinates {(2.55,28.11)(2.86,28.78)(3.48,29.62)(4.73,31.1)(6.17,31.65)}; 

\addplot[color=color1,mark=triangle*, mark size=1.8pt,line width=0.8pt] coordinates {(3.0,29.12)(3.28,29.37)(5.02,30.6)(5.75,30.95)(7.93,31.5)}; 

\addplot[color=color3,mark=triangle*, mark size=1.8pt,line width=0.8pt] coordinates {(3.03,28.54)(3.42,28.97)(5.36,29.81)(5.91,30.02)(8.29,30.38)}; 

\legend{EAST, EAST-w/o-Offline, EAST-Only-Stage-II, EAST-Single-Stage, EAST-Stage-I} 
\end{axis}
\end{tikzpicture}}}
\hspace{-2mm}
\resizebox{0.248\textwidth}{!}{
\subfigure[Zh$\rightarrow$En]{
\begin{tikzpicture}[baseline]
\begin{axis}[
    xlabel=AL,
    enlargelimits=0.05,
    width=6.4cm,height=5.6cm,
    font=\small,
    legend style={font=\tiny,
    at={(0.5,0.35)},
    anchor=south,
    legend columns=1},
    xmajorgrids=true,
    ymajorgrids=true,
    grid style=dashed,
    xmax=40,
]

\addplot[color=color4,mark=triangle*, mark size=1.8pt,line width=0.8pt] coordinates {(1.31,18.74)(1.56,19.08)(3.6,21.43)(4.88,22.34)(13.79,25.0)}; 

\addplot[color=color2,mark=triangle*, mark size=1.8pt,line width=0.8pt] coordinates {(1.42,18.35)(1.77,18.74)(3.77,20.9)(5.31,22.02)(12.22,24.96)}; 

\addplot[color=color7,mark=triangle*, mark size=1.8pt,line width=0.8pt] coordinates {(0.22,16.9)(2.12,19.45)(3.66,21.17)(8.82,23.53)(14.69,24.3)}; 

\addplot[color=color1,mark=triangle*, mark size=1.8pt,line width=0.8pt] coordinates {((1.23,18.45)(1.82,19.35)(3.87,21.39)(5.56,22.55)(15.87,25.1)}; 

\addplot[color=color3,mark=triangle*, mark size=1.8pt,line width=0.8pt] coordinates {(31.1,21.09)(32.09,20.69)(34.91,20.62)(35.34,20.23)(38.43,19.88)};

\end{axis}
\end{tikzpicture}}
}
\hspace{-2mm}
\resizebox{0.248\textwidth}{!}{
\subfigure[Ru$\rightarrow$En]{
\begin{tikzpicture}[baseline]
\begin{axis}[
    xlabel=AL,
    enlargelimits=0.05,
    width=6.4cm,height=5.6cm,
    font=\small,
    legend style={font=\tiny,
    at={(0.64,0.45)},
    anchor=south,
    legend columns=1},
    xmajorgrids=true,
    ymajorgrids=true,
    grid style=dashed,
    xmax=10,
]
\addplot[color=color4,mark=triangle*, mark size=1.8pt,line width=0.8pt] coordinates {(2.24,36.35)(2.43,36.49)(3.41,37.42)(4.01,37.73)(6.78,38.92)}; 

\addplot[color=color2,mark=triangle*, mark size=1.8pt,line width=0.8pt] coordinates {(2.2,35.79)(2.42,36.2)(3.51,37.3)(4.36,37.97)(6.98,39.24)}; 

\addplot[color=color7,mark=triangle*, mark size=1.8pt,line width=0.8pt] coordinates {(2.39,35.51)(2.83,36.13)(3.63,37.05)(5.21,38.05)(6.97,38.53)}; 

\addplot[color=color1,mark=triangle*, mark size=1.8pt,line width=0.8pt] coordinates {(2.51,35.98)(2.85,36.27)(4.06,37.08)(4.61,37.55)(7.14,38.77)}; 

\addplot[color=color3,mark=triangle*, mark size=1.8pt,line width=0.8pt] coordinates {(4.69,35.61)(5.14,36.09)(6.6,36.83)(6.97,36.91)(9.66,37.43)};

\end{axis}
\end{tikzpicture}}
}
\hspace{-2mm}
\resizebox{0.246\textwidth}{!}{
\subfigure[Cs$\rightarrow$En]{
\begin{tikzpicture}[baseline]
\begin{axis}[
    xlabel=AL,
    enlargelimits=0.05,
    width=6.4cm,height=5.6cm,
    font=\small,
    legend style={font=\tiny,
    at={(0.5,0.35)},
    anchor=south,
    legend columns=1},
    xmajorgrids=true,
    ymajorgrids=true,
    grid style=dashed,
    xtick={2,3,4,5,6,7},
]
\addplot[color=color4,mark=triangle*, mark size=1.8pt,line width=0.8pt] coordinates {(2.3,37.47)(2.55,38.32)(3.82,40.8)(4.43,41.28)(6.96,42.99)}; 

\addplot[color=color2,mark=triangle*, mark size=1.8pt,line width=0.8pt] coordinates {(2.28,37.29)(2.52,38.15)(3.71,40.39)(4.43,41.35)(7.02,42.57)}; 

\addplot[color=color7,mark=triangle*, mark size=1.8pt,line width=0.8pt] coordinates {(2.34,36.04)(2.93,37.27)(4.04,38.84)(5.39,40.08)(7.25,41.11)}; 

\addplot[color=color1,mark=triangle*, mark size=1.8pt,line width=0.8pt] coordinates {(2.33,36.69)(2.7,37.92)(4.03,40.09)(4.76,40.7)(7.38,42.3)}; 

\addplot[color=color3,mark=triangle*, mark size=1.8pt,line width=0.8pt] coordinates {(2.29,33.86)(2.62,34.54)(4.18,37.37)(4.64,37.54)(6.67,39.56)}; 

\end{axis}
\end{tikzpicture}}
}
~
\resizebox{0.265\textwidth}{!}{
\subfigure[En$\rightarrow$De]{
\begin{tikzpicture}[baseline]
\begin{axis}[
    ylabel=SacreBLEU,
    xlabel=AL,
    enlargelimits=0.05,
    font=\small,
    width=6.4cm,height=5.6cm,
    legend style={font=\tiny,
    at={(0.7,0.01)},
    anchor=south,
    legend columns=1},
    legend cell align=left,
    xmajorgrids=true,
    ymajorgrids=true,
    grid style=dashed, 
    ymax=30,
    xtick={4,6,8,10,12,14},
]
\addplot[color=color4,mark=triangle*, mark size=1.8pt,line width=0.8pt] coordinates {(2.58,23.27)(3.09,24.57)(5.39,27.0)(7.54,27.99)(12.2,28.77)}; 

\addplot[color=color2,mark=triangle*, mark size=1.8pt,line width=0.8pt] coordinates {(2.59,23.32)(3.04,24.46)(5.35,26.96)(7.01,27.57)(11.5,28.6)}; 

\addplot[color=color7,mark=triangle*, mark size=1.8pt,line width=0.8pt] coordinates {(2.74,23.45)(3.63,25.24)(5.66,26.83)(9.13,28.23)(12.18,28.73)}; 

\addplot[color=color1,mark=triangle*, mark size=1.8pt,line width=0.8pt] coordinates {(2.76,23.97)(3.24,24.93)(5.74,27.75)(7.93,28.64)(12.48,29.57)}; 

\legend{EAST, EAST-w/o-Offline, EAST-Only-Stage-II, EAST-Single-Stage, }

\end{axis}
\end{tikzpicture}}
}
\hspace{-2mm}
\resizebox{0.245\textwidth}{!}{
\subfigure[En$\rightarrow$Zh]{
\begin{tikzpicture}[baseline]
\begin{axis}[
    xlabel=AL,
    enlargelimits=0.05,
    font=\small,
    width=6.4cm,height=5.6cm,
    legend style={font=\tiny,
    at={(0.7,0.35)},
    anchor=south,
    legend columns=1},
    xmajorgrids=true,
    ymajorgrids=true,
    grid style=dashed,
    ymax=38,
    ytick={30,32,34,36,38},
]
\addplot[color=color4,mark=triangle*, mark size=1.8pt,line width=0.8pt] coordinates {(2.64,31.86)(2.95,32.18)(4.68,34.37)(6.46,35.32)(10.22,37.15)}; 

\addplot[color=color2,mark=triangle*, mark size=1.8pt,line width=0.8pt] coordinates {(2.66,31.59)(2.99,32.39)(4.6,34.16)(6.89,35.24)(10.19,36.64)}; 

\addplot[color=color7,mark=triangle*, mark size=1.8pt,line width=0.8pt] coordinates {(2.79,31.85)(3.43,33.48)(4.91,34.97)(8.29,37.0)(10.59,37.63)}; 

\addplot[color=color1,mark=triangle*, mark size=1.8pt,line width=0.8pt] coordinates {(2.73,31.07)(3.23,32.25)(4.79,34.41)(6.24,35.09)(10.88,37.36)}; 

\end{axis}
\end{tikzpicture}}
}
\hspace{-2mm}
\resizebox{0.245\textwidth}{!}{
\subfigure[En$\rightarrow$Ru]{
\begin{tikzpicture}[baseline]
\begin{axis}[
    xlabel=AL,
    enlargelimits=0.05,
    font=\small,
    width=6.4cm,height=5.6cm,
    legend style={font=\tiny,
    at={(0.64,0.45)},
    anchor=south,
    legend columns=1},
    xmajorgrids=true,
    ymajorgrids=true,
    grid style=dashed,
]
\addplot[color=color4,mark=triangle*, mark size=1.8pt,line width=0.8pt] coordinates {(2.97,23.69)(3.48,24.63)(5.82,25.69)(7.4,26.21)(10.1,26.67)}; 

\addplot[color=color2,mark=triangle*, mark size=1.8pt,line width=0.8pt] coordinates {(2.99,23.61)(3.47,24.29)(5.76,25.28)(7.26,25.78)(10.03,26.07)}; 

\addplot[color=color7,mark=triangle*, mark size=1.8pt,line width=0.8pt] coordinates {(3.06,23.44)(4.0,24.56)(5.95,25.5)(8.61,26.02)(10.15,26.41)}; 

\addplot[color=color1,mark=triangle*, mark size=1.8pt,line width=0.8pt] coordinates {(3.15,23.73)(3.76,24.69)(6.15,25.85)(7.05,26.12)(10.54,26.45)}; 

\end{axis}
\end{tikzpicture}}
}
\hspace{-2mm}
\resizebox{0.247\textwidth}{!}{
\subfigure[En$\rightarrow$Cs]{
\begin{tikzpicture}[baseline]
\begin{axis}[
    xlabel=AL,
    enlargelimits=0.05,
    font=\small,
    width=6.4cm,height=5.6cm,
    legend style={font=\tiny,
    at={(0.5,0.35)},
    anchor=south,
    legend columns=1},
    xmajorgrids=true,
    ymajorgrids=true,
    grid style=dashed,
]
\addplot[color=color4,mark=triangle*, mark size=1.8pt,line width=0.8pt] coordinates {(3.13,24.88)(3.77,25.48)(6.07,26.96)(7.53,27.28)(9.56,27.62)}; 

\addplot[color=color2,mark=triangle*, mark size=1.8pt,line width=0.8pt] coordinates {(3.07,24.61)(3.58,25.52)(5.73,26.71)(7.21,27.02)(9.72,27.07)}; 

\addplot[color=color7,mark=triangle*, mark size=1.8pt,line width=0.8pt] coordinates {(3.38,24.61)(4.21,25.59)(6.1,27.0)(8.41,27.69)(9.7,27.95)}; 

\addplot[color=color1,mark=triangle*, mark size=1.8pt,line width=0.8pt] coordinates {(3.38,25.5)(4.18,26.46)(6.36,27.29)(7.43,27.55)(10.0,28.22)}; 

\end{axis}
\end{tikzpicture}}
}

\caption{BLEU-AL curves for different training strategies on the WMT22 X$\rightarrow$En and En$\rightarrow$X test sets. }
\label{fig:result_wmt22_bleu_ablation}
\end{figure*}
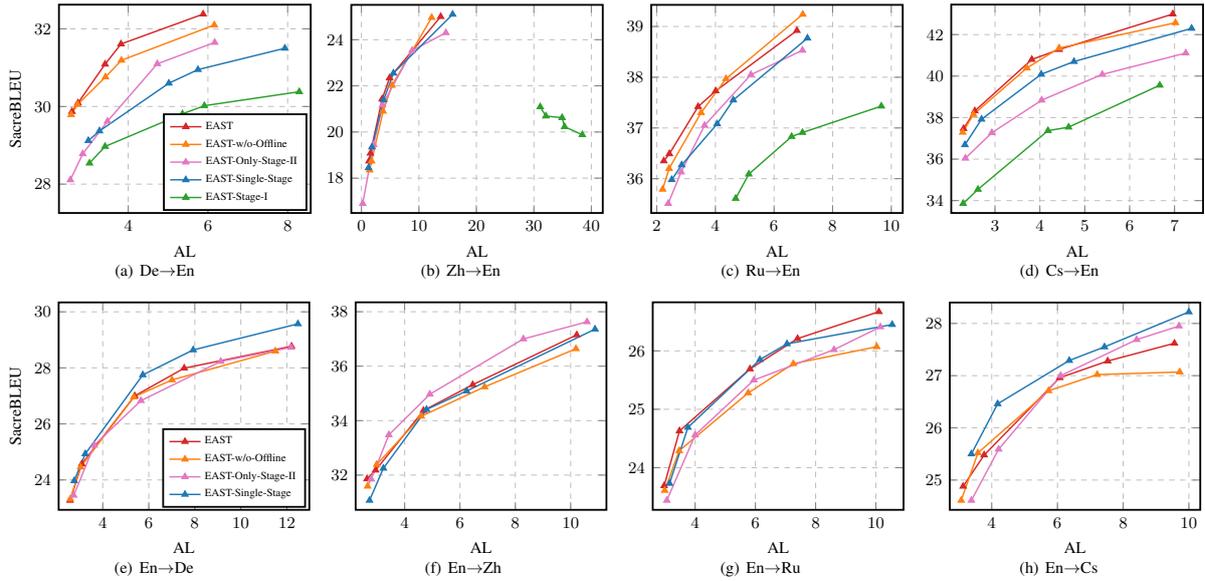

\section{Data processing}
\label{apd:data_filter}
For each sentence pair from WMT15 De$\rightarrow$En training set, we first filter out source sentences with less than 20 words.
We then utilize LLMs to generate SiMT chunk sequences at three different latency levels, as outlined in Eq.(\ref{eq:simt_data}), while filtering out invalid examples generated by LLMs with unequal numbers of source and target chunks. We found that such mismatches often result from non-monotonic translations. 
Next, we compute BLEURT scores between the LLM-generated translations and the ground-truth references, removing examples with scores below 80 to ensure translation quality.
Additionally, we merge any chunk that contains fewer than two words (or four characters for Chinese) with its subsequent chunk to avoid overly short segments.
This processing pipeline ensures that the final dataset is well-aligned and preserves monotonicity.

\section{Implementation Details} 
\label{apd:train_detail}
Our implementation is based on LLaMA-Factory\footnote{\url{https://github.com/hiyouga/LLaMA-Factory}} \cite{zheng-etal-2024-llamafactory}.
We train our models using Llama-3-8B-Instruct \cite{dubey2024llama} as the backbone, with full parameter tuning on Stage I and LoRA tuning on Stage II. 
All models are trained on 8 Nvidia A100 GPUs with a total batch size of 256, a learning rate of 1e-5, a cosine learning rate scheduler, a warm-up ratio of 0.1 and a maximum sequence length of 1024. 
The number of epochs is set to 1 for full parameter tuning and 2 for LoRA tuning, respectively.
When using LoRA, the LoRA rank, alpha, and dropout rate are set to 64, 128, and 0.05, respectively.

\section{Additional Results}
\label{app:additional_results}

\subsection{COMET-AL and BLEURT-AL Curves for Main Results}
\label{apd:comet_bleurt}

In Figures \ref{fig:result_wmt22_comet} and \ref{fig:result_wmt22_brt}, we present the COMET-AL and BLEURT-AL curves across multiple language pairs for the WMT22 X$\rightarrow$En and En$\rightarrow$X test sets.  

\noindent \textbf{COMET-AL Results:} EAST achieves consistently superior COMET scores compared to both Conversational SimulMT and Llama3-MNMT with wait-k across all language pairs, especially in the low-latency region. The adaptive policy of EAST effectively balances latency (AL) and quality, achieving a higher COMET score.

\noindent \textbf{BLEURT-AL Results:} EAST demonstrates strong BLEURT performance across the language pairs and achieves the best BLEURT scores on average. However, in En-De, Llama3-MNMT w/ wait-k achieves comparable BLEURT scores.

\subsection{How Different Training Strategies Affect Performance}
\label{apd:diff_train}

We conduct comprehensive experiments between EAST and the following variants.
\begin{enumerate}

\item \textbf{EAST-Stage-I}: Full-weight fine-tuning on the SiMT-De-En-660K dataset. 

\item \textbf{EAST-Single-Stage}: Full-weight fine-tuning on all the three datasets for one single epoch. 

\item \textbf{EAST-w/o-Offline}: Removing the Off-Multi-120K datasets in Stage II.

\item \textbf{EAST-Only-Stage-II}: Removing the Stage I fine-tuning.

\end{enumerate}

\noindent \textbf{SiMT Performance} The BLEU-AL curves of SiMT X$\rightarrow$En tasks are illustrated in first row of Figures \ref{fig:result_wmt22_bleu_ablation}. 
Notably, the EAST-Stage-I model, which is obtained from tuning the 660K De$\rightarrow$En SiMT data alone, shows reasonable performance under varying latency instructions in language pairs like Ru$\rightarrow$En and Cs$\rightarrow$En due to linguistic similarities among these Indo-European languages, facilitating better transfer learning. 
The Stage I model struggles with correct translation for Zh$\rightarrow$En because of the significant structural and grammatical differences between Chinese and Indo-European languages. 
However, EAST with two-stage training greatly improves the performance of Zh$\rightarrow$En becomes normal, underscoring the importance of this approach. 
Additionally, EAST with two-stage training further enhances performance across multiple latency ranges for De$\rightarrow$En, Ru$\rightarrow$En, and Cs$\rightarrow$En, with improvements of 0.5 to 1 in BLEU.

The Stage I model completely fails to perform En$\rightarrow$X translations, by repeating the source English sentence. 
This underperformance in En$\rightarrow$X directions without specific fine-tuning on those language pairs highlights the challenges of cross-linguistic semantic structures in SiMT. 
Fortunately, a very smaller multilingual dataset with 90K SiMT parallel pairs enable the LLM excellent performance on the reversed language directions En$\rightarrow$X. 
Compared with LLM-based SiMT baselines, EAST demonstrates superior performance across all 8 translation directions.
In Figure~\ref{fig:result_wmt22_comet_ablation} and \ref{fig:result_wmt22_brt_ablation} of the Appendix, we also plot the quality-latency curves of all methods with respect to COMET and BLEURT, revealing a trend similar to that of the BLEU-AL curves.

The EAST-w/o-Offline variant, which removes the OMT data in Stage II, shows a slight performance decline in De$\rightarrow$En and Ru$\rightarrow$En but maintained similar performance in other language pairs. 
The EAST-Only-Stage-II that omits the Stage I fine-tuning results in a performance degradation of about 1 BLEU for the X$\rightarrow$En on average, whereas the translation performance remains relatively unchanged for En$\rightarrow$X.
This suggests that learning a novel SiMT task may require a larger scale dataset. 
When fine-tuning with all three high-quality datasets in a single stage, EAST-Single-Stage demonstrates competitive performance across various delays and language orientations. 
Specifically, it achieves even higher BLEU-AL curves than the full two-stage training pipeline for both En$\rightarrow$De and En$\rightarrow$Cs. 
However, the two-stage training approach offers the advantage of better generalization to novel language directions with a reduced training schedule, avoiding re-training on the extensive Stage I dataset.

\begin{table*}[t]
\centering
\resizebox{0.88\textwidth}{!}{
\begin{tabular}{lcccccccccc}
\toprule
\multirow{2}{*}{Models} & \multicolumn{2}{c}{De$\rightarrow$En} & \multicolumn{2}{c}{Zh$\rightarrow$En} & \multicolumn{2}{c}{Ru$\rightarrow$En} & \multicolumn{2}{c}{Cs$\rightarrow$En} & \multicolumn{2}{c}{Average} \\
\cmidrule(lr){2-3} \cmidrule(lr){4-5} \cmidrule(lr){6-7} \cmidrule(lr){8-9} \cmidrule(lr){10-11}
 & BLEU & COMET & BLEU & COMET & BLEU & COMET & BLEU & COMET & BLEU & COMET \\
\midrule 
Llama3-MOMT & 31.98 & \textbf{84.89} & \textbf{25.48} & \textbf{81.26} & \underline{39.83} & \textbf{85.19} & \underline{44.92} & \textbf{86.23} & \textbf{35.55} & \textbf{84.39} \\
EAST & \textbf{32.55} & \underline{84.77} & 23.80 & 80.86 & \underline{39.83} & \underline{85.04} & \textbf{45.61} & \underline{86.20} & \underline{35.45} & \underline{84.22} \\
EAST-Single-Stage & 30.01 & 84.15 & 24.05 & 80.20 & 36.06 & 84.39 & 39.12 & 84.63 & 32.31 & 83.34 \\
EAST-w/o-Offline & \underline{32.37} & 84.55 & 22.42 & 80.85 & \textbf{40.29} & 84.80 & 41.21 & 85.41 & 34.07 & 83.90 \\
EAST-Only-Stage-II & 31.34 & 84.34 & \underline{24.90} & \underline{80.89} & 38.48 & 84.78 & 42.77 & 85.97 & 34.37 & 84.00 \\
\bottomrule
\toprule
\multirow{2}{*}{Models} & \multicolumn{2}{c}{En$\rightarrow$De} & \multicolumn{2}{c}{En$\rightarrow$Zh} & \multicolumn{2}{c}{En$\rightarrow$Ru} & \multicolumn{2}{c}{En$\rightarrow$Cs} & \multicolumn{2}{c}{Average} \\ \cmidrule(lr){2-3} \cmidrule(lr){4-5} \cmidrule(lr){6-7} \cmidrule(lr){8-9} \cmidrule(lr){10-11}
 & BLEU & COMET & BLEU & COMET & BLEU & COMET & BLEU & COMET & BLEU & COMET \\
\midrule 
Llama3-MOMT & 30.45 & \textbf{85.63} & \textbf{40.68} & \textbf{86.53} & 24.83 & 87.27 & \textbf{27.92} & \underline{88.36} & 30.97 & \textbf{86.95} \\
EAST & \underline{30.84} & 85.49 & \underline{40.17} & 86.31 & \textbf{26.79} & 87.13 & 26.63 & 88.17 & \underline{31.11} & 86.78 \\
EAST-Single-Stage & \textbf{30.85} & \underline{85.51} & 39.69 & \underline{86.43} & 26.57 & \underline{87.32} & \underline{27.76} & \textbf{88.40} & \textbf{31.22} & \underline{86.92} \\
EAST-w/o-Offline & 26.77 & 84.34 & 28.27 & 84.69 & 23.17 & 85.89 & 23.27 & 87.00 & 25.37 & 85.48 \\
EAST-Only-Stage-II & 30.50 & 85.44 & 39.03 & 86.31 & \underline{26.62 }& \textbf{87.40} & 26.85 & 88.32 & 30.75 & 86.87 \\
\bottomrule
\end{tabular}
}
\caption{Offline results for different training strategies on the WMT22 X$\rightarrow$En and En$\rightarrow$X test sets. \textbf{Bold} values denote the highest scores, while the \underline{underlined} values indicate the second highest scores.}
\label{tab:offline_mt_ablation}
\end{table*}

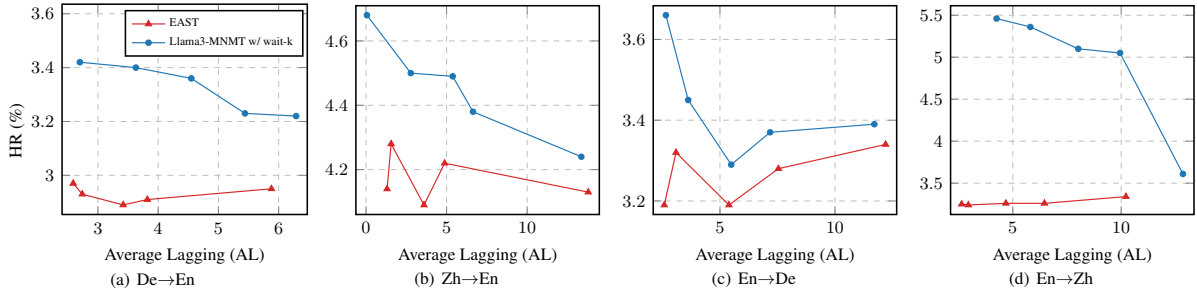
\begin{figure*}[t]
\pgfplotsset{
    every axis y label/.append style={at={(-0.12,0.4)}},
    every axis/.append style={line width=0.8pt},
}
\centering
\resizebox{0.26\textwidth}{!}{
\subfigure[De$\rightarrow$En]{
\begin{tikzpicture}[baseline]
\begin{axis}[
    ylabel=HR (\%),
    xlabel=Average Lagging (AL),
    enlargelimits=0.05,
    font=\small,
    width=6cm,height=5.4cm,
    legend style={font=\tiny,
    at={(0.62,0.75)},
    anchor=south,
    legend columns=1},
    legend cell align=left,
    xmajorgrids=true,
    ymajorgrids=true,
    grid style=dashed,
    ymax=3.6,
]
\addplot[color=color4,mark=triangle*, mark size=1.8pt,line width=0.6pt] coordinates {(2.59,2.97)(2.74,2.93)(3.42,2.89)(3.82,2.91)(5.88,2.95)};

\addplot[color=color1,mark=*, mark size=1.4pt,line width=0.6pt] coordinates {(2.7,3.42)(3.63,3.4)(4.55,3.36)(5.44,3.23)(6.29,3.22)};

\legend{EAST, Llama3-MNMT w/ wait-k}
\end{axis}
\end{tikzpicture}}}
\hspace{-2mm}
\resizebox{0.245\textwidth}{!}{
\subfigure[Zh$\rightarrow$En]{
\begin{tikzpicture}[baseline]
\begin{axis}[
    xlabel=Average Lagging (AL),
    enlargelimits=0.05,
    width=6cm,height=5.4cm,
    font=\small,
    legend style={font=\tiny,
    at={(0.5,0.35)},
    anchor=south,
    legend columns=1},
    xmajorgrids=true,
    ymajorgrids=true,
    grid style=dashed,
]

\addplot[color=color4,mark=triangle*, mark size=1.8pt,line width=0.6pt] coordinates {(1.31,4.14)(1.56,4.28)(3.6,4.09)(4.88,4.22)(13.79,4.13)};

\addplot[color=color1,mark=*, mark size=1.4pt,line width=0.6pt] coordinates {(0.06,4.68)(2.78,4.5)(5.38,4.49)(6.64,4.38)(13.36,4.24)};

\end{axis}
\end{tikzpicture}}
}
\hspace{-2mm}
\resizebox{0.245\textwidth}{!}{
\subfigure[En$\rightarrow$De]{
\begin{tikzpicture}[baseline]
\begin{axis}[
    xlabel=Average Lagging (AL),
    enlargelimits=0.05,
    font=\small,
    width=6cm,height=5.4cm,
    legend style={font=\tiny,
    at={(0.62,0.01)},
    anchor=south,
    legend columns=1},
    xmajorgrids=true,
    ymajorgrids=true,
    grid style=dashed,
]
\addplot[color=color4,mark=triangle*, mark size=1.8pt,line width=0.6pt] coordinates {(2.58,3.19)(3.09,3.32)(5.39,3.19)(7.54,3.28)(12.2,3.34)}; 

\addplot[color=color1,mark=*, mark size=1.4pt,line width=0.6pt] coordinates {(2.65,3.66)(3.62,3.45)(5.49,3.29)(7.18,3.37)(11.71,3.39)};

\end{axis}
\end{tikzpicture}}
}
\hspace{-2mm}
\resizebox{0.245\textwidth}{!}{
\subfigure[En$\rightarrow$Zh]{
\begin{tikzpicture}[baseline]
\begin{axis}[
    xlabel=Average Lagging (AL),
    enlargelimits=0.05,
    font=\small,
    width=6cm,height=5.4cm,
    legend style={font=\tiny,
    at={(0.5,0.35)},
    anchor=south,
    legend columns=1},
    xmajorgrids=true,
    ymajorgrids=true,
    grid style=dashed,
    ymax=5.5,
    ytick={3.0,3.5,4.0,4.5,5.0,5.5},
]
\addplot[color=color4,mark=triangle*, mark size=1.8pt,line width=0.6pt] coordinates {(2.64,3.25)(2.95,3.24)(4.68,3.26)(6.46,3.26)(10.22,3.34)}; 

\addplot[color=color1,mark=*, mark size=1.4pt,line width=0.6pt] coordinates {(4.25,5.46)(5.81,5.36)(8.02,5.1)(9.95,5.05)(12.85,3.61)}; 

\end{axis}
\end{tikzpicture}}
}

\caption{The hallucination rate (HR) against the latency metrics (AL) on the WMT22 test sets. }
\label{fig:hr}
\end{figure*}

\noindent \textbf{Offline Performance} 
We also evaluate the performance of offline translation on the WMT22 test set, as presented in Table \ref{tab:offline_mt_ablation}. 
Compared to offline NMT model Llama3-MNMT and other variants, EAST maintained comparable or superior offline translation performance across the eight language directions, indicating that our two-stage SFT process effectively maintains translation quality for offline NMT. 
EAST-Single-Stage shows excellent translation performance for En$\rightarrow$X, although it slightly underperforms by about 3 BLEU and 1 COMET for X$\rightarrow$En.
The EAST-w/o-Offline model, not even trained on offline translation data, still performed well, particularly for X$\rightarrow$En.
This can be attributed to the fact that our high-latency SiMT data has context that is close to being as informative as the offline NMT data.
Similar to the trend in SiMT, the offline performance of the EAST-Only-Stage-II drops by 1 BLEU and 0.22 COMET for X$\rightarrow$En, while the performance remains relatively stable for En$\rightarrow$X. 
In summary, these results highlight EAST not only excels in high-quality simultaneous translation but also ensures that the offline translation capabilities are not compromised.

\subsection{What Is hallucination Rate of The LLM-based SiMT?}
\label{apd:hallu_rate}

Hallucination is a significant challenge in traditional SiMT, as the models begin translating while receiving input. 
This can prompt incorrect assumptions about the content yet to be received, resulting in hallucinated outputs. 
Additionally, hallucination is a common issue in the outputs of LLMs across various generation tasks. 
Therefore, it is more essential to evaluate the hallucination phenomenon in LLM-based SiMT. 
To effectively measure the hallucinations in our case, we utilize the hallucination rate (HR) metric \citep{chen-etal-2021-improving-simultaneous}, which quantifies the proportion of target words in the hypothesis that do not align with any source words. 
For this, we employ the fast-align\footnote{\url{https://github.com/clab/fast\_align}} tool to identify word-level alignments between the source text and the target translation.

Figure \ref{fig:hr} illustrates the HR comparison on En$\leftrightarrow$De and En$\leftrightarrow$Zh test sets. 
EAST consistently demonstrates a lower hallucination rate across all latency levels and test sets compared to the Llama3-MOMT w/ wait-$k$.
Unlike the wait-$k$ policy, EAST can adaptively determine reading and writing actions based on the semantic context. 
This prevents the model from prematurely generating translations, thereby reducing the production of hallucinated content and ensuring translations that are more accurate and faithful to the source text.

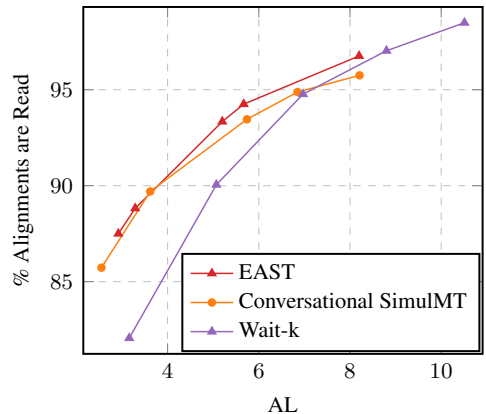
\begin{figure}[t]
\pgfplotsset{
    every axis y label/.append style={at={(-0.1,0.5)}},
    every axis/.append style={line width=0.8pt},
}
\centering
\resizebox{0.4\textwidth}{!}{
\begin{tikzpicture}[baseline]
\begin{axis}[
    ylabel=\% Alignments are Read,
    xlabel=AL,
    enlargelimits=0.05,
    font=\small,
    width=7.2cm,height=6.5cm,
    legend cell align=left,
    legend style={font=\footnotesize,
    at={(0.62,0.005)},
    anchor=south,
    legend columns=1},
    xmajorgrids=true,
    ymajorgrids=true,
    grid style=dashed,
]

\addplot[color=color4, mark=triangle*, mark size=1.8pt,line width=0.6pt] coordinates {(2.92,87.5)(3.29,88.82)(5.2,93.34)(5.67,94.25)(8.2,96.76)
}; 

\addplot[color=color2,mark=*, mark size=1.4pt,line width=0.6pt] coordinates {(2.55,85.73)(3.62,89.7)(5.74,93.46)(6.85,94.88)(8.21,95.75)};

\addplot[color=color5,mark=triangle*, mark size=1.8pt,line width=0.6pt] coordinates {(3.16,82.06)(5.07,90.05)(6.97,94.77)(8.80,97.03)(10.51,98.49)};

\legend{EAST, Conversational SimulMT, Wait-k} 

\end{axis}
\end{tikzpicture}}
\caption{The proportion of the ground-truth aligned source tokens received before translating.}
\label{fig:policy_eval}
\end{figure}

\subsection{Quality of Translation Policy}
\label{apd:qua_policy}

To evaluate the quality of our translation policy, we conduct experiments on the manually aligned RWTH De$\rightarrow$En alignment dataset\footnote{\url{https://www-i6.informatik.rwth-aachen.de/goldAlignment/}}. Following \citep{zhang-feng-2022-information}, we measure the proportion of ground-truth aligned source tokens that are read before generating each target token. Specifically, for a target token $y_i$, the number of source tokens read ($g_i$) must be at least equal to the ground-truth aligned source position ($a_i$). This ensures that the alignment between $y_i$ and $x_{a_i}$ is satisfied during the SiMT process.
The proportion is calculated as follows:
\begin{equation}
    \text{A} = \frac{1}{T} \sum_{i=1}^T \mathbb{I}_{a_i \leq g_i}
\end{equation}
where $T$ is the total number of target tokens and $\mathbb{I}_{a_i \leq g_i}$ counts the number of $a_i \leq g_i$.

As shown in Figure \ref{fig:policy_eval}, EAST consistently achieves the higher percentage of aligned source tokens read before translating across most latency levels compared to Conversational SimulMT and Wait-$k$. This result indicates that EAST better adheres to the ground-truth alignment, ensuring sufficient source context is read before generating target tokens.

\subsection{Analysis on Different Backbone LLMs}
\label{apd:diff_backbone}
In this section, we analyze the impact of different backbone models, Llama-3-8B-Instruct, Llama-3.1-8B-Instruct and Qwen2.5-7B-Instruct, on translation performance.

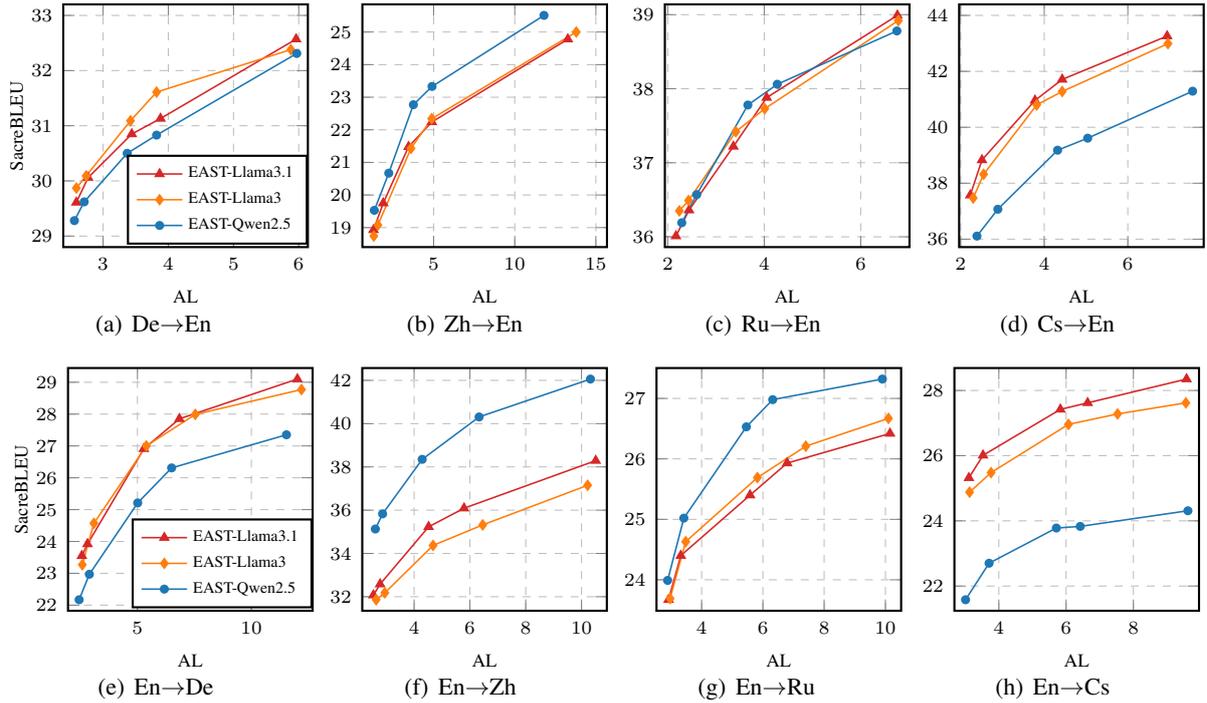
\begin{figure*}[t]
\pgfplotsset{
    every axis y label/.append style={at={(-0.12,0.5)}},
    every axis/.append style={line width=0.8pt},
}
\centering
\subfigure[De$\rightarrow$En]{
\begin{tikzpicture}[baseline]
\begin{axis}[
    ylabel=SacreBLEU,
    xlabel=AL,
    enlargelimits=0.05,
    font=\scriptsize,
    width=4.8cm,height=4.8cm,
    legend cell align=left,
    legend style={font=\tiny,
    at={(0.63,0.01)},
    anchor=south,
    legend columns=1},
    xmajorgrids=true,
    ymajorgrids=true,
    grid style=dashed,
    ymin=29,ymax=33,
]

\addplot[color=color4,mark=triangle*, mark size=1.8pt,line width=0.6pt] coordinates {(2.59,29.61)(2.77,30.06)(3.44,30.85)(3.88,31.13)(5.96,32.57)}; 

\addplot[color=color2,mark=diamond*, mark size=1.8pt,line width=0.6pt] coordinates {(2.59,29.87)(2.74,30.09)(3.42,31.09)(3.82,31.61)(5.88,32.38)};

\addplot[color=color1,mark=*, mark size=1.4pt,line width=0.6pt] coordinates {(2.56,29.28)(2.71,29.62)(3.37,30.5)(3.82,30.83)(5.97,32.31)}; 

\legend{EAST-Llama3.1, EAST-Llama3, EAST-Qwen2.5}

\end{axis}
\end{tikzpicture}}
\hspace{-2mm}
\subfigure[Zh$\rightarrow$En]{
\begin{tikzpicture}[baseline]
\begin{axis}[
    xlabel=AL,
    enlargelimits=0.05,
    width=4.8cm,height=4.8cm,
    font=\scriptsize,
    legend style={font=\tiny,
    at={(0.5,0.35)},
    anchor=south,
    legend columns=1},
    xmajorgrids=true,
    ymajorgrids=true,
    grid style=dashed,
    xmax=15,
    ytick={19,20,21,22,23,24,25},
]

\addplot[color=color4,mark=triangle*, mark size=1.8pt,line width=0.6pt] coordinates {(1.29,18.93)(1.89,19.75)(3.45,21.48)(4.89,22.24)(13.27,24.78)}; 

\addplot[color=color2,mark=diamond*, mark size=1.8pt,line width=0.6pt] coordinates {(1.31,18.74)(1.56,19.08)(3.6,21.43)(4.88,22.34)(13.79,25.0)}; 

\addplot[color=color1,mark=*, mark size=1.4pt,line width=0.6pt] coordinates {(1.34,19.53)(2.23,20.67)(3.75,22.77)(4.9,23.33)(11.8,25.51)}; 

\end{axis}
\end{tikzpicture}}
\hspace{-2mm}
\subfigure[Ru$\rightarrow$En]{
\begin{tikzpicture}[baseline]
\begin{axis}[
    xlabel=AL,
    enlargelimits=0.05,
    width=4.8cm,height=4.8cm,
    font=\scriptsize,
    legend style={font=\tiny,
    at={(0.64,0.45)},
    anchor=south,
    legend columns=1},
    xmajorgrids=true,
    ymajorgrids=true,
    grid style=dashed,
]
\addplot[color=color4,mark=triangle*, mark size=1.8pt,line width=0.6pt] coordinates {(2.17,36.01)(2.44,36.36)(3.36,37.22)(4.05,37.88)(6.76,38.99)}; 

\addplot[color=color2,mark=diamond*, mark size=1.8pt,line width=0.6pt] coordinates {(2.24,36.35)(2.43,36.49)(3.41,37.42)(4.01,37.73)(6.78,38.92)}; 

\addplot[color=color1,mark=*, mark size=1.4pt,line width=0.6pt] coordinates {(2.29,36.19)(2.6,36.57)(3.66,37.78)(4.27,38.06)(6.75,38.78)}; 

\end{axis}
\end{tikzpicture}}
\hspace{-2mm}
\subfigure[Cs$\rightarrow$En]{
\begin{tikzpicture}[baseline]
\begin{axis}[
    xlabel=AL,
    enlargelimits=0.05,
    width=4.8cm,height=4.8cm,
    font=\scriptsize,
    legend style={font=\tiny,
    at={(0.5,0.35)},
    anchor=south,
    legend columns=1},
    xmajorgrids=true,
    ymajorgrids=true,
    grid style=dashed,
    ymax=44,
]
\addplot[color=color4,mark=triangle*, mark size=1.8pt,line width=0.6pt] coordinates {(2.23,37.57)(2.51,38.83)(3.78,40.96)(4.43,41.71)(6.94,43.26)}; 

\addplot[color=color2,mark=diamond*, mark size=1.8pt,line width=0.6pt] coordinates {(2.3,37.47)(2.55,38.32)(3.82,40.8)(4.43,41.28)(6.96,42.99)};

\addplot[color=color1,mark=*, mark size=1.4pt,line width=0.6pt] coordinates {(2.39,36.11)(2.89,37.07)(4.32,39.18)(5.04,39.61)(7.55,41.29)}; 

\end{axis}
\end{tikzpicture}}
\subfigure[En$\rightarrow$De]{
\begin{tikzpicture}[baseline]
\begin{axis}[
    ylabel=SacreBLEU,
    xlabel=AL,
    enlargelimits=0.05,
    font=\scriptsize,
    width=4.8cm,height=4.8cm,
    legend style={font=\tiny,
    at={(0.63,0.01)},
    anchor=south,
    legend columns=1},
    legend cell align=left,
    xmajorgrids=true,
    ymajorgrids=true,
    grid style=dashed,
    ytick={22,23,24,25,26,27,28,29},
]
\addplot[color=color4,mark=triangle*, mark size=1.8pt,line width=0.6pt] coordinates {(2.56,23.55)(2.81,23.92)(5.28,26.91)(6.84,27.85)(12.02,29.1)}; 

\addplot[color=color2,mark=diamond*, mark size=1.8pt,line width=0.6pt] coordinates {(2.58,23.27)(3.09,24.57)(5.39,27.0)(7.54,27.99)(12.2,28.77)};

\addplot[color=color1,mark=*, mark size=1.4pt,line width=0.6pt] coordinates {(2.44,22.17)(2.89,22.97)(5.01,25.21)(6.5,26.31)(11.54,27.35)}; 

\legend{EAST-Llama3.1, EAST-Llama3, EAST-Qwen2.5}

\end{axis}
\end{tikzpicture}}
\hspace{-2mm}
\subfigure[En$\rightarrow$Zh]{
\begin{tikzpicture}[baseline]
\begin{axis}[
    xlabel=AL,
    enlargelimits=0.05,
    font=\scriptsize,
    width=4.8cm,height=4.8cm,
    legend style={font=\tiny,
    at={(0.5,0.35)},
    anchor=south,
    legend columns=1},
    xmajorgrids=true,
    ymajorgrids=true,
    grid style=dashed,
    ytick={32,34,36,38,40,42},
]
\addplot[color=color4,mark=triangle*, mark size=1.8pt,line width=0.6pt] coordinates {(2.54,32.07)(2.78,32.58)(4.52,35.23)(5.79,36.09)(10.51,38.29)}; 

\addplot[color=color2,mark=diamond*, mark size=1.8pt,line width=0.6pt] coordinates {(2.64,31.86)(2.95,32.18)(4.68,34.37)(6.46,35.32)(10.22,37.15)};

\addplot[color=color1,mark=*, mark size=1.4pt,line width=0.6pt] coordinates {(2.61,35.13)(2.87,35.83)(4.29,38.35)(6.33,40.31)(10.32,42.06)}; 

\end{axis}
\end{tikzpicture}}
\hspace{-2mm}
\subfigure[En$\rightarrow$Ru]{
\begin{tikzpicture}[baseline]
\begin{axis}[
    xlabel=AL,
    enlargelimits=0.05,
    font=\scriptsize,
    width=4.8cm,height=4.8cm,
    legend style={font=\tiny,
    at={(0.64,0.45)},
    anchor=south,
    legend columns=1},
    xmajorgrids=true,
    ymajorgrids=true,
    grid style=dashed,
]
\addplot[color=color4,mark=triangle*, mark size=1.8pt,line width=0.6pt] coordinates {(2.92,23.67)(3.32,24.4)(5.58,25.4)(6.79,25.93)(10.15,26.42)}; 

\addplot[color=color2,mark=diamond*, mark size=1.8pt,line width=0.6pt] coordinates {(2.97,23.69)(3.48,24.63)(5.82,25.69)(7.4,26.21)(10.1,26.67)};

\addplot[color=color1,mark=*, mark size=1.4pt,line width=0.6pt] coordinates {(2.89,23.99)(3.42,25.02)(5.46,26.53)(6.32,26.98)(9.9,27.32)}; 

\end{axis}
\end{tikzpicture}}
\hspace{-2mm}
\subfigure[En$\rightarrow$Cs]{
\begin{tikzpicture}[baseline]
\begin{axis}[
    xlabel=AL,
    enlargelimits=0.05,
    font=\scriptsize,
    width=4.8cm,height=4.8cm,
    legend style={font=\tiny,
    at={(0.5,0.35)},
    anchor=south,
    legend columns=1},
    xmajorgrids=true,
    ymajorgrids=true,
    grid style=dashed,
]
\addplot[color=color4,mark=triangle*, mark size=1.8pt,line width=0.6pt] coordinates {(3.11,25.32)(3.53,26.01)(5.83,27.42)(6.64,27.62)(9.58,28.35)}; 

\addplot[color=color2,mark=diamond*, mark size=1.8pt,line width=0.6pt] coordinates {(3.13,24.88)(3.77,25.48)(6.07,26.96)(7.53,27.28)(9.56,27.62)};

\addplot[color=color1,mark=*, mark size=1.4pt,line width=0.6pt] coordinates {(3.01,21.58)(3.71,22.7)(5.71,23.78)(6.42,23.83)(9.62,24.31)}; 

\end{axis}
\end{tikzpicture}}

\caption{BLEU-AL curves for different backbone LLMs on the WMT22 X$\rightarrow$En and En$\rightarrow$X test sets. }
\label{fig:result_llm_wmt22_bleu}
\end{figure*}

\paragraph{SiMT Performance} As shown in Figure \ref{fig:result_llm_wmt22_bleu}, EAST-Llama3.1 achieves similar or higher performance compared to EAST-Llama3 across all eight language pairs, indicating that a more powerful backbone model leads to better translation quality. 
The curves for EAST-Llama3.1 and EAST-Llama3 are generally higher than those for EAST-Qwen2.5, demonstrating the superiority of Llama-based models in multilingual SiMT tasks.
However, for Zh$\rightarrow$En, En$\rightarrow$Zh and En$\rightarrow$Ru, EAST-Qwen2.5 achieves better performance than both Llama models. 
This can be attributed to the pretraining data distribution, where Qwen2.5 likely benefits from having been pre-trained on a larger proportion of Chinese monolingual data, enhancing its performance on tasks involving the Chinese and Russian.

\paragraph{Offline Performance} The performance trends observed in Table \ref{tab:offline_llm_mt} align closely with those seen in SiMT. EAST-Llama3.1 consistently outperforms EAST-Llama3, while EAST-Qwen2.5 delivers competitive results only in Zh$\rightarrow$En and En$\rightarrow$Zh. Importantly, EAST models achieve translation performance comparable to their offline counterparts, demonstrating that the EAST framework effectively preserves high translation quality.

\subsection{Comparison with GPT-4}
\label{apd:gpt4}
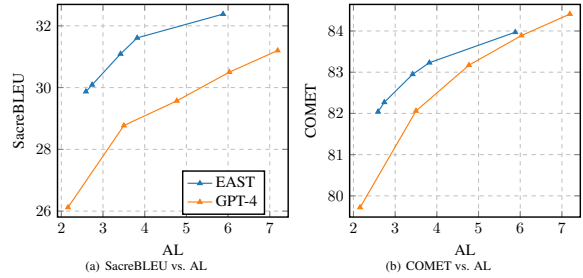
\begin{figure}[t]
\pgfplotsset{
  width=7.2cm,height=6.8cm,
    every axis y label/.append style={at={(-0.12,0.5)}},
    every axis/.append style={line width=1.0pt},
}
\centering
\resizebox{0.241\textwidth}{!}{
\subfigure[SacreBLEU vs. AL]{
\begin{tikzpicture}[baseline]
\begin{axis}[
    ylabel=SacreBLEU,
    xlabel=AL,
    enlargelimits=0.05,
    legend cell align=left,
    legend style={
    at={(0.72,0.01)},
    anchor=south,
    legend columns=1},
    xmajorgrids=true,
    ymajorgrids=true,
    grid style=dashed,
]

\addplot[color=color1,mark=triangle*, mark size=1.8pt,line width=0.8pt] coordinates {(2.59,29.87)(2.74,30.09)(3.42,31.09)(3.82,31.61)(5.88,32.38)};  

\addplot[color=color2,mark=triangle*, mark size=1.8pt,line width=0.8pt] coordinates {(2.16,26.12)(3.5,28.77)(4.77,29.57)(6.03,30.5)(7.19,31.2)}; 

\legend{EAST, GPT-4}
\end{axis}
\end{tikzpicture}}
}
\hspace{-2mm}
\resizebox{0.241\textwidth}{!}{
\subfigure[COMET vs. AL]{
\begin{tikzpicture}[baseline]
\begin{axis}[
    ylabel=COMET,
    xlabel=AL,
    enlargelimits=0.05,
    legend style={font=\small,
    at={(0.72,0.01)},
    anchor=south,
    legend columns=1},
    xmajorgrids=true,
    ymajorgrids=true,
    grid style=dashed,
]

\addplot[color=color1,mark=triangle*, mark size=1.8pt,line width=0.8pt] coordinates {(2.59,82.04)(2.74,82.27)(3.42,82.95)(3.82,83.23)(5.88,83.97)}; 

\addplot[color=color2,mark=triangle*, mark size=1.8pt,line width=0.8pt] coordinates {(2.16,79.72)(3.5,82.06)(4.77,83.17)(6.03,83.89)(7.19,84.41)}; 

\end{axis}
\end{tikzpicture}}
}
\caption{Translation quality and latency results on the WMT22 De$\rightarrow$En test set.}
\label{fig:result_gpt_deen}
\end{figure}

\begin{figure}[t]
\centering
\includegraphics[width=0.86\linewidth]{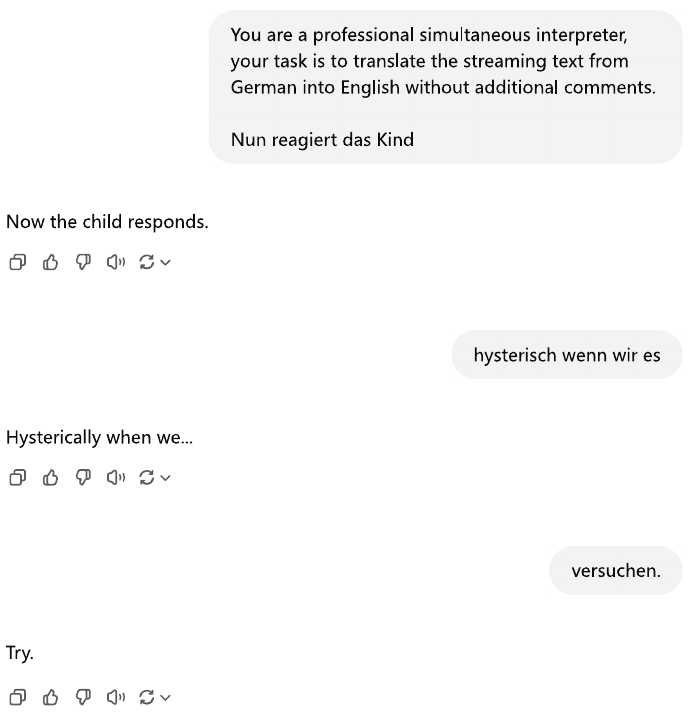}
\caption{A SiMT case for GPT-4.}
\label{fig:gpt_case}
\end{figure}

GPT-4 demonstrates superior offline translation quality in Table \ref{tab:offline_mt}, largely due to its extensive offline translation training corpus and larger model size.
Since GPT-4 is a closed-source model, it cannot effectively execute adaptive SiMT as EAST does. Instead, we use a fixed-policy inference strategy similar to Conversational SimulMT, where a fixed number of words are read before GPT-4 generates the corresponding translation. 
The experimental results are shown in Figure \ref{fig:result_gpt_deen}.
The results indicate that GPT-4 significantly underperforms EAST across all latency levels. A key reason for this performance gap is that GPT-4 has not been fine-tuned on SiMT-specific data, which can't efficiently perform zero-shot simultaneous translations. For example, when provided with partial source input, GPT-4 tends to add a period at the end of this response (Figure \ref{fig:gpt_case}), leading to suboptimal translations and reduced performance.

\begin{table}[t]
\centering
\small
\begin{tabular}{lccc}
\toprule
\textbf{Latency Level} & \textbf{Fluency ($\uparrow$)} & \textbf{BLEU ($\uparrow$)} & \textbf{AL ($\downarrow$)} \\
\midrule
Low     & 7.25 & 28.27 & 2.47 \\
Medium  & 7.55 & 30.60 & 4.53 \\
High    & 7.81 & 32.44 & 9.44 \\
Offline & 8.29 & 33.28 & 18.80 \\
\bottomrule
\end{tabular}
\caption{Translation fluency, BLEU, and AL of EAST under different latency levels. Fluency scores are rated by GPT-4 on a 0--10 scale.}
\label{tab:fluency}
\end{table}

\subsection{Fluency Evaluation}
\label{apd:fluency}
Table~\ref{tab:fluency} presents the fluency scores of translations generated by EAST under varying latency levels, as rated by GPT-4 on a 0–10 scale. The results are averaged over eight translation directions on the WMT22 test set (X$\rightarrow$En and En$\rightarrow$X).
These results show that while offline translation achieves the highest fluency, our method maintains consistently high fluency across all latency levels, with only a minor drop compared to offline. This demonstrates that our method preserves fluency well.

\subsection{Numeric Results for the Figures}
\label{app:appendix_numeric_results}
We also provide the numeric results for Figures \ref{fig:result_wmt15_bleu} in Tables \ref{tab:numberic_results_wmt15} and for Figures \ref{fig:result_wmt22_bleu}, \ref{fig:result_wmt22_comet}, and \ref{fig:result_wmt22_brt} in Tables \ref{tab:numberic_results_wmt22}.

\begin{table*}[t]
\centering
\resizebox{\textwidth}{!}{
\begin{tabular}{lcccccccccc}
\toprule
\multirow{2}{*}{Models} & \multicolumn{2}{c}{De$\rightarrow$En} & \multicolumn{2}{c}{Zh$\rightarrow$En} & \multicolumn{2}{c}{Ru$\rightarrow$En} & \multicolumn{2}{c}{Cs$\rightarrow$En} & \multicolumn{2}{c}{Average} \\
\cmidrule(lr){2-3} \cmidrule(lr){4-5} \cmidrule(lr){6-7} \cmidrule(lr){8-9} \cmidrule(lr){10-11}
 & BLEU & COMET & BLEU & COMET & BLEU & COMET & BLEU & COMET & BLEU & COMET \\
\midrule
Llama3-MOMT & 31.98 & 84.89 & 25.48 & 81.26 & 39.83 & 85.19 & 44.92 & 86.23 & 35.55 & 84.39 \\
Llama3.1-MOMT & 31.80 & \textbf{84.90} & 26.87 & 81.46 & \textbf{40.45} & 85.42 & \textbf{45.92} & \textbf{86.50} & \textbf{36.26} & \textbf{84.57} \\
Qwen2.5-MOMT & 31.62 & 84.73 & \textbf{27.33} & \textbf{81.78} & 40.15 & \textbf{85.60} & 43.85 & 85.48 & 35.74 & 84.40 \\
EAST-Llama3 & 32.55 & 84.77 & 23.80 & 80.86 & 39.83 & 85.04 & 45.61 & 86.20 & 35.45 & 84.22 \\
EAST-Llama3.1 & \textbf{32.62} & 84.80 & 26.22 & 81.12 & 39.98 & 85.08 & 44.89 & 86.24 & 35.93 & 84.31 \\
EAST-Qwen2.5 & 32.33 & 84.52 & 25.62 & 81.46 & 40.29 & 85.44 & 44.24 & 85.58 & 35.62 & 84.25 \\
\bottomrule
\toprule
\multirow{2}{*}{Models} & \multicolumn{2}{c}{En$\rightarrow$De} & \multicolumn{2}{c}{En$\rightarrow$Zh} & \multicolumn{2}{c}{En$\rightarrow$Ru} & \multicolumn{2}{c}{En$\rightarrow$Cs} & \multicolumn{2}{c}{Average} \\ \cmidrule(lr){2-3} \cmidrule(lr){4-5} \cmidrule(lr){6-7} \cmidrule(lr){8-9} \cmidrule(lr){10-11}
 & BLEU & COMET & BLEU & COMET & BLEU & COMET & BLEU & COMET & BLEU & COMET \\
\midrule 
Llama3-MOMT & 30.45 & 85.63 & 40.68 & 86.53 & 24.83 & 87.27 & 27.92 & 88.36 & 30.97 & 86.95 \\
Llama3.1-MOMT & \textbf{32.11} & \textbf{85.78} & 40.65 & 86.70 & 27.28 & \textbf{87.51} & \textbf{29.75} & \textbf{89.10} & \textbf{32.45} & \textbf{87.27} \\
Qwen2.5-MOMT & 29.57 & 84.62 & \textbf{43.88} & \textbf{87.56} & \textbf{27.86} & \textbf{87.51} & 21.48 & 86.26 & 30.70 & 86.49 \\
EAST-Llama3 & 30.84 & 85.49 & 40.17 & 86.31 & 26.79 & 87.13 & 26.63 & 88.17 & 31.11 & 86.78 \\
EAST-Llama3.1 & 31.13 & 85.68 & 40.75 & 86.38 & 27.10 & 87.20 & 28.53 & 88.67 & 31.88 & 86.98 \\
EAST-Qwen2.5 & 28.70 & 84.32 & 41.97 & 87.17 & 27.66 & 87.29 & 23.75 & 86.45 & 30.52 & 86.31 \\
\bottomrule
\end{tabular}
}
\caption{Offline results for different backbone LLMs on the WMT22 X$\rightarrow$En and En$\rightarrow$X test sets. }
\label{tab:offline_llm_mt}
\end{table*}

\begin{figure*}[t]
\pgfplotsset{
    every axis y label/.append style={at={(-0.12,0.5)}},
    every axis/.append style={line width=1.0pt},
}
\centering
\resizebox{0.262\textwidth}{!}{
\subfigure[De$\rightarrow$En]{
\begin{tikzpicture}[baseline]
\begin{axis}[
    ylabel=COMET,
    xlabel=AL,
    enlargelimits=0.05,
    font=\small,
    width=6.4cm,height=5.6cm,
    legend cell align=left,
    legend style={font=\tiny,
    at={(0.7,0.01)},
    anchor=south,
    legend columns=1},
    xmajorgrids=true,
    ymajorgrids=true,
    grid style=dashed,
    xtick={3,4,5,6,7,8},
]
\addplot[color=color4,mark=triangle*, mark size=1.8pt,line width=0.8pt] coordinates {(2.59,82.04)(2.74,82.27)(3.42,82.95)(3.82,83.23)(5.88,83.97)}; 

\addplot[color=color2,mark=triangle*, mark size=1.8pt,line width=0.8pt] coordinates {(2.57,82.01)(2.73,82.24)(3.43,82.87)(3.83,83.27)(6.16,84.13)}; 

\addplot[color=color7,mark=triangle*, mark size=1.8pt,line width=0.8pt] coordinates {(2.55,81.18)(2.86,81.68)(3.48,82.38)(4.73,83.23)(6.17,83.65)}; 

\addplot[color=color1,mark=triangle*, mark size=1.8pt,line width=0.8pt] coordinates {(3.0,82.12)(3.28,82.28)(5.02,83.36)(5.75,83.61)(7.93,84.11)}; 

\addplot[color=color3,mark=triangle*, mark size=1.8pt,line width=0.8pt] coordinates {(3.03,81.97)(3.42,82.3)(5.36,83.43)(5.91,83.48)(8.29,83.74)}; 

\legend{EAST, EAST-w/o-Offline, EAST-Only-Stage-II, EAST-Single-Stage, EAST-Stage-I}

\end{axis}
\end{tikzpicture}}
}
\hspace{-2mm}
\resizebox{0.242\textwidth}{!}{
\subfigure[Zh$\rightarrow$En]{
\begin{tikzpicture}[baseline]
\begin{axis}[
    xlabel=AL,
    enlargelimits=0.05,
    font=\small,
    width=6.4cm,height=5.6cm,
    legend style={font=\tiny,
    at={(0.5,0.35)},
    anchor=south,
    legend columns=1},
    xmajorgrids=true,
    ymajorgrids=true,
    grid style=dashed,
    xmax=41,
]
\addplot[color=color4,mark=triangle*, mark size=1.8pt,line width=0.8pt] coordinates {(1.31,77.14)(1.56,77.4)(3.6,78.74)(4.88,79.24)(13.79,80.59)}; 

\addplot[color=color2,mark=triangle*, mark size=1.8pt,line width=0.8pt] coordinates {(1.42,77.33)(1.77,77.5)(3.77,78.48)(5.31,79.13)(12.22,80.45)}; 

\addplot[color=color7,mark=triangle*, mark size=1.8pt,line width=0.8pt] coordinates {(0.22,75.87)(2.12,77.22)(3.66,78.25)(8.82,79.75)(14.69,80.26)}; 

\addplot[color=color1,mark=triangle*, mark size=1.8pt,line width=0.8pt] coordinates {(1.23,76.58)(1.82,77.05)(3.87,78.47)(5.56,79.13)(15.87,80.68)}; 

\addplot[color=color3,mark=triangle*, mark size=1.8pt,line width=0.8pt] coordinates {(31.1,78.03)(32.09,77.84)(34.91,77.43)(35.34,77.03)(38.43,76.63)}; 

\end{axis}
\end{tikzpicture}}
}
\hspace{-2mm}
\resizebox{0.255\textwidth}{!}{
\subfigure[Ru$\rightarrow$En]{
\begin{tikzpicture}[baseline]
\begin{axis}[
    xlabel=AL,
    enlargelimits=0.05,
    font=\small,
    width=6.4cm,height=5.6cm,
    legend style={font=\tiny,
    at={(0.64,0.45)},
    anchor=south,
    legend columns=1},
    xmajorgrids=true,
    ymajorgrids=true,
    grid style=dashed,
    xmax=10.0,
]
\addplot[color=color4,mark=triangle*, mark size=1.8pt,line width=0.8pt] coordinates {(2.24,83.66)(2.43,83.83)(3.41,84.22)(4.01,84.35)(6.78,84.6)}; 

\addplot[color=color2,mark=triangle*, mark size=1.8pt,line width=0.8pt] coordinates {(2.2,83.47)(2.42,83.57)(3.51,84.05)(4.36,84.26)(6.98,84.52)}; 

\addplot[color=color7,mark=triangle*, mark size=1.8pt,line width=0.8pt] coordinates {(2.39,83.35)(2.83,83.61)(3.63,83.96)(5.21,84.37)(6.97,84.59)}; 

\addplot[color=color1,mark=triangle*, mark size=1.8pt,line width=0.8pt] coordinates {(2.51,83.45)(2.85,83.6)(4.06,84.02)(4.61,84.21)(7.14,84.55)}; 

\addplot[color=color3,mark=triangle*, mark size=1.8pt,line width=0.8pt] coordinates {(4.69,83.37)(5.14,83.41)(6.6,83.58)(6.97,83.48)(9.66,83.63)}; 

\end{axis}
\end{tikzpicture}}
}
\hspace{-2mm}
\resizebox{0.242\textwidth}{!}{
\subfigure[Cs$\rightarrow$En]{
\begin{tikzpicture}[baseline]
\begin{axis}[
    xlabel=AL,
    enlargelimits=0.05,
    font=\small,
    width=6.4cm,height=5.6cm,
    legend style={font=\tiny,
    at={(0.5,0.35)},
    anchor=south,
    legend columns=1},
    xmajorgrids=true,
    ymajorgrids=true,
    grid style=dashed,
    ytick={75,77,79,81,83,85},
]
\addplot[color=color4,mark=triangle*, mark size=1.8pt,line width=0.8pt] coordinates {(2.3,83.41)(2.55,83.73)(3.82,84.72)(4.43,84.94)(6.96,85.43)}; 

\addplot[color=color2,mark=triangle*, mark size=1.8pt,line width=0.8pt] coordinates {(2.28,83.21)(2.52,83.38)(3.71,84.41)(4.43,84.76)(7.02,85.2)}; 

\addplot[color=color7,mark=triangle*, mark size=1.8pt,line width=0.8pt] coordinates {(2.34,82.94)(2.93,83.54)(4.04,84.25)(5.39,84.69)(7.25,85.05)}; 

\addplot[color=color1,mark=triangle*, mark size=1.8pt,line width=0.8pt] coordinates {(2.33,82.95)(2.7,83.55)(4.03,84.36)(4.76,84.63)(7.38,85.23)};

\addplot[color=color3,mark=triangle*, mark size=1.8pt,line width=0.8pt] coordinates {(2.29,81.35)(2.62,81.76)(4.18,82.93)(4.64,83.29)(6.67,84.01)}; 

\end{axis}
\end{tikzpicture}}
}
~
\resizebox{0.265\textwidth}{!}{
\subfigure[En$\rightarrow$De]{
\begin{tikzpicture}[baseline]
\begin{axis}[
    ylabel=COMET,
    xlabel=AL,
    enlargelimits=0.05,
    font=\small,
    width=6.4cm,height=5.6cm,
    legend style={font=\tiny,
    at={(0.7,0.01)},
    anchor=south,
    legend columns=1},
    legend cell align=left,
    xmajorgrids=true,
    ymajorgrids=true,
    grid style=dashed,
    ytick={75,77,79,81,83,85},
]
\addplot[color=color4,mark=triangle*, mark size=1.8pt,line width=0.8pt] coordinates {(2.58,80.31)(3.09,81.7)(5.39,83.49)(7.54,84.01)(12.2,84.37)}; 

\addplot[color=color2,mark=triangle*, mark size=1.8pt,line width=0.8pt] coordinates {(2.59,80.42)(3.04,81.28)(5.35,83.21)(7.01,83.78)(11.5,84.22)}; 

\addplot[color=color7,mark=triangle*, mark size=1.8pt,line width=0.8pt] coordinates {(2.74,80.41)(3.63,81.86)(5.66,83.34)(9.13,84.39)(12.18,84.52)}; 

\addplot[color=color1,mark=triangle*, mark size=1.8pt,line width=0.8pt] coordinates {(2.76,81.33)(3.24,81.95)(5.74,83.99)(7.93,84.51)(12.48,84.98)}; 

\legend{EAST, EAST-w/o-Offline, EAST-Only-Stage-II, EAST-Single-Stage}

\end{axis}
\end{tikzpicture}}
}
\hspace{-2mm}
\resizebox{0.245\textwidth}{!}{
\subfigure[En$\rightarrow$Zh]{
\begin{tikzpicture}[baseline]
\begin{axis}[
    xlabel=AL,
    enlargelimits=0.05,
    font=\small,
    width=6.4cm,height=5.6cm,
    legend style={font=\tiny,
    at={(0.7,0.01)},
    anchor=south,
    legend columns=1},
    xmajorgrids=true,
    ymajorgrids=true,
    grid style=dashed,
    ymax=86,
]
\addplot[color=color4,mark=triangle*, mark size=1.8pt,line width=0.8pt] coordinates {(2.64,81.69)(2.95,82.11)(4.68,83.7)(6.46,84.3)(10.22,84.87)}; 

\addplot[color=color2,mark=triangle*, mark size=1.8pt,line width=0.8pt] coordinates {(2.66,81.36)(2.99,81.97)(4.6,83.25)(6.89,84.08)(10.19,84.74)}; 

\addplot[color=color7,mark=triangle*, mark size=1.8pt,line width=0.8pt] coordinates {(2.79,81.69)(3.43,82.55)(4.91,83.83)(8.29,85.27)(10.59,85.18)}; 

\addplot[color=color1,mark=triangle*, mark size=1.8pt,line width=0.8pt] coordinates {(2.73,82.0)(3.23,82.8)(4.79,84.05)(6.24,84.67)(10.88,85.55)}; 

\end{axis}
\end{tikzpicture}}
}
\hspace{-2mm}
\resizebox{0.245\textwidth}{!}{
\subfigure[En$\rightarrow$Ru]{
\begin{tikzpicture}[baseline]
\begin{axis}[
    xlabel=AL,
    enlargelimits=0.05,
    font=\small,
    width=6.4cm,height=5.6cm,
    legend style={font=\tiny,
    at={(0.64,0.45)},
    anchor=south,
    legend columns=1},
    xmajorgrids=true,
    ymajorgrids=true,
    grid style=dashed,
]
\addplot[color=color4,mark=triangle*, mark size=1.8pt,line width=0.8pt] coordinates {(2.97,83.73)(3.48,84.6)(5.82,85.52)(7.4,85.63)(10.1,85.84)}; 

\addplot[color=color2,mark=triangle*, mark size=1.8pt,line width=0.8pt] coordinates {(2.99,83.89)(3.47,84.41)(5.76,85.45)(7.26,85.7)(10.03,85.86)}; 

\addplot[color=color7,mark=triangle*, mark size=1.8pt,line width=0.8pt] coordinates {(3.06,83.69)(4.0,84.49)(5.95,85.55)(8.61,85.93)(10.15,86.08)}; 

\addplot[color=color1,mark=triangle*, mark size=1.8pt,line width=0.8pt] coordinates {(3.15,84.31)(3.76,85.04)(6.15,85.66)(7.05,85.83)(10.54,86.1)}; 

\end{axis}
\end{tikzpicture}}
}
\hspace{-2mm}
\resizebox{0.247\textwidth}{!}{
\subfigure[En$\rightarrow$Cs]{
\begin{tikzpicture}[baseline]
\begin{axis}[
    xlabel=AL,
    enlargelimits=0.05,
    font=\small,
    width=6.4cm,height=5.6cm,
    legend style={font=\tiny,
    at={(0.5,0.35)},
    anchor=south,
    legend columns=1},
    xmajorgrids=true,
    ymajorgrids=true,
    grid style=dashed,
]
\addplot[color=color4,mark=triangle*, mark size=1.8pt,line width=0.8pt] coordinates {(3.13,85.23)(3.77,85.98)(6.07,86.75)(7.53,86.83)(9.56,87.2)}; 

\addplot[color=color2,mark=triangle*, mark size=1.8pt,line width=0.8pt] coordinates {(3.07,85.28)(3.58,85.94)(5.73,86.54)(7.21,86.91)(9.72,87.12)}; 

\addplot[color=color7,mark=triangle*, mark size=1.8pt,line width=0.8pt] coordinates {(3.38,85.49)(4.21,86.18)(6.1,86.84)(8.41,87.4)(9.7,87.4)}; 

\addplot[color=color1,mark=triangle*, mark size=1.8pt,line width=0.8pt] coordinates {(3.38,85.94)(4.18,86.49)(6.36,87.3)(7.43,87.37)(10.0,87.66)}; 

\end{axis}
\end{tikzpicture}}
}

\caption{COMET-AL curves for different training strategies on the WMT22 X$\rightarrow$En and En$\rightarrow$X test sets.}
\label{fig:result_wmt22_comet_ablation}
\end{figure*}

\begin{figure*}[t]
\pgfplotsset{
    every axis y label/.append style={at={(-0.12,0.5)}},
    every axis/.append style={line width=1.0pt},
}
\centering
\resizebox{0.262\textwidth}{!}{
\subfigure[De$\rightarrow$En]{
\begin{tikzpicture}[baseline]
\begin{axis}[
    ylabel=BLEURT,
    xlabel=AL,
    enlargelimits=0.05,
    font=\small,
    width=6.4cm,height=5.6cm,
    legend cell align=left,
    legend style={font=\tiny,
    at={(0.7,0.01)},
    anchor=south,
    legend columns=1},
    xmajorgrids=true,
    ymajorgrids=true,
    grid style=dashed,
]
\addplot[color=color4,mark=triangle*, mark size=1.8pt,line width=0.8pt] coordinates {(2.59,69.63)(2.74,69.94)(3.42,70.75)(3.82,71.09)(5.88,72.1)}; 

\addplot[color=color2,mark=triangle*, mark size=1.8pt,line width=0.8pt] coordinates {(2.57,69.72)(2.73,69.93)(3.43,70.65)(3.83,71.16)(6.16,72.27)}; 

\addplot[color=color7,mark=triangle*, mark size=1.8pt,line width=0.8pt] coordinates {(2.55,68.75)(2.86,69.38)(3.48,70.22)(4.73,71.15)(6.17,71.67)}; 

\addplot[color=color1,mark=triangle*, mark size=1.8pt,line width=0.8pt] coordinates {(3.0,69.86)(3.28,70.12)(5.02,71.49)(5.75,71.88)(7.93,72.44)}; 

\addplot[color=color3,mark=triangle*, mark size=1.8pt,line width=0.8pt] coordinates {(3.03,69.93)(3.42,70.28)(5.36,71.79)(5.91,71.81)(8.29,72.28)}; 

\legend{EAST, EAST-w/o-Offline, EAST-Only-Stage-II, EAST-Single-Stage, EAST-Stage-I}

\end{axis}
\end{tikzpicture}}
}
\hspace{-2mm}
\resizebox{0.242\textwidth}{!}{
\subfigure[Zh$\rightarrow$En]{
\begin{tikzpicture}[baseline]
\begin{axis}[
    xlabel=AL,
    enlargelimits=0.05,
    font=\small,
    width=6.4cm,height=5.6cm,
    legend style={font=\tiny,
    at={(0.5,0.35)},
    anchor=south,
    legend columns=1},
    xmajorgrids=true,
    ymajorgrids=true,
    grid style=dashed,
    xmax=41
]
\addplot[color=color4,mark=triangle*, mark size=1.8pt,line width=0.8pt] coordinates {(1.31,63.27)(1.56,63.56)(3.6,65.3)(4.88,65.99)(13.79,67.87)}; 

\addplot[color=color2,mark=triangle*, mark size=1.8pt,line width=0.8pt] coordinates {(1.42,63.45)(1.77,63.73)(3.77,65.0)(5.31,65.72)(12.22,67.7)}; 

\addplot[color=color7,mark=triangle*, mark size=1.8pt,line width=0.8pt] coordinates {(0.22,62.2)(2.12,63.51)(3.66,64.8)(8.82,67.0)(14.69,67.61)}; 

\addplot[color=color1,mark=triangle*, mark size=1.8pt,line width=0.8pt] coordinates {(1.23,62.62)(1.82,63.17)(3.87,65.02)(5.56,66.02)(15.87,68.2)}; 

\addplot[color=color3,mark=triangle*, mark size=1.8pt,line width=0.8pt] coordinates {(31.1,64.27)(32.09,63.93)(34.91,63.39)(35.34,62.84)(38.43,62.37)}; 

\end{axis}
\end{tikzpicture}}
}
\hspace{-2mm}
\resizebox{0.255\textwidth}{!}{
\subfigure[Ru$\rightarrow$En]{
\begin{tikzpicture}[baseline]
\begin{axis}[
    xlabel=AL,
    enlargelimits=0.05,
    font=\small,
    width=6.4cm,height=5.6cm,
    legend style={font=\tiny,
    at={(0.64,0.45)},
    anchor=south,
    legend columns=1},
    xmajorgrids=true,
    ymajorgrids=true,
    grid style=dashed,
    xmax=10,
]
\addplot[color=color4,mark=triangle*, mark size=1.8pt,line width=0.8pt] coordinates {(2.24,74.06)(2.43,74.26)(3.41,74.82)(4.01,74.97)(6.78,75.43)}; 

\addplot[color=color2,mark=triangle*, mark size=1.8pt,line width=0.8pt] coordinates {(2.2,73.96)(2.42,74.07)(3.51,74.73)(4.36,75.07)(6.98,75.36)}; 

\addplot[color=color7,mark=triangle*, mark size=1.8pt,line width=0.8pt] coordinates {(2.39,73.82)(2.83,74.25)(3.63,74.7)(5.21,75.29)(6.97,75.6)}; 

\addplot[color=color1,mark=triangle*, mark size=1.8pt,line width=0.8pt] coordinates {(2.51,73.87)(2.85,74.07)(4.06,74.7)(4.61,74.91)(7.14,75.38)}; 

\addplot[color=color3,mark=triangle*, mark size=1.8pt,line width=0.8pt] coordinates {(4.69,73.98)(5.14,74.13)(6.6,74.27)(6.97,74.18)(9.66,74.31)}; 

\end{axis}
\end{tikzpicture}}
}
\hspace{-2mm}
\resizebox{0.242\textwidth}{!}{
\subfigure[Cs$\rightarrow$En]{
\begin{tikzpicture}[baseline]
\begin{axis}[
    xlabel=AL,
    enlargelimits=0.05,
    font=\small,
    width=6.4cm,height=5.6cm,
    legend style={font=\tiny,
    at={(0.5,0.35)},
    anchor=south,
    legend columns=1},
    xmajorgrids=true,
    ymajorgrids=true,
    grid style=dashed,
    ytick={65,67,69,71,73,75},
]
\addplot[color=color4,mark=triangle*, mark size=1.8pt,line width=0.8pt] coordinates {(2.3,72.05)(2.55,72.54)(3.82,73.95)(4.43,74.27)(6.96,74.92)}; 

\addplot[color=color2,mark=triangle*, mark size=1.8pt,line width=0.8pt] coordinates {(2.28,71.83)(2.52,72.07)(3.71,73.45)(4.43,73.95)(7.02,74.57)}; 

\addplot[color=color7,mark=triangle*, mark size=1.8pt,line width=0.8pt] coordinates {(2.34,71.55)(2.93,72.34)(4.04,73.09)(5.39,73.85)(7.25,74.34)}; 

\addplot[color=color1,mark=triangle*, mark size=1.8pt,line width=0.8pt] coordinates {(2.33,71.71)(2.7,72.48)(4.03,73.55)(4.76,73.9)(7.38,74.61)}; 

\addplot[color=color3,mark=triangle*, mark size=1.8pt,line width=0.8pt] coordinates {(2.29,69.76)(2.62,70.2)(4.18,71.68)(4.64,72.03)(6.67,73.05)}; 

\end{axis}
\end{tikzpicture}}
}
~
\resizebox{0.265\textwidth}{!}{
\subfigure[En$\rightarrow$De]{
\begin{tikzpicture}[baseline]
\begin{axis}[
    ylabel=BLEURT,
    xlabel=AL,
    enlargelimits=0.05,
    font=\small,
    width=6.4cm,height=5.6cm,
    legend style={font=\tiny,
    at={(0.7,0.01)},
    anchor=south,
    legend columns=1},
    legend cell align=left,
    xmajorgrids=true,
    ymajorgrids=true,
    grid style=dashed,
]
\addplot[color=color4,mark=triangle*, mark size=1.8pt,line width=0.8pt] coordinates {(2.58,68.0)(3.09,69.6)(5.39,71.91)(7.54,72.66)(12.2,73.25)}; 

\addplot[color=color2,mark=triangle*, mark size=1.8pt,line width=0.8pt] coordinates {(2.59,68.23)(3.04,69.26)(5.35,71.57)(7.01,72.41)(11.5,73.14)}; 

\addplot[color=color7,mark=triangle*, mark size=1.8pt,line width=0.8pt] coordinates {(2.74,68.46)(3.63,69.96)(5.66,72.06)(9.13,73.35)(12.18,73.55)}; 

\addplot[color=color1,mark=triangle*, mark size=1.8pt,line width=0.8pt] coordinates {(2.76,69.36)(3.24,70.24)(5.74,72.84)(7.93,73.51)(12.48,74.17)}; 

\legend{EAST, EAST-w/o-Offline, EAST-Only-Stage-II, EAST-Single-Stage}

\end{axis}
\end{tikzpicture}}
}
\hspace{-2mm}
\resizebox{0.245\textwidth}{!}{
\subfigure[En$\rightarrow$Zh]{
\begin{tikzpicture}[baseline]
\begin{axis}[
    xlabel=AL,
    enlargelimits=0.05,
    font=\small,
    width=6.4cm,height=5.6cm,
    legend style={font=\tiny,
    at={(0.5,0.35)},
    anchor=south,
    legend columns=1},
    xmajorgrids=true,
    ymajorgrids=true,
    grid style=dashed,
    ytick={60,62,64,66,68,70},
]
\addplot[color=color4,mark=triangle*, mark size=1.8pt,line width=0.8pt] coordinates {(2.64,65.58)(2.95,66.1)(4.68,67.99)(6.46,68.74)(10.22,69.37)}; 

\addplot[color=color2,mark=triangle*, mark size=1.8pt,line width=0.8pt] coordinates {(2.66,65.28)(2.99,65.94)(4.6,67.35)(6.89,68.22)(10.19,68.95)}; 

\addplot[color=color7,mark=triangle*, mark size=1.8pt,line width=0.8pt] coordinates {(2.79,65.93)(3.43,67.06)(4.91,68.26)(8.29,69.85)(10.59,69.84)}; 

\addplot[color=color1,mark=triangle*, mark size=1.8pt,line width=0.8pt] coordinates {(2.73,65.82)(3.23,66.65)(4.79,68.11)(6.24,69.06)(10.88,69.96)}; 

\end{axis}
\end{tikzpicture}}
}
\hspace{-2mm}
\resizebox{0.245\textwidth}{!}{
\subfigure[En$\rightarrow$Ru]{
\begin{tikzpicture}[baseline]
\begin{axis}[
    xlabel=AL,
    enlargelimits=0.05,
    font=\small,
    width=6.4cm,height=5.6cm,
    legend style={font=\tiny,
    at={(0.64,0.45)},
    anchor=south,
    legend columns=1},
    xmajorgrids=true,
    ymajorgrids=true,
    grid style=dashed,
]
\addplot[color=color4,mark=triangle*, mark size=1.8pt,line width=0.8pt] coordinates {(2.97,69.31)(3.48,70.32)(5.82,71.63)(7.4,71.83)(10.1,72.0)}; 

\addplot[color=color2,mark=triangle*, mark size=1.8pt,line width=0.8pt] coordinates {(2.99,69.43)(3.47,70.1)(5.76,71.44)(7.26,71.73)(10.03,72.19)}; 

\addplot[color=color7,mark=triangle*, mark size=1.8pt,line width=0.8pt] coordinates {(3.06,69.3)(4.0,70.52)(5.95,71.77)(8.61,72.34)(10.15,72.53)}; 

\addplot[color=color1,mark=triangle*, mark size=1.8pt,line width=0.8pt] coordinates {(3.15,69.92)(3.76,70.98)(6.15,71.93)(7.05,72.13)(10.54,72.6)}; 

\end{axis}
\end{tikzpicture}}
}
\hspace{-2mm}
\resizebox{0.247\textwidth}{!}{
\subfigure[En$\rightarrow$Cs]{
\begin{tikzpicture}[baseline]
\begin{axis}[
    xlabel=AL,
    enlargelimits=0.05,
    font=\small,
    width=6.4cm,height=5.6cm,
    legend style={font=\tiny,
    at={(0.5,0.35)},
    anchor=south,
    legend columns=1},
    xmajorgrids=true,
    ymajorgrids=true,
    grid style=dashed,
]
\addplot[color=color4,mark=triangle*, mark size=1.8pt,line width=0.8pt] coordinates {(3.13,73.83)(3.77,74.8)(6.07,76.11)(7.53,76.22)(9.56,76.48)}; 

\addplot[color=color2,mark=triangle*, mark size=1.8pt,line width=0.8pt] coordinates {(3.07,73.63)(3.58,74.41)(5.73,75.55)(7.21,76.0)(9.72,76.32)}; 

\addplot[color=color7,mark=triangle*, mark size=1.8pt,line width=0.8pt] coordinates {(3.38,73.71)(4.21,74.95)(6.1,75.88)(8.41,76.56)(9.7,76.62)}; 

\addplot[color=color1,mark=triangle*, mark size=1.8pt,line width=0.8pt] coordinates {(3.38,74.4)(4.18,75.43)(6.36,76.38)(7.43,76.36)(10.0,76.81)}; 

\end{axis}
\end{tikzpicture}}
}

\caption{BLEURT-AL curves for different training strategies on the WMT22 X$\rightarrow$En and En$\rightarrow$X test sets.}
\label{fig:result_wmt22_brt_ablation}
\end{figure*}

\section{Prompt}
\begin{figure*}[ht]
\centering
\includegraphics[width=\textwidth]{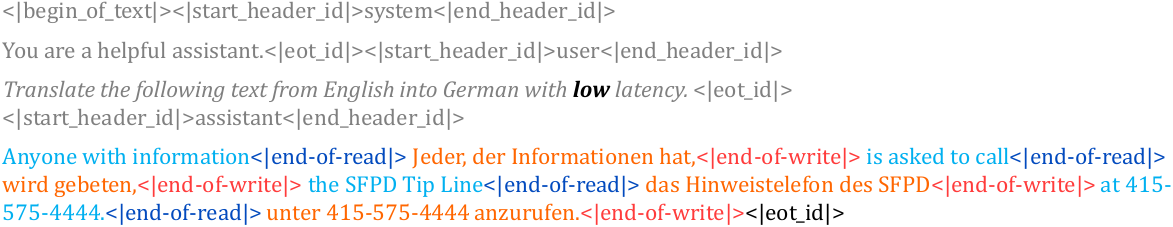}
\caption{An example of the SiMT SFT data for Llama-3. Prompt is colored in gray. The source and target texts are highlighted in cyan and orange, respectively. The read-write tokens are highlighted in blue and red, respectively. We calculate the loss for all tokens other than the prompt during training.}
\label{fig:data_example}
\end{figure*}

\begin{figure*}[t]
\centering
\includegraphics[width=0.9\linewidth]{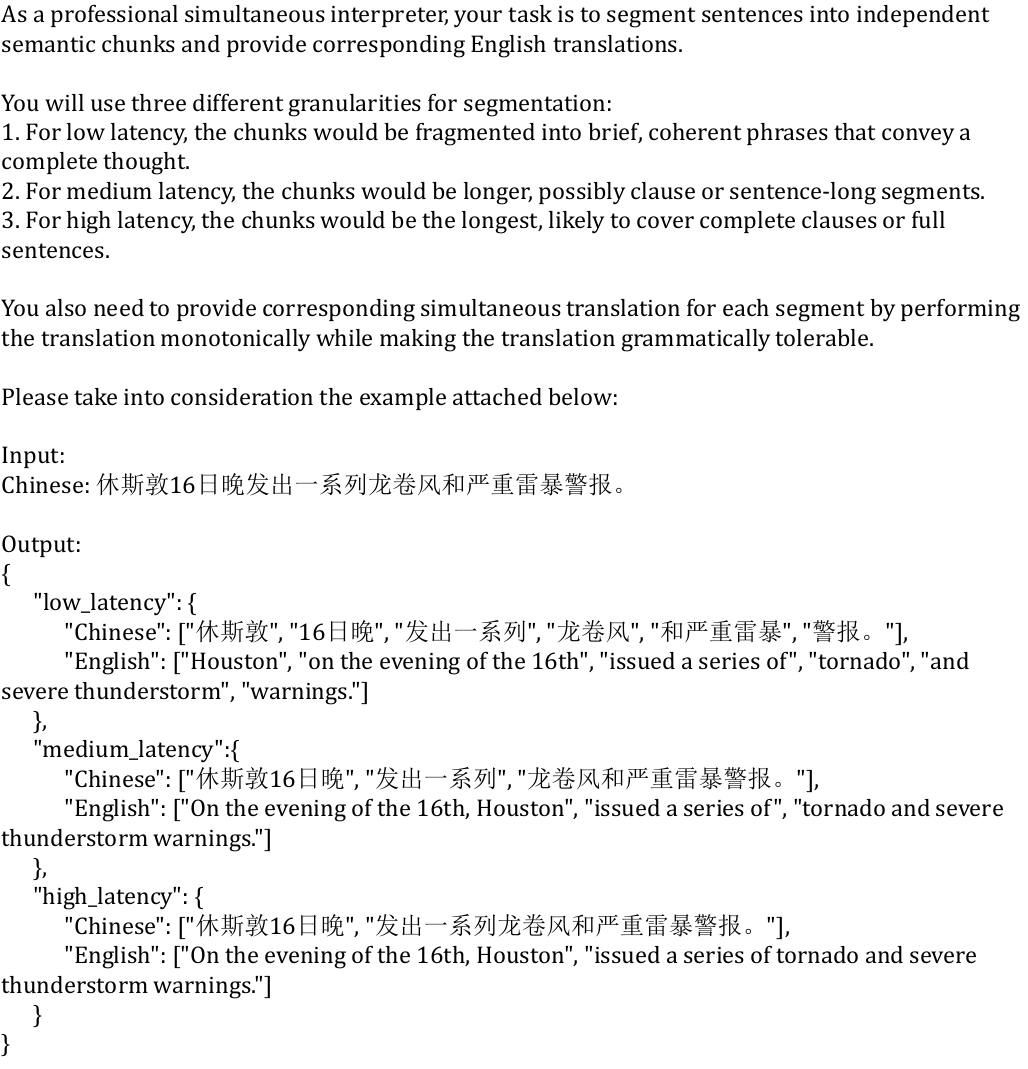}
\caption{The prompt template for GPT-4 to generate SiMT data.}
\label{fig:gpt_prompt}
\end{figure*}

\begin{figure*}[t]
\centering
\includegraphics[width=\linewidth]{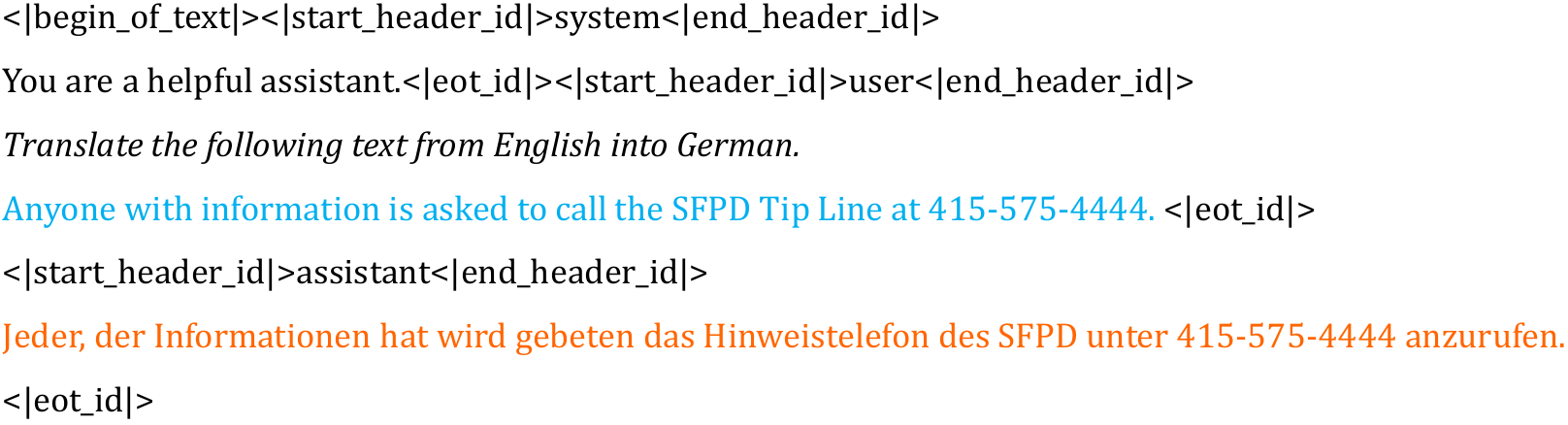}
\caption{An example of the OMT SFT data for Llama-3. The source and target texts are highlighted in cyan and orange, respectively. We compute the loss on the target tokens during training.}
\label{fig:nmt_example}
\end{figure*}

The example for SiMT SFT data is shown in the Figure \ref{fig:data_example}.
The prompt template for generating SiMT data is provided in Figure \ref{fig:gpt_prompt}.
The instruction data for offline translation is shown in the Figure \ref{fig:nmt_example}.
The fluency evaluation prompt is provided in Figure~\ref{fig:fluency_prompt}.

\begin{figure*}
\centering
\begin{tcolorbox}[
  colback=white,
  colframe=gray,
  coltitle=white,
  colbacktitle=gray,
  title=\textbf{Fluency Evaluation},
  fonttitle=\bfseries,
  arc=2mm,
  boxrule=0.8pt,
  left=4mm,
  right=4mm,
  top=2mm,
  bottom=2mm
]
Evaluate the fluency of the translated text based on the following scoring criteria from 0 to 10, with a minimum granularity of 1 point. 
\\
\\
Scoring Criteria:
\begin{itemize}
  \item 10: Perfectly matches the target language's expression habits, with no grammatical, spelling, punctuation, or word order issues. The language is idiomatic and reads as if written by a native speaker.
  \item 8: Mostly natural and fluent, with only minor grammar or word usage issues. The reader can easily understand the entire text and almost won't notice any unnaturalness.
  \item 6: The translation is somewhat stilted, with several grammar or word usage issues. Some sentences require the reader to adjust their understanding. Overall, the original meaning can still be understood.
  \item 4: Poor fluency, with frequent grammatical and expression errors. Reading is laborious but the general idea can still be understood.
  \item 2: Extremely stilted, with most content being hard to understand. The language is disorganized and almost fails to convey the message.
  \item 0: Completely incomprehensible.
\end{itemize}

Please output the score in the following JSON format: 
\{"fluency": "score" \}
\\
\\
Translated text:

\end{tcolorbox}
\caption{Prompt template for fluency evaluation.}
\label{fig:fluency_prompt}
\end{figure*}

\begin{table*}[t]
\centering
\small
\begin{tabular}{lcccccccccc}
\toprule
\multirow{2}{*}{Latency} & \multicolumn{4}{c}{EAST-Stage-I w/ Llama3} & \multicolumn{4}{c}{EAST-Stage-I w/ Llama2} \\
\cmidrule(lr){2-5} \cmidrule(lr){6-9}
 & AL & BLEU & COMET & BLEURT & AL & BLEU & COMET & BLEURT \\
\midrule
Low & 2.68 & 32.46 & 84.28 & 72.52 & 3.18 & 32.55 & 84.32 & 72.62 \\
Low-Medium & 3.06 & 33.12 & 84.63 & 72.90 & 3.46 & 32.80 & 84.51 & 72.88 \\
Medium & 4.77 & 35.21 & 85.66 & 74.29 & 5.84 & 34.62 & 85.49 & 74.28 \\
Medium-High & 5.33 & 35.55 & 85.72 & 74.47 & 7.05 & 34.93 & 85.70 & 74.57 \\
High & 7.78 & 36.62 & 86.11 & 75.02 & 9.73 & 35.49 & 85.91 & 74.83 \\
\bottomrule
\end{tabular}
\caption{Numeric results on WMT15 De$\rightarrow$En test set for EAST-Stage-I w/ Llama3 and EAST-Stage-I w/ Llama2 (Figure \ref{fig:result_wmt15_bleu}).}
\label{tab:numberic_results_wmt15}
\end{table*}

\begin{table*}[h]
\centering
\small
\begin{tabular}{lcccccccccc}
\toprule
\multirow{2}{*}{Latency} & \multicolumn{4}{c}{De$\rightarrow$En} & \multicolumn{4}{c}{En$\rightarrow$De} \\ 
\cmidrule(lr){2-5} \cmidrule(lr){6-9}
 & AL & BLEU & COMET & BLEURT & AL & BLEU & COMET & BLEURT \\
\midrule
Low & 2.59 & 29.87 & 82.04 & 69.63 & 2.58 & 23.27 & 80.31 & 68.00 \\
Low-Medium & 2.74 & 30.09 & 82.27 & 69.94 & 3.09 & 24.57 & 81.70 & 69.60 \\
Medium & 3.42 & 31.09 & 82.95 & 70.75 & 5.39 & 27.00 & 83.49 & 71.91 \\
Medium-High & 3.82 & 31.61 & 83.23 & 71.09 & 7.54 & 27.99 & 84.01 & 72.66 \\
High & 5.88 & 32.38 & 83.97 & 72.10 & 12.20 & 28.77 & 84.37 & 73.25 \\
\bottomrule
\toprule
\multirow{2}{*}{Latency} & \multicolumn{4}{c}{Zh$\rightarrow$En} & \multicolumn{4}{c}{En$\rightarrow$Zh} \\
\cmidrule(lr){2-5} \cmidrule(lr){6-9}
 & AL & BLEU & COMET & BLEURT & AL & BLEU & COMET & BLEURT \\
\midrule
Low & 1.31 & 18.74 & 77.14 & 63.27 & 2.64 & 31.86 & 81.69 & 65.58 \\
Low-Medium & 1.56 & 19.08 & 77.40 & 63.56 & 2.95 & 32.18 & 82.11 & 66.10 \\
Medium & 3.60 & 21.43 & 78.74 & 65.30 & 4.68 & 34.37 & 83.70 & 67.99 \\
Medium-High & 4.88 & 22.34 & 79.24 & 65.99 & 6.46 & 35.32 & 84.30 & 68.74 \\
High & 13.79 & 25.00 & 80.59 & 67.87 & 10.22 & 37.15 & 84.87 & 69.37 \\
\bottomrule
\toprule
\multirow{2}{*}{Latency} & \multicolumn{4}{c}{Ru$\rightarrow$En} & \multicolumn{4}{c}{En$\rightarrow$Ru} \\
\cmidrule(lr){2-5} \cmidrule(lr){6-9}
 & AL & BLEU & COMET & BLEURT & AL & BLEU & COMET & BLEURT \\
\midrule
Low & 2.24 & 36.35 & 83.66 & 74.06 & 2.97 & 23.69 & 83.73 & 69.31 \\
Low-Medium & 2.43 & 36.49 & 83.83 & 74.26 & 3.48 & 24.63 & 84.60 & 70.32 \\
Medium & 3.41 & 37.42 & 84.22 & 74.82 & 5.82 & 25.69 & 85.52 & 71.63 \\
Medium-High & 4.01 & 37.73 & 84.35 & 74.97 & 7.40 & 26.21 & 85.63 & 71.83 \\
High & 6.78 & 38.92 & 84.60 & 75.43 & 10.10 & 26.67 & 85.84 & 72.00 \\
\bottomrule
\toprule
\multirow{2}{*}{Latency} & \multicolumn{4}{c}{Cs$\rightarrow$En} & \multicolumn{4}{c}{En$\rightarrow$Cs} \\
\cmidrule(lr){2-5} \cmidrule(lr){6-9}
 & AL & BLEU & COMET & BLEURT & AL & BLEU & COMET & BLEURT \\
\midrule
Low & 2.30 & 37.47 & 83.41 & 72.05 & 3.13 & 24.88 & 85.23 & 73.83 \\
Low-Medium & 2.55 & 38.32 & 83.73 & 72.54 & 3.77 & 25.48 & 85.98 & 74.80 \\
Medium & 3.82 & 40.80 & 84.72 & 73.95 & 6.07 & 26.96 & 86.75 & 76.11 \\
Medium-High & 4.43 & 41.28 & 84.94 & 74.27 & 7.53 & 27.28 & 86.83 & 76.22 \\
High & 6.96 & 42.99 & 85.43 & 74.92 & 9.56 & 27.62 & 87.20 & 76.48 \\
\bottomrule
\end{tabular}
\caption{Numeric results on WMT22 X$\rightarrow$En and En$\rightarrow$X test sets for EAST (Figures \ref{fig:result_wmt22_bleu}, \ref{fig:result_wmt22_comet}, and \ref{fig:result_wmt22_brt}).}
\label{tab:numberic_results_wmt22}
\end{table*}

\end{document}